%% file: main.tex
\documentclass[pdflatex,bst/sn-mathphys-num]{sn-jnl}

\usepackage{graphicx}%
\usepackage{multirow}%
\usepackage{amsmath,amssymb,amsfonts}%
\usepackage{amsthm}%
\usepackage{mathrsfs}%
\usepackage{bbm}
\usepackage{makecell}
\usepackage[title]{appendix}%
\usepackage{xcolor}%
\usepackage{textcomp}%
\usepackage{manyfoot}%
\usepackage{booktabs}%
\usepackage{algorithm}%
\usepackage{algorithmicx}%
\usepackage{algpseudocode}%
\usepackage{listings}%
\usepackage{float}
\usepackage{cleveref}

\usepackage{fontawesome5}
\usepackage[inline]{enumitem}
\usepackage{array}

\theoremstyle{thmstyleone}

\theoremstyle{thmstyletwo}

\theoremstyle{thmstylethree}

\input{preamble.tex}

\definecolor{mydarkred} {rgb}{0.6,0.0,0.0}
\definecolor{metabg}    {HTML}{F1F6FE}  
\definecolor{metarule}  {HTML}{D8E5F8}
\definecolor{metalink}  {HTML}{2E77D0}
\definecolor{metalabel} {HTML}{1A1A1A}
\definecolor{ghblack}   {HTML}{24292F}
\definecolor{hfyellow}  {HTML}{FFB100}
\definecolor{dbteal}    {HTML}{1B9AAA}
\definecolor{globeblue} {HTML}{3B7DD8}
\definecolor{ytred}     {HTML}{E02F2F}
\definecolor{mailgray}  {HTML}{5A6572}

\newcommand{\hflogo}{\textcolor{hfyellow}{\faSmileBeam}}

\newcommand{\metaicon}[2]{%
  \makebox[1.25em][c]{\textcolor{#1}{\raisebox{-0.1ex}{#2}}}}

\newcommand{\metarow}[5]{%
  \textbf{\textcolor{metalabel}{#3}} &
  \metaicon{#1}{#2}\,\href{#4}{\textcolor{metalink}{#5}}\\}

\providecommand{\firstpagemeta}{}
\renewcommand{\firstpagemeta}{%
  \begingroup
  \centering
  \setlength{\fboxsep}{11pt}%
  \fcolorbox{metarule}{metabg}{%
    \begin{minipage}{0.90\linewidth}
      \small
      \setlength{\tabcolsep}{6pt}%
      \renewcommand{\arraystretch}{1.28}%
      \begin{tabular}{@{}l@{\hspace{0.9em}}l@{}}
        \metarow{globeblue}{\faGlobe}    {Project Page}  {https://targo-benchmark.github.io}                 {targo-benchmark.github.io}
        \metarow{ghblack}  {\faGithub}   {Code}          {https://github.com/targo-benchmark/targo}          {targo-benchmark/targo}
        \metarow{dbteal}   {\faDatabase} {Dataset}       {https://huggingface.co/datasets/randing2000/TARGO} {randing2000/TARGO}
        \metarow{hfyellow} {\hflogo}     {Model Weights} {https://huggingface.co/randing2000/TARGO-Net}      {randing2000/TARGO-Net}
        \metarow{ytred}    {\faYoutube}  {Video}         {https://youtu.be/1aIxo5b2Ppw}                      {youtu.be/1aIxo5b2Ppw}
        \textbf{\textcolor{metalabel}{Contact}} &
        \metaicon{mailgray}{\faEnvelope}\,%
        \href{mailto:yan.xia@ustc.edu.cn}{\textcolor{metalink}{yan.xia@ustc.edu.cn}},\;
        \href{mailto:randing2000@gmail.com}{\textcolor{metalink}{randing2000@gmail.com}},\;
        \href{mailto:ziyuanqin6@gmail.com}{\textcolor{metalink}{ziyuanqin6@gmail.com}}\\
      \end{tabular}
    \end{minipage}}%
  \par\vspace{0.35em}
  {\small\bfseries\textcolor{mydarkred}%
    {Published at International Journal of Computer Vision (IJCV), 2026}\par}
  \vspace{0.2cm}
  \endgroup
}

\begin{document}

\title{TARGO and TARGO-Net: Benchmarking Target-driven Object Grasping under Occlusions}

\author*[1]{\fnm{Yan} \sur{Xia}\textsuperscript{\dag}}
\author[2]{\fnm{Ran} \sur{Ding}\textsuperscript{\dag}}
\author[2]{\fnm{Ziyuan} \sur{Qin}\textsuperscript{\dag}}

\author[4]{\fnm{Guanqi} \sur{Zhan}}
\author[4]{\fnm{Kaichen} \sur{Zhou}}

\author[3]{\fnm{Long} \sur{Yang}}
\author[3]{\fnm{Hao} \sur{Dong}}

\author[2,5]{\fnm{Daniel} \sur{Cremers}}

\affil[1]{University of Science and Technology of China}
\affil[2]{Technical University of Munich}
\affil[3]{CFCS, School of Computer Science, Peking University}
\affil[4]{University of Oxford}
\affil[5]{Munich Center for Machine Learning (MCML)}
\affil[\dag]{Equal Contribution}
\affil[*]{Corresponding Author}

\abstract{Predicting 6-DoF grasp poses from a single RGB-D frame has recently achieved impressive accuracy, yet performance can degrade significantly when the target object is heavily occluded by clutter.
In this paper, we establish TARGO, the first benchmark dataset for TARget-driven Grasping under Occlusions, and introduce TARGO-Net, a grasp prediction model designed for visual occlusion. Our main contributions are fourfold:
\begin{enumerate*}[label=\arabic*)]
    \item We identify the visual occlusion challenge in single-view RGB-D 6-DoF grasping and show that current state-of-the-art models still suffer under severe occlusion.
    \item We propose the TARGO dataset, which supports training and testing 6-DoF grasp models across different visual occlusion levels and enables robustness evaluation in real-world scenarios.
    \item We devise TARGO-Net, a transformer-based grasping model with a target completion module and target-scene cross-attention, which achieves the most robust performance across all visual occlusion levels.
    \item We show that, target completion, number of occluders, and target minimum dimension also relate to grasp success in addition to visual occlusion.
\end{enumerate*}}

\maketitle


\newpage
\input{sections/1_intro}
\input{sections/2_related}
\input{sections/3_data}
\input{sections/4_method}
\input{sections/5_exp}
\input{sections/6_disc}
\input{sections/7_concl}
\input{sections/appendix}
\input{sections/da_statement}

\newpage
\bibliographystyle{ieeetr}
\bibliography{main}

\end{document}

%% file: preamble.tex
\usepackage[utf8]{inputenc} 
\usepackage[T1]{fontenc}    
\usepackage{url}            
\usepackage{booktabs}       
\usepackage{amsfonts}       
\usepackage{fontawesome5}
\usepackage{nicefrac}       
\usepackage[expansion=false]{microtype} 
\usepackage{xcolor}         
\usepackage{enumitem}
\usepackage[inline]{enumitem}
\usepackage{multirow}
\usepackage{longtable}
\usepackage{supertabular}
\usepackage{makecell}
\usepackage{soul}
\usepackage{xcolor}
\usepackage{amsmath}
\usepackage{amssymb}
\usepackage{cleveref}       
\usepackage{bbm}
\usepackage{graphicx}
\usepackage{caption}
\usepackage{subcaption}
\usepackage{wrapfig}
\usepackage{colortbl}
\usepackage{adjustbox}
\usepackage{siunitx}
\usepackage{titlesec}
\usepackage{array}
\usepackage{pifont}             
\usepackage{manfnt}

\newcommand{\cmark}{\ding{51}}   
\newcommand{\xmark}{\ding{55}}   

\titleformat{\paragraph}
{\normalfont\normalsize\bfseries}{\theparagraph}{1em}{}
\titlespacing*{\paragraph}
{0pt}{3.25ex plus 1ex minus .2ex}{1.5ex plus .2ex}

\newcommand{\OL}{\text{OL}}
\newcommand{\SO}{\mathrm{SO}}

\renewcommand\thesection{\Alph{section}}
\renewcommand\thesubsection{\thesection.\arabic{subsection}}
\renewcommand\thesubsubsection{\thesubsection.\arabic{subsubsection}}

\renewcommand{\thefigure}{\Roman{figure}}
\renewcommand{\thetable}{\Roman{table}}

\usepackage{fontawesome5}

\newcommand{\papermetadata}[4][\relax]{%
  \par\noindent
  {\small\bfseries \ifx\relax#1\relax\else#1~\fi#2:}~{\small\href{#3}{#4}}\par
}

\makeatletter
\newcommand{\myfrontleft}{%
  \leftskip=0pt \rightskip=0pt plus 1fil \parfillskip=0pt \parindent=0pt
}
\renewcommand{\@maketitle}{\newpage\null%
    \vskip21pt%
    \hsize\textwidth\parindent0pt%
    \myfrontleft
    {\hbox to \textwidth{{\Artcatfont\ArtType\hfill}\par}}%
    \ifx\@title\empty\else%
        \removelastskip\vskip20pt\nointerlineskip%
        {\Titlefont\@title\par}%
    \fi%
    \ifx\@subtitle\empty\else%
        \vskip9pt%
        {{\SubTitlefont\@subtitle\par}}%
    \fi%
    \ifnum\aucount>0%
        \global\punctcount\aucount%
        \vskip20pt%
        {\myfrontleft\artauthors\par}%
        {\vskip5pt\myfrontleft\addressfont\auaddress\par%
        \ifequalcont{\par$^{\dagger}$\@equalconttext\par}\fi%
	}%
     \fi%
     \removelastskip\vskip18pt%
     {\printabstract\par}%
     {\printkeywords\par}%
     \removelastskip\vskip18pt%
     {\firstpagemeta}%
    \removelastskip\vskip30pt\vskip0pt}%
\makeatother

%% file: sections/1_intro.tex
\section{Introduction}
\label{sec:intro}

Predicting 6-DoF grasps in cluttered environments is a fundamental problem in robotics and computer vision. From a partial and noisy point cloud captured by a single fixed RGB-D camera, a robot must infer object geometry and estimate robust grasp poses for a parallel-jaw gripper. Recent advances in clutter removal~\cite{giga, vgn, zurbrugg2024icgnet} and the availability of large-scale simulated and real-world datasets~\cite{cornell, levine2018learning, eppner2021acronym, fang2020graspnet} have significantly propelled this field forward. However, in target-driven grasping, where the robot must grasp a specified object in clutter, performance can degrade severely when other objects occlude the target.
Although several recent datasets~\cite{wang2021ocid, jung2024housecat6d, back2025graspclutter6d} include occluded scenes, they do not explicitly isolate or quantify how target-specific occlusion severity affects 6-DoF target-driven grasp pose estimation.

To study this problem, we propose TARGO, a benchmark for analyzing the robustness of target-driven grasp models under visual occlusion. TARGO comprises a synthetic dataset, TARGO-Synthetic, and a real-world dataset, TARGO-Real. Derived from VGN~\cite{vgn}, TARGO-Synthetic arranges multiple objects in cluttered tabletop scenes and pairs each with a single scene, enabling controlled analysis of both the presence and the severity of target occlusion. TARGO-Real contains $1{,}000$ scenes selected from DexYCB~\cite{dexycb}, featuring $21$ YCB objects under natural occlusion. It enables zero-shot evaluation of models trained in simulation from real RGB-D observations, with predicted grasps executed in reconstructed simulation.

We investigate the impact of visual occlusion on six representative grasping baselines adapted to the target-driven task. Across controlled occlusion levels in both synthetic and real settings, every baseline shows a consistent decline in grasp success rate as target occlusion increases.

Motivated by data augmentation as a common strategy for mitigating distribution shift~\cite{grover2024revealing}, we find that sampling grasps in single scenes, where only the target is present, improves robustness to visual occlusion. We further propose TARGO-Net, a transformer-based model that captures target--occluder interactions through cross-attention. Unlike GIGA's scene reconstruction~\cite{giga}, TARGO-Net explicitly completes the target shape, which our ablations show is more effective for target-driven grasping. Across the TARGO-Synthetic test set, TARGO-Real, and real-world experiments, TARGO-Net maintains higher grasp success than the evaluated baselines as visual occlusion increases.

In summary, we make the following contributions:
\begin{itemize}
    \item We systematically evaluate target-driven 6-DoF grasping across visual occlusion levels with a simple \textit{visual occlusion level} metric.
    \item We create TARGO, a benchmark comprising large-scale simulated data for training and balanced evaluation, together with real RGB-D observations with natural occlusion for zero-shot evaluation.
    \item We propose TARGO-Net, a transformer-based model with 3D target completion and target-scene fusion for grasp detection. TARGO-Net remains robust even under high visual occlusion and outperforms the baselines in simulation and the real world at all occlusion levels.
    \item We empirically show that target completion, the number of occluders, and the target minimum dimension affect grasp success in addition to visual occlusion level.
\end{itemize}

%% file: sections/2_related.tex
\section{Related Work}
Grasping is widely studied in robotics. We first review occlusion handling in vision and robotics, then grasping datasets and methods for grasping in clutter, and finally 3D shape completion.

\subsection{Occlusion Handling}
Visual occlusion affects both computer vision and robotic systems: it reduces the information available for perception and introduces obstacles for manipulation.
It remains a major challenge in vision tasks such as object recognition~\cite{kortylewski2020compositional,ozguroglu2024pix2gestalt}, detection~\cite{wang2020robust,ke2021deep}, instance segmentation~\cite{yuan2021robust,ke2021deep,zhan2022tri}, tracking~\cite{vanhoorick2023tracking,hsieh2023tracking}, point cloud registration~\cite{qin2024icpuncertainty}, and video action recognition~\cite{modi2024occlusions}. Amodal segmentation further seeks to recover complete object masks from partial observations~\cite{zhu2017semantic,zhan2020self,xu2023amodal,ozguroglu2024pix2gestalt}.
In robotics, occlusion complicates mapping~\cite{stolzle2022reconstructing,zhou2024dynpoint,jefferies2008robot}, object retrieval~\cite{kurenkov2020visuomotor,li2024broadcasting}, manipulation~\cite{li2024mobileafford,wu2024learning}, navigation~\cite{zhou2022devnet,chung2009safe,wang2021learning}, grasping~\cite{sundermeyer2021contact,zeng2022robotic}, and rearrangement~\cite{cheong2020relocate,lee2019efficient}.
Wen et al.~\cite{wen2020robust} study pose estimation when a robotic hand occludes a grasped object, without providing grasp labels, while Yu et al.~\cite{yu2020novel} combine 2D inpainting with RGB-D grasp detection. Our work instead studies how target-specific visual occlusion affects single-view RGB-D 6-DoF grasp prediction. To our knowledge, we are the first to formalize a visual occlusion level for target-driven grasping from single RGB-D views, build an occlusion-centered benchmark around it, and provide a strong baseline for this setting. A complete grasping system also requires motion planning and control~\cite{chi2025diffusion,ze20243d}, but these components are beyond the scope of this paper, and our baseline can be extended with them in future work.

\subsection{Grasping Datasets}
Recent grasp datasets have significantly advanced grasp pose estimation. The Cornell dataset~\cite{cornell} uses manually labeled rectangles, GraspNet-1Billion~\cite{fang2020graspnet} provides dense analytical grasp annotations, and ACRONYM~\cite{eppner2021acronym} and VGN~\cite{vgn} generate grasp labels entirely in simulation.
Specialized datasets add complementary supervision. REGRAD~\cite{zhang2022regrad} models relationships between objects and object-specific grasps but does not control occlusion severity. OCID-Ref~\cite{wang2021ocid} provides scene segmentation and referring expressions but no 6-DoF grasp annotations, and DexGraspNet~\cite{wang2023dexgraspnet} targets dexterous multi-finger hands rather than the parallel-jaw gripper we consider. HouseCat6D~\cite{jung2024housecat6d} and GraspClutter6D~\cite{back2025graspclutter6d} provide parallel-jaw grasps in realistic occluded scenes, and MetaGraspNet and MetaGraspNetV2~\cite{chen2021metagraspnet_v0, gilles2023metagraspnetv2} annotate amodal perception, occlusion reasoning, and both parallel-jaw and vacuum grasping. However, none of these datasets isolates or quantifies target-specific occlusion severity, so a systematic benchmark for parallel-jaw grasping under controlled visual occlusion is still missing.
TARGO fills this gap with paired single and cluttered scenes, a target-specific visual occlusion level, and balanced evaluation across controlled occlusion levels for 6-DoF parallel-jaw grasping, in both synthetic and real scenes.

\subsection{Methods for Grasping in Clutter}
Grasp planning in clutter must reason about collisions with neighboring objects, turning a single-object problem into one of \emph{scene-level} geometry~\cite{kiatos2022learning}.
Early voxel-based approaches such as VGN~\cite{vgn} and GIGA~\cite{giga} use 3D CNNs to regress grasp quality over a TSDF. Contact-based methods such as Edge Grasp Networks~\cite{huang2023edge} and Contact-GraspNet~\cite{sundermeyer2021contact} predict an $\mathrm{SE}(3)$ pose and width directly for points in the cloud, whereas ICGNet~\cite{zurbrugg2024icgnet} augments this with single-view instance masks and object reconstructions.
Although these systems excel at \emph{clutter removal} (removing objects from a tabletop one at a time), their performance drops markedly in \emph{target-driven grasping} (grasping only a specified object) due to visual occlusion (see~\Cref{sec:target_driven_res}).
Target-driven methods address clutter in several ways. Murali et al.~\cite{murali2020} directly plan 6-DoF grasps from partial point clouds using learned collision checking, while G2N2~\cite{lou2022learning} models inter-object relations and uses shape-completion-assisted grasp sampling. Sequential methods push or remove obstructing objects before grasping the target~\cite{yang2020, liu2022ge, motoda2023multi}, whereas active-view methods reposition the camera to obtain a less occluded observation~\cite{breyer2022nbvp, zhang2023nbvp}. The latter two strategies require intermediate manipulation or camera motion, which adds computational and manipulation overhead and limits real-time applicability.
In contrast, TARGO-Net predicts a target grasp from a single fixed RGB-D observation, without rearranging objects or moving the camera. Its single-scene grasp augmentation, target completion network, and target-scene transformer make this one-stage pipeline robust even under severe visual occlusion.

\subsection{Shape Completion}
Thanks to advances in deep learning and the availability of large 3D datasets, shape completion performance has significantly improved~\cite{xia2021vpc, xia2021asfm}. AdaPoinTr~\cite{yu2023adapointr} uses adaptive query generation and a denoising task for efficient point-cloud completion. Shape completion has also been integrated into grasp prediction. GIGA~\cite{giga} shows that grasp affordance and 3D reconstruction are synergistic, and learns both from a shared implicit representation by combining VGN~\cite{vgn} with Convolutional Occupancy Networks~\cite{convonet}. ICGNet~\cite{zurbrugg2024icgnet} likewise reconstructs objects for instance-centric grasp prediction. Rather than completing the entire scene, TARGO-Net adapts AdaPoinTr to complete only the selected target before grasp prediction.

%% file: sections/3_data.tex
\section{Benchmarking Visual Occlusion in Grasping: TARGO Dataset}

To improve learning and evaluate robustness of robotic grasping models under visual occlusion, we introduce a new benchmark dataset named TARGO (TARget-driven Grasping under Occlusion). 
TARGO encompasses scenes with varying levels of visual occlusion and includes both synthetic and real-world datasets.
The synthetic dataset focuses on occlusion learning and evaluation, while the real-world dataset evaluates grasp performance under practical conditions.

\subsection{Design Parameters}
\label{sec:design_params}
We investigate two key attributes of visual occlusion:
\begin{enumerate*}[label=\arabic*)]
    \item \textbf{Existence of occluders} compares scenes containing only the target object (single scenes) vs. those with all objects (cluttered scenes).
    \item \textbf{Level of visual occlusion} quantifies the degree to which the target object is blocked in the depth image.
\end{enumerate*}
By analyzing these attributes, we aim to comprehensively understand the impact of visual occlusion on robotic grasping performance.

\subsubsection{Existence of Occluders}
\label{sec:existence_occluders}
\begin{figure*}[hbtp]
    \centering
    \begin{minipage}[t]{0.63\linewidth}
        \vspace{0pt}  
        \centering
        \begin{subfigure}[b]{0.48\linewidth}
            \centering
            \includegraphics[width=\linewidth]{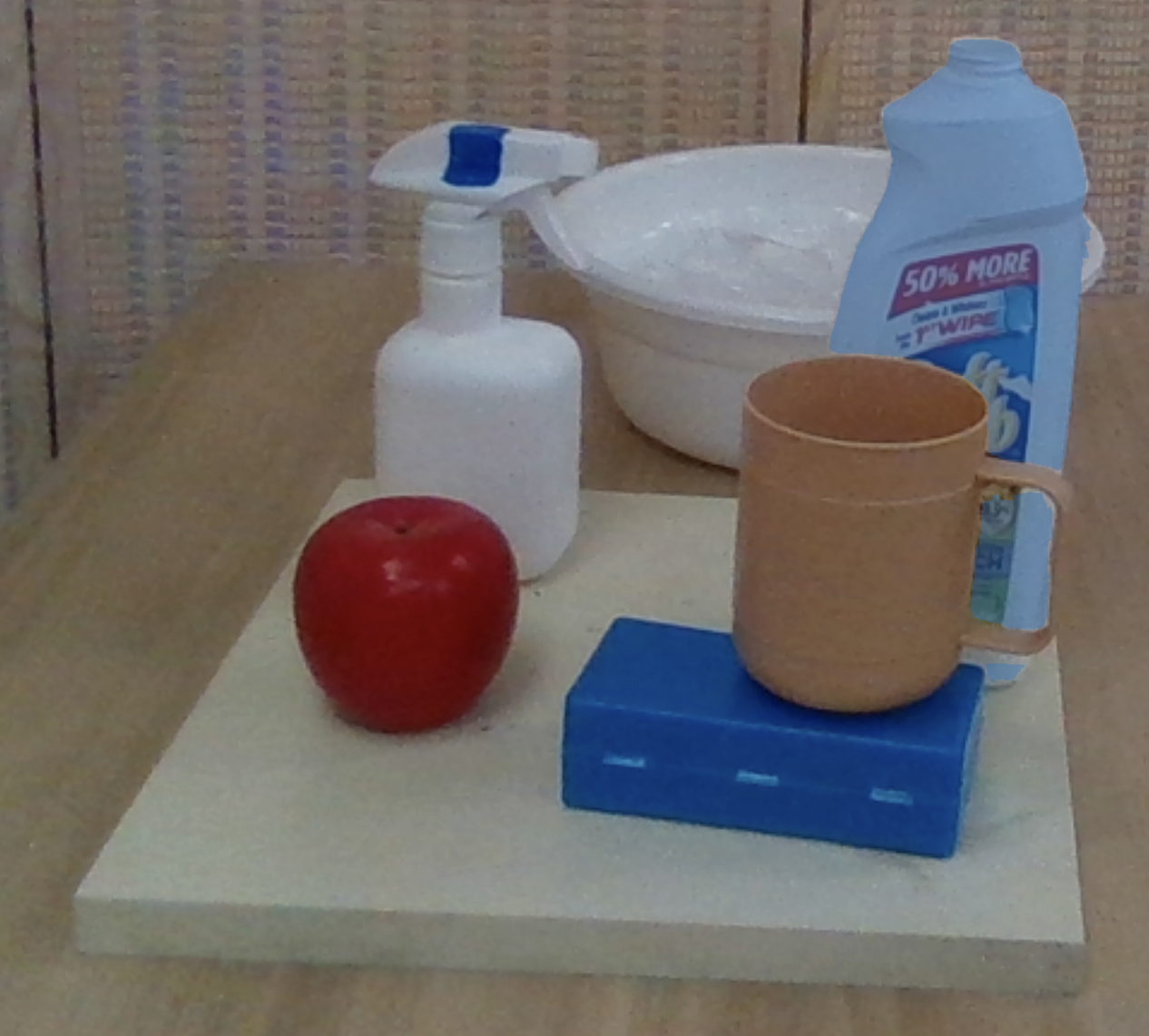}
            \caption{Cluttered scene}
            \label{fig:clutter_scene}
        \end{subfigure}
        \hfill
        \begin{subfigure}[b]{0.48\linewidth} 
            \centering
            \includegraphics[width=\linewidth]{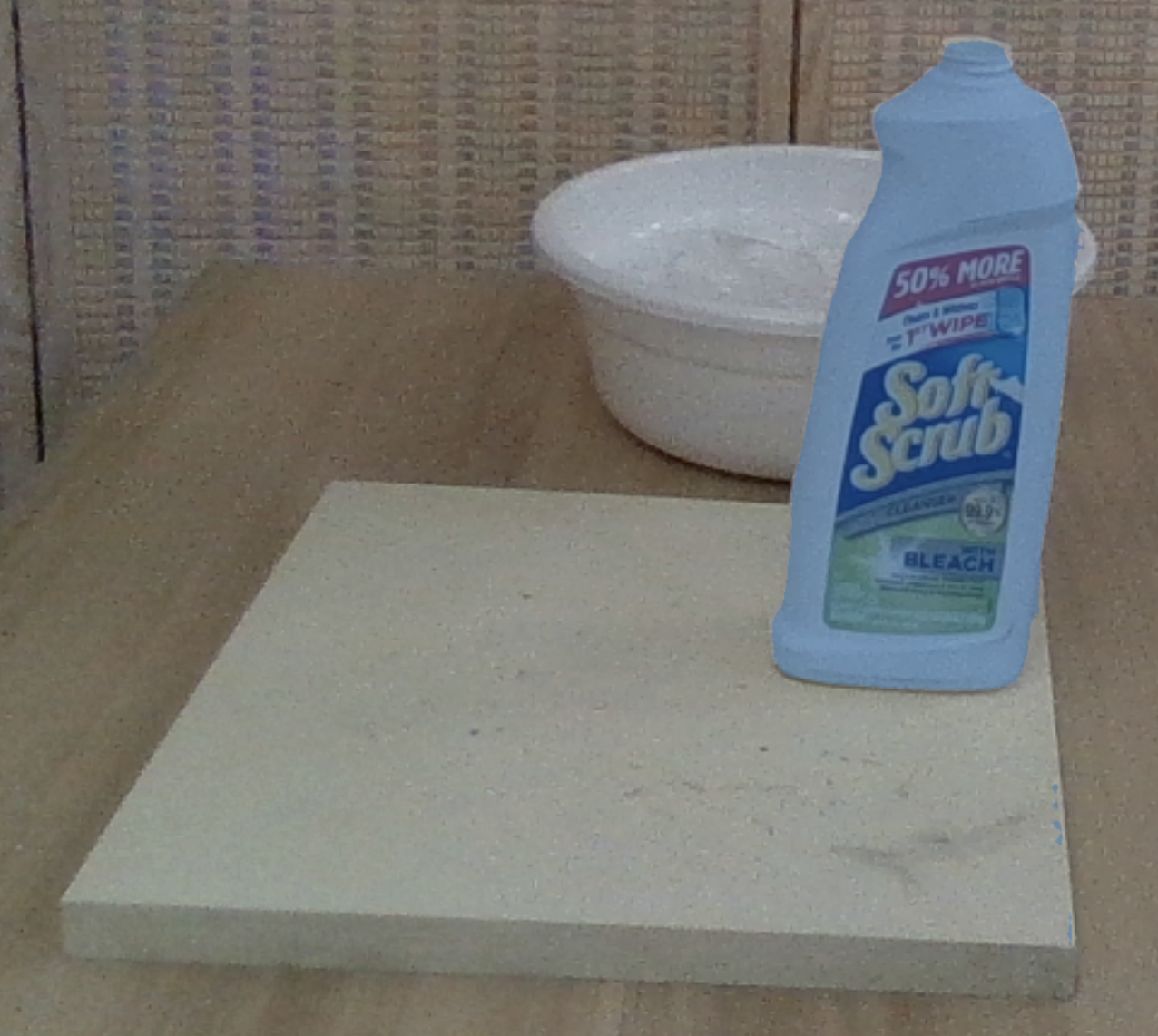}
            \caption{Single scene}
            \label{fig:single_scene}
        \end{subfigure}
        \caption{The target bleacher, marked by a blue overlay, in the cluttered and single scenes.}
        \label{fig:single_cluttered}
    \end{minipage}
    \hfill
    \begin{minipage}[t]{0.33\linewidth}
        \vspace{0pt}  
        \centering
        \includegraphics[width=\linewidth]{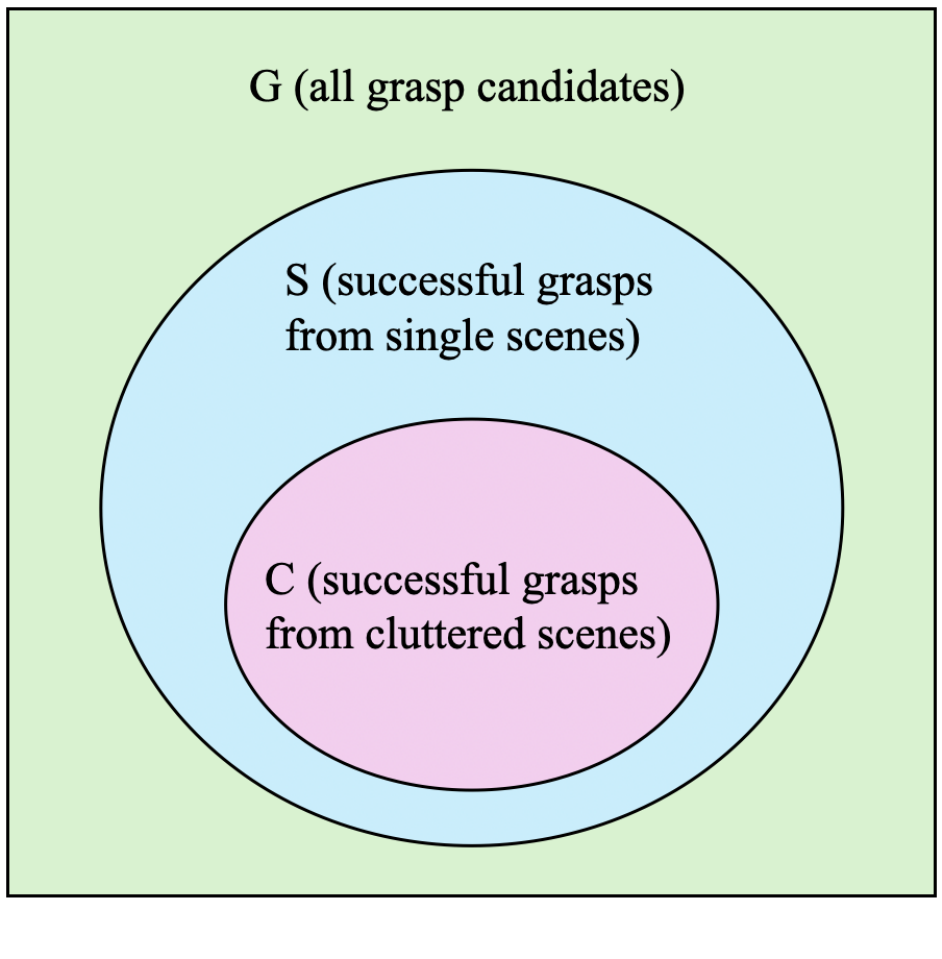}
        \caption{Relationship of the three sets: $G$, $S$, and $C$.}
        \label{fig:venn_diagram}
    \end{minipage}
\end{figure*}

In target-driven grasping, a parallel-jaw gripper picks a designated target from a \emph{cluttered scene} observed by a fixed, single-view RGB-D camera (\Cref{fig:demos_rw}). Other objects can both obscure the target and physically constrain feasible grasps. We call these objects \emph{occluders}. The corresponding \emph{single scene} contains only the target, with its pose and the camera pose unchanged (\Cref{fig:single_cluttered}).

For a given target, let $G$ denote the set of all candidate grasps, $S$ the subset that succeeds in the single scene, and $C$ the subset of $S$ that remains successful when executed in the cluttered scene. Thus, $C \subseteq S \subseteq G$ (\Cref{fig:venn_diagram}), and the candidate grasps form three mutually exclusive groups:
\begin{enumerate}
    \item $G \setminus S$: inherently unsuccessful grasps that fail even in the single scene (e.g., due to table collisions or gripper limits),
    \item $S \setminus C$: clutter-induced failures that succeed in the single scene but fail when occluders are present,
    \item $S \cap C = C$: robust grasps that succeed both with and without occluders.
\end{enumerate}

In TARGO-Synthetic training, rather than sampling grasps directly in \emph{cluttered scenes}, we first sample grasps in the corresponding \emph{single scenes}. We propose a two-stage procedure that preserves both robust grasps and those that are feasible but occlusion-sensitive. Concretely, for \emph{each target}:
\begin{enumerate*}[label=\arabic*)]
  \item In the \emph{single scene}, sample and execute $N$ grasps from $G$, of which $N_S$ succeed and belong to $S$, while the remaining $N-N_S$ fail and belong to $G\setminus S$.
  \item In the \emph{cluttered scene}, re-execute only the $N_S$ successful grasps\footnote{Grasps that fail in the single scene are not re-executed in the corresponding cluttered scene, because they would definitely fail with occluders present.}, of which $N_C$ remain successful and belong to $C$, while the remaining $N_S - N_C$ fail and belong to $S\setminus C$.
\end{enumerate*}

This design offers two key advantages:
\begin{enumerate*}[label=\arabic*)]
  \item \textbf{Fine-grained failure analysis.} We explicitly separate inherent failures ($G\setminus S$) from occlusion-induced ones ($S\setminus C$), enabling the model to learn precisely how occluders invalidate feasible grasps.
  \item \textbf{Efficient data augmentation.} Since any grasp that fails in the single scene also fails in cluttered scene ($G\setminus S \; \subseteq \; G\setminus C$), we only have to re-test the grasps that succeed in the single scene when evaluating the cluttered scene. As a result, with only $N_S$ extra executions per scene\footnote{Empirically, $N_S/N \approx 1/4$.}, TARGO-Net gains over $10\%$ in GSR (see~\Cref{sec:single_scene_analysis}).
\end{enumerate*}

\subsubsection{Levels of Visual Occlusion}
\label{sec:visual_occlusion_level}

We quantify the \emph{visual occlusion level} ($\OL$) from the target masks in a paired single and cluttered depth image. Let $n_{\mathrm{single}}$ and $n_{\mathrm{cluttered}}$ be the respective target-pixel counts. We define
\begin{equation}
    \OL=1-\frac{n_{\mathrm{cluttered}}}{n_{\mathrm{single}}}.
\end{equation}
Because the target and camera poses are fixed within each pair, this ratio measures the fraction of the target's single-view projection hidden by other objects.

\begin{figure}
    \centering
    \includegraphics[width=0.9\linewidth]{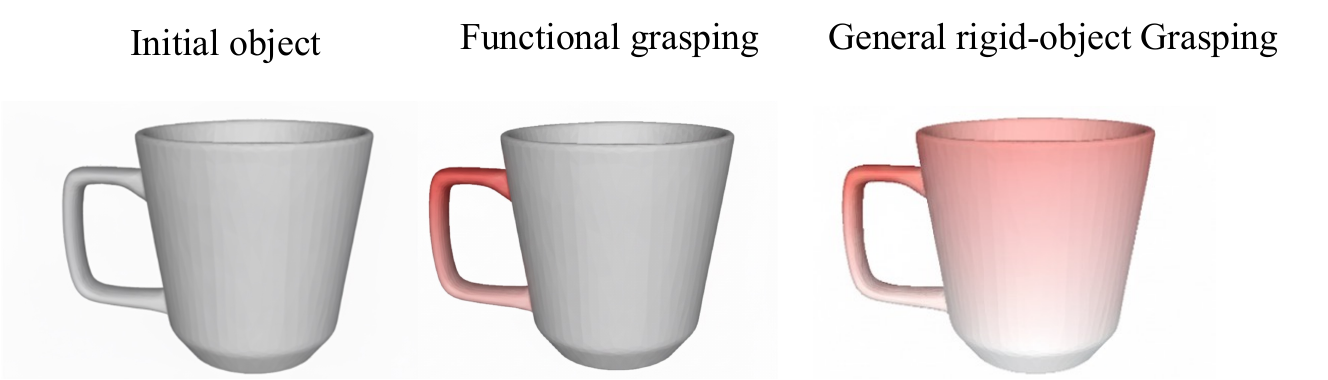}
    \caption{Functional grasping vs. general rigid-object grasping.}
    \label{fig:grasping_type}
\end{figure}

This image-space metric is intentionally simple and aligned with our goal of general rigid-object grasping rather than functional grasping.~\Cref{fig:grasping_type} illustrates the difference on a mug: functional grasping designates a specific region such as the handle~\cite{guo2023handal}, whereas general rigid-object grasping accepts any mechanically valid region, including the handle, lip, body, or base. Because TARGO evaluates the latter, $\OL$ does not weight particular object parts and instead stratifies evaluation by overall target visibility. Despite its simplicity, this measure is sufficient to expose the steep decline of all evaluated baselines as visibility decreases, whereas TARGO-Net degrades far less (\Cref{fig:abl_baseline_comparison}).

\begin{figure*}[htbp]
    \centering
        \includegraphics[width=\linewidth]{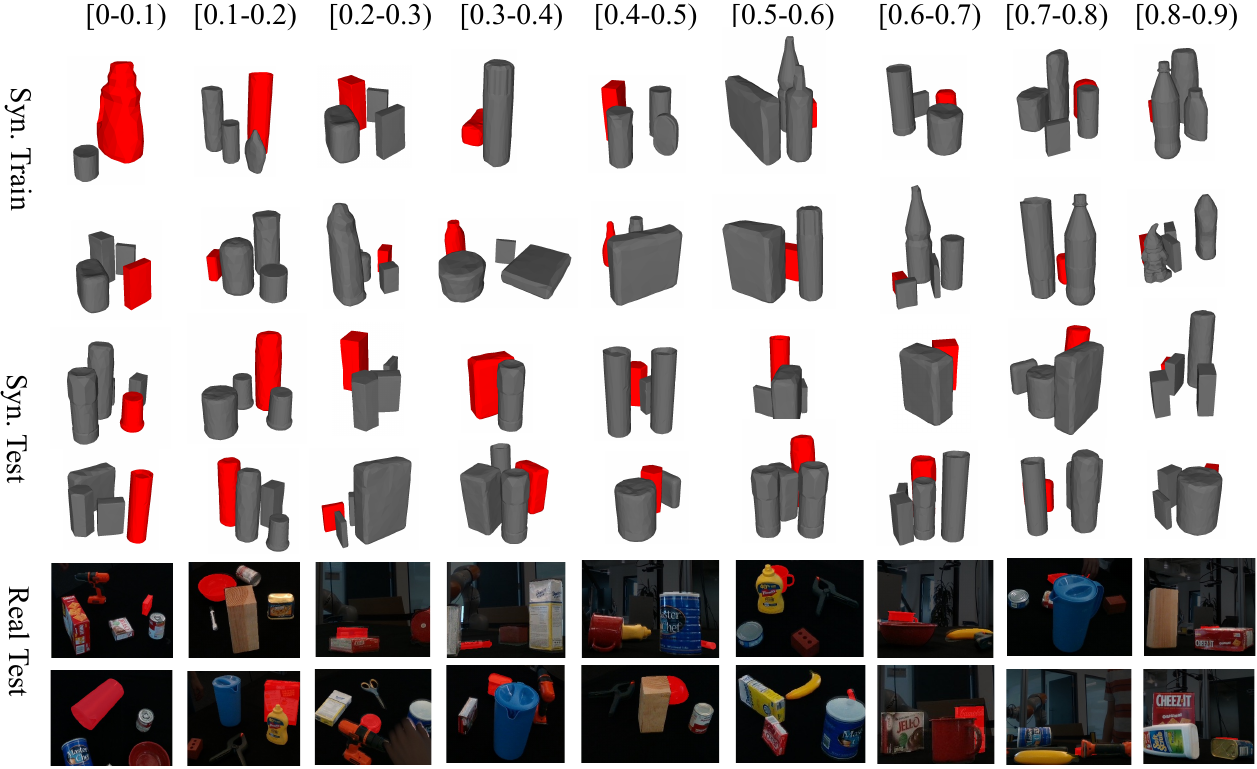}
        \caption{Examples from TARGO-Synthetic training and test sets, as well as TARGO-Real dataset, spanning visual occlusion levels from $[0,0.1)$ to $[0.8, 0.9)$. Red denotes the grasp target.}
        \label{fig:data_overview}
        \vspace{5mm}
\end{figure*}

\begin{figure*}[htbp]
        \centering
        \begin{subfigure}[b]{0.29\textwidth}
            \centering
            \includegraphics[width=\linewidth]{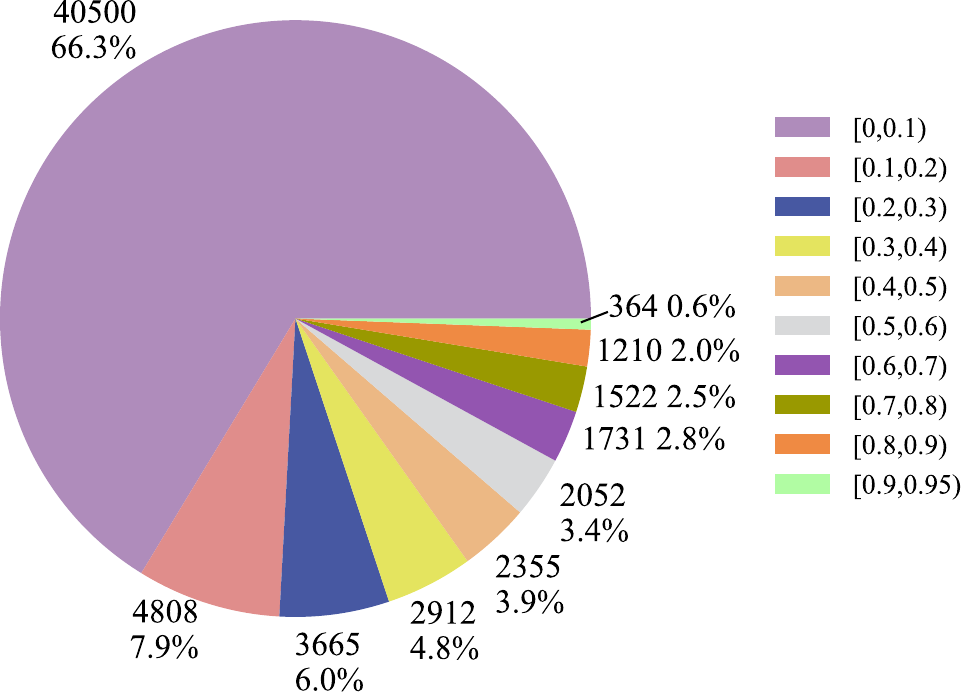}
            \caption{TARGO-Synthetic train}
            \label{fig:target_counts_vs_occ_level_syn_train}
        \end{subfigure}
        \hspace{0.01\textwidth}
        \begin{subfigure}[b]{0.325\textwidth}
            \centering
            \includegraphics[width=\linewidth]{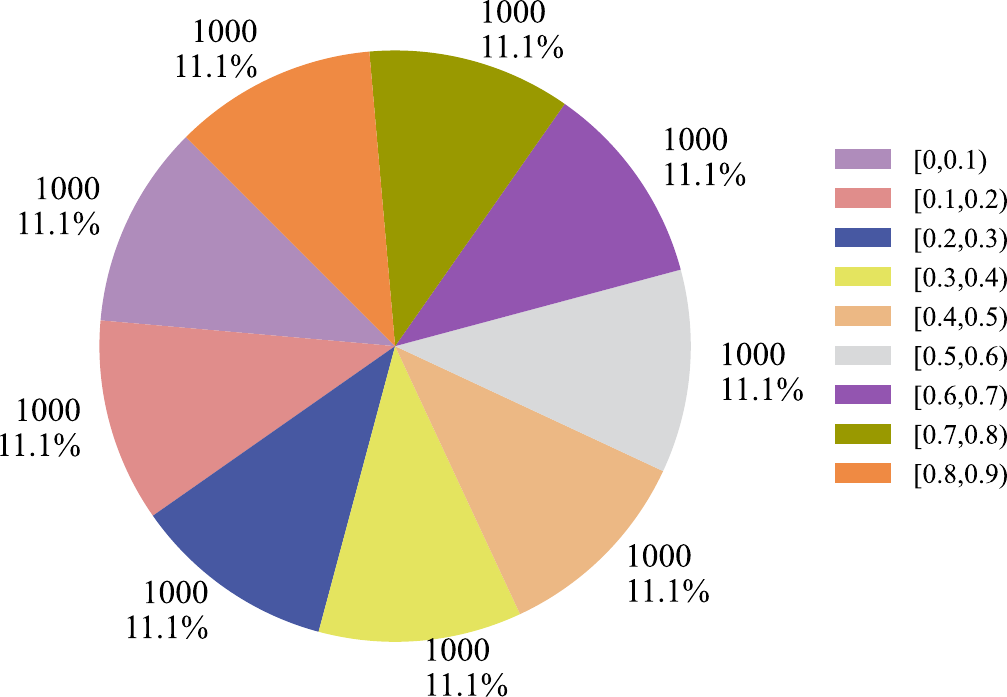}
            \caption{TARGO-Synthetic test}
            \label{fig:target_counts_vs_occ_level_syn_test}
        \end{subfigure}
        \hspace{0.01\textwidth}
        \begin{subfigure}[b]{0.33\textwidth}
            \centering
            \includegraphics[width=\linewidth]{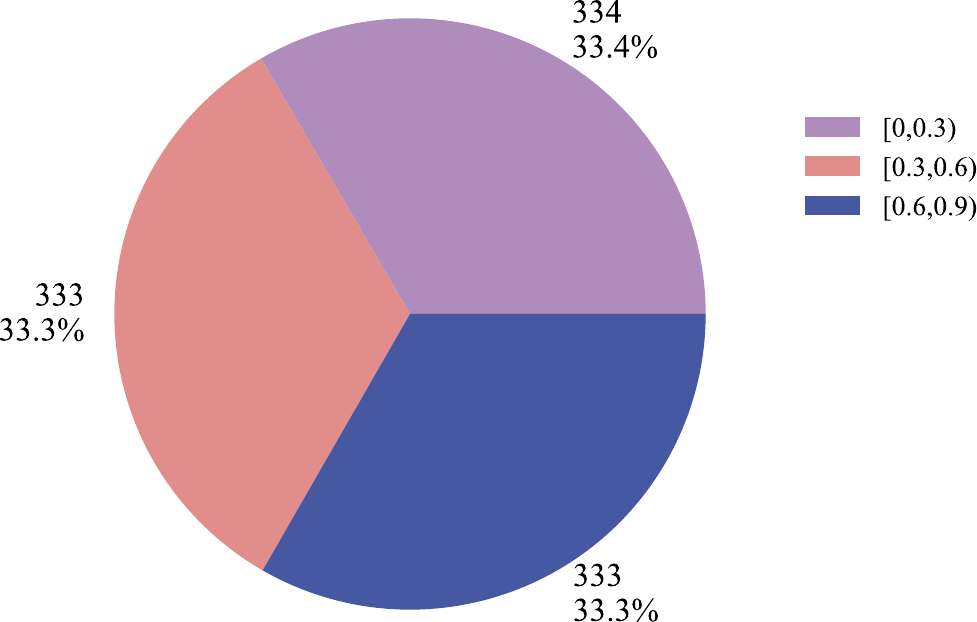}
            \caption{TARGO-Real}
            \label{fig:target_counts_vs_occ_level_real_test}
        \end{subfigure}
        \caption{TARGO target counts in different visual occlusion levels.}
        \label{fig:target_counts_vs_occ_level}
\end{figure*}

\begin{table*}[t!]
    \centering
    \caption{Comparison of publicly available grasp datasets. Grasp labels are generated either manually
    (\ding{45}), by physics simulation
    (\faPlayCircle), analytical models
    ($f$), or real-world trials
    (\faEye). Planar/3D: $\square$ denotes planar grasp, \mancube~denotes 3D grasp.}
    \vspace{0.2cm}
    \setlength{\tabcolsep}{3.5pt} 
    \small 
    \setlength{\extrarowheight}{0.25cm}
    \begin{tabular}{|l|c|c|c|c|c|c|c|}
        \hline
        \textbf{Dataset} & \rotatebox{90}{\makecell[c]{\\Occlusion}} & \rotatebox{90}{\makecell[c]{Planar/3D}} & \rotatebox{90}{\makecell[c]{Grasp\\Type}} & \rotatebox{90}{\makecell[c]{Observa-\\tions}} & \rotatebox{90}{Labels} & \rotatebox{90}{\parbox{0.9cm}{Objects\\(Per Scene)}} & \rotatebox{90}{\parbox{0.9cm}{Grasps\\(Per Scene)}} \\
        \hline\hline

        Cornell \cite{cornell} & \xmark & $\square$ & Rect. & Real & \ding{45} & {\footnotesize\scalebox{1.05}{240}} & \parbox{0.9cm}{\centering {\footnotesize\scalebox{1.05}{8019}}\\{\footnotesize\scalebox{1.05}{($\sim$8)}}} \\
        
        \rowcolor{gray!15}
        Levine et al.~\cite{levine2018learning} & \xmark & $\square$ & Rect. & Real & \faEye & - & \parbox{0.9cm}{\centering {\footnotesize\scalebox{1.05}{800k}}\\{\footnotesize\scalebox{1.05}{(1)}}} \\
        
        REGRAD~\cite{zhang2022regrad} & \xmark & \mancube & 6-DoF + Rect. & Sim & \faPlayCircle & - & \parbox{0.9cm}{\centering {\footnotesize\scalebox{1.05}{100M}}\\{\footnotesize\scalebox{1.05}{(1.02K)}}} \\
        
        \rowcolor{gray!15}
        MetaGraspNetV2~\cite{gilles2023metagraspnetv2} & \xmark & \mancube & 6-DoF & Sim + Real & $f$ & - & - \\
        
        GraspNet-1Billion \cite{fang2020graspnet} & \xmark & \mancube & 6-DoF + Rect. & Real & $f$ & \parbox{0.9cm}{\centering {\footnotesize\scalebox{1.05}{88}}\\{\footnotesize\scalebox{1.05}{($\sim$10)}}} 
        & \parbox{0.9cm}{\centering {\footnotesize\scalebox{1.05}{$\sim$1.2B}}\\{\footnotesize\scalebox{1.05}{(3-9M)}}} \\
        
        \rowcolor{gray!15}
        CGNet \cite{yu2022cgnet} & \xmark & \mancube & 6-DoF + Rect. & Real & $f$ & \parbox{0.9cm}{\centering {\footnotesize\scalebox{1.05}{88}}\\{\footnotesize\scalebox{1.05}{($\sim$10)}}} & \parbox{0.9cm}{\centering {\footnotesize\scalebox{1.05}{$\sim$1.2B}}\\{\footnotesize\scalebox{1.05}{(3-9M)}}} \\
        
        ACRONYM~\cite{eppner2021acronym} & \xmark & \mancube & 6-DoF & Sim & \faPlayCircle & {\footnotesize\scalebox{1.05}{8872}} & \parbox{0.9cm}{\centering {\footnotesize\scalebox{1.05}{17.7M}}\\{\footnotesize\scalebox{1.05}{(2000)}}} \\
        
        \rowcolor{gray!15}
        VGN \cite{vgn} & \xmark & \mancube & 6-DoF & Sim & \faPlayCircle & \parbox{0.9cm}{\centering {\footnotesize\scalebox{1.05}{114}}\\{\footnotesize\scalebox{1.05}{($\sim$5)}}} & \parbox{0.9cm}{\centering {\footnotesize\scalebox{1.05}{$\sim$6M}}\\{\footnotesize\scalebox{1.05}{(120)}}} \\
        
        \textbf{TARGO (Ours)} & \cmark & \mancube & 6-DoF & \textbf{Sim + Real} & \faPlayCircle & \parbox{0.9cm}{\centering {\footnotesize\scalebox{1.05}{135}}\\{\footnotesize\scalebox{1.05}{($\sim$5)}}} & \parbox{0.9cm}{\centering {\footnotesize\scalebox{1.05}{$\sim$6M}}\\{\footnotesize\scalebox{1.05}{($\sim$350)}}} \\
        \hline
    \end{tabular}
    \label{tab:dataset_info}
    \vspace{-0.2cm}
\end{table*}

\subsection{Benchmark Datasets}
\label{sec:benchmark_dataset}

We present synthetic and real-world datasets to thoroughly investigate the occlusion challenge in robotic grasping. The TARGO-Synthetic dataset can be used to train and evaluate the 6-DoF grasping models under various visual occlusion levels, and the TARGO-Real dataset can be tested for zero-shot transfer to evaluate real-world grasping performance.

\Cref{tab:dataset_info} compares the TARGO dataset with publicly available grasp datasets~\cite{zhang2022regrad, gilles2023metagraspnetv2, cornell, levine2018learning, fang2020graspnet, yu2022cgnet, eppner2021acronym, vgn}. TARGO uniquely samples 6-DoF grasps from both single and cluttered scenes, utilizes $21$ YCB objects~\cite{calli2015ycb} in real-world testing, and provides approximately $350$ grasps per scene. While simulation datasets offer scalability, their depth images and grasps may differ from real-world conditions. Conversely, real-world datasets provide realistic inputs and natural occlusion but may have fewer scenes and limited object types. Our combination of simulated and real-world data allows scalable, realistic grasp generation applicable to practical scenarios. Notably, our dataset is the first to explicitly address the effect of visual occlusion on robotic grasping, as illustrated in~\Cref{fig:data_overview}.

\subsubsection{TARGO-Synthetic Dataset}
Generated with PyBullet~\cite{pybullet}, TARGO-Synthetic has disjoint training and test object sets. Collision-free grasps are sampled within a $30 \times 30 \times 30\,\mathrm{cm}^3$ workspace using 114 training objects and 16 test objects from VGN~\cite{vgn}. Dataset generation took approximately 11 hours with 20 parallel processes on an Intel Xeon E5-2697 server.

\paragraph{TARGO-Synthetic Training Dataset}
As mentioned in~\Cref{sec:existence_occluders}, our TARGO-Synthetic training dataset samples from single scenes and tests the successful grasps in the cluttered scenes. This design helps grasping models better learn how occlusion affects feasible grasps with little additional cost. As single scenes offer complete target point clouds corresponding to the partial occluded point clouds, our TARGO-Synthetic dataset can also be utilized to train and test our TARGO-Net's \emph{target completion module} (detailed in~\Cref{sec:shape_completion}). In addition, our TARGO-Synthetic training dataset can directly support other robotic grasping tasks such as \emph{clutter removal}, as all scene objects are treated as targets and include sampled grasps augmented with single scenes.

The dataset generation process is as follows. For each cluttered scene, approximately five objects are randomly placed in the workspace, with each object selected as a target in turn. By comparing these cluttered scenes to their corresponding single scenes, visual occlusion levels can be calculated (\Cref{fig:single_cluttered}). Overall, the TARGO-Synthetic training dataset comprises $61{,}119$ target objects across $16{,}640$ cluttered scenes, categorized into ten visual occlusion bins ranging from $[0, 0.1)$ to $[0.9, 0.95)$ (\Cref{fig:target_counts_vs_occ_level_syn_train}). Due to random placement, most targets fall within the lowest occlusion bin $[0, 0.1)$.

\paragraph{TARGO-Synthetic Test Dataset}
Prior clutter-removal evaluations commonly run approximately five online rounds of about $100$ randomly generated scenes each~\cite{vgn, giga, huang2023edge}. Because random placement leaves most targets only lightly occluded (\Cref{fig:target_counts_vs_occ_level_syn_train}), such evaluations can mask the performance decline under \emph{high occlusion}. Online generation also adds evaluation time, limits the number of trials, and hinders repeatability. Therefore, we build the TARGO-Synthetic test set with
$9{,}000$ \emph{offline} scenes--$1{,}000$ scenes for each of the nine occlusion-level bins from $[0,0.1)$ to $[0.8,0.9)$ (illustrated in~\Cref{fig:target_counts_vs_occ_level_syn_test}).
The \emph{large, balanced and offline} TARGO-Synthetic test set allows for reliable and fast evaluation with attention to \emph{high occlusion}.

We generate the TARGO-Synthetic test dataset by rejection sampling. In each round, approximately five objects are randomly selected and placed on the tabletop at their canonical poses. For each target, the paired single scene provides its unoccluded single-view projection, from which $\OL$ is calculated. Scenes are accepted according to the remaining occlusion level quotas until every level contains exactly $1{,}000$ scenes.

\subsubsection{TARGO-Real Dataset}
Unlike the TARGO-Synthetic dataset, TARGO-Real captures real-world occlusion with raw noisy depth, varied occlusion configurations, and distinct object shapes.
The dataset includes $21$ YCB objects~\cite{calli2015ycb} across $1{,}000$ scenes selected from DexYCB~\cite{dexycb}.
Each occlusion level bin $[0, 0.3)$, $[0.3, 0.6)$, and $[0.6, 0.9)$ is balanced, as depicted in~\Cref{fig:target_counts_vs_occ_level_real_test}, ensuring accurate and comprehensive evaluations across all occlusion ranges.

We use the ground-truth object poses and meshes provided by DexYCB to reconstruct each scene in PyBullet~\cite{pybullet}. Each scene contains around five objects in a $70 \times 70 \times 70\,\mathrm{cm}^3$ workspace, larger than the $30 \times 30 \times 30\,\mathrm{cm}^3$ workspace of TARGO-Synthetic. With real-world observations from DexYCB, we predict and evaluate grasp poses within the simulated environment\footnote{A grasp is considered successful if a collision-free grasp can pick up the target object.}. This real-to-sim evaluation closely aligns with a real-world setup, accounting for noisy observations, inaccurate calibration, planning inaccuracies, and control challenges~\cite{vgn, giga}.

Because TARGO-Real is used for execution-based evaluation rather than supervised training, it does not include ground-truth 6-DoF grasp poses.

\subsection{Target-driven Grasping Models}
While many existing grasping models~\cite{vgn, giga, huang2023edge, zurbrugg2024icgnet} remove all objects in a cluttered scene, our task is to \emph{directly grasp the target}. Thus, we adapt them specifically to target-driven grasping in cluttered environments as our baseline models.
All these models are trained under supervision on the TARGO-Synthetic training dataset, assessed on the TARGO-Synthetic test dataset, and zero-shot transferred to the TARGO-Real dataset (still evaluated in simulation), to comprehensively evaluate their practical grasping performance. Additionally, we conduct real-world experiments using our proposed model TARGO-Net and GIGA~\cite{giga} to validate their real-world applicability.

\subsection{Evaluation Metric}
\label{sec:eval_metric}

Following~\cite{giga,vgn}, we use grasp success rate (GSR) as the evaluation metric for target-driven grasping. If $m_{\mathrm{success}}$ of $m$ attempted grasps succeed, then
\begin{equation}
    \mathrm{GSR}=\frac{m_{\mathrm{success}}}{m}.
\end{equation}

%% file: sections/4_method.tex
\section{Tackling Visual Occlusion in Grasping: TARGO-Net}
\label{sec:targo}

\begin{figure}[ht]
    \centering
    \includegraphics[width=\textwidth]{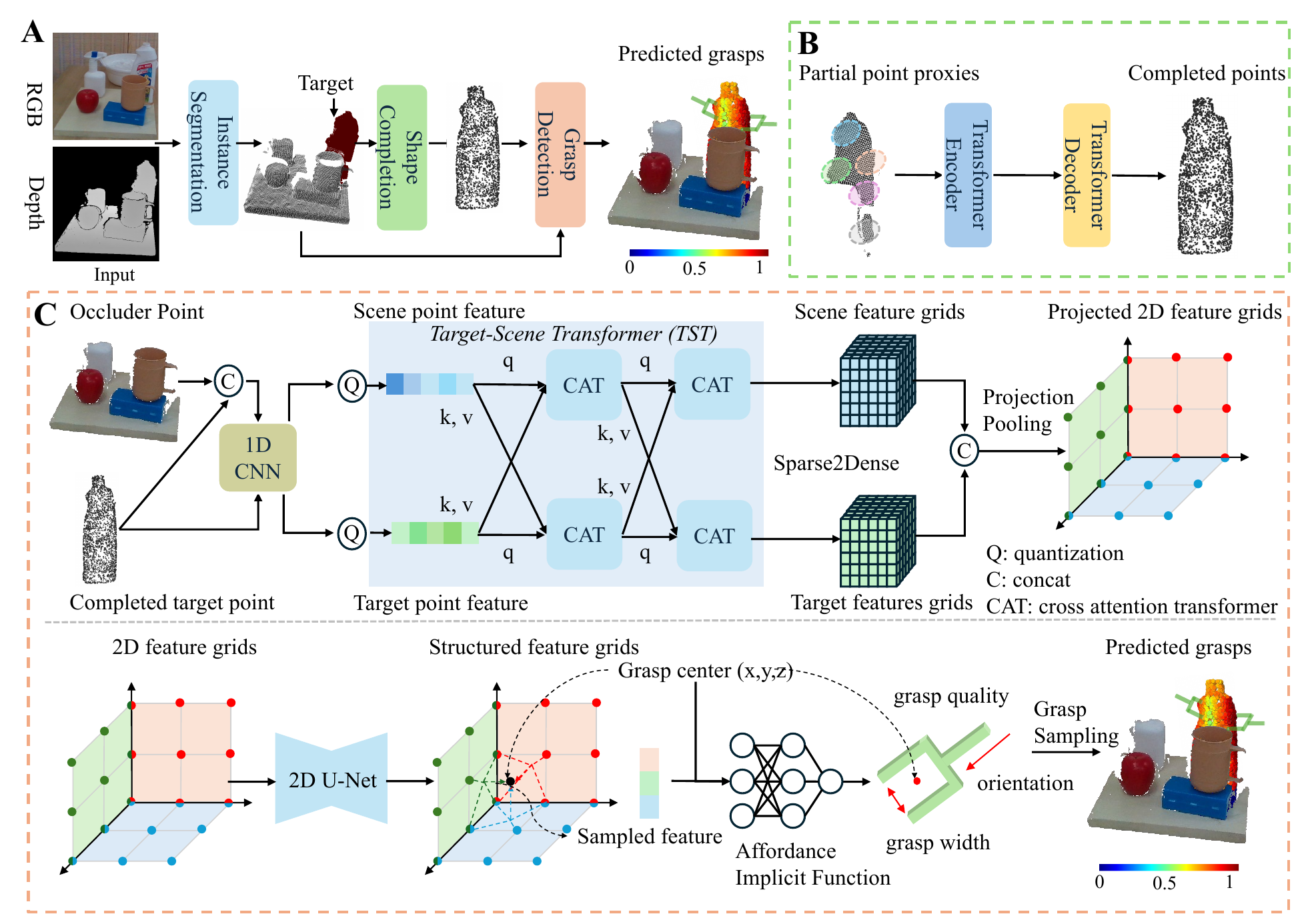}
    \caption{TARGO-Net: (A) overall pipeline, (B) target completion network, and (C) grasp detection module. The segmented partial target is completed to alleviate visual occlusion and then combined with the scene. The Target-Scene Transformer (TST) fuses sparse target and scene features to learn their occlusion relationship, and the implicit decoder predicts grasp quality, rotation, and width at each voxel. The voxel with the highest predicted quality yields the final 6-DoF grasp.}
    \label{fig:pipeline}
\end{figure}

\subsection{Problem Formulation}
\label{sec:problem_formulation}

Our setup comprises a tabletop workspace, a robotic arm with a parallel-jaw gripper, and a fixed RGB-D camera. We randomly place $K$ rigid objects upright in a packed arrangement. The camera observes the cluttered scene as $P\in\mathbb R^{\mathcal N_s\times3}$, where $\mathcal N_s$ is the total number of scene points. SAM~\cite{sam} segments the partial target $P_k\subset P$, and a pretrained completion network produces $P'_k\in\mathbb R^{\mathcal N_t\times3}$, where $\mathcal N_t$ is the number of completed target points.

To predict a 6-DoF pose that grasps the target while avoiding collisions with occluders, TARGO-Net learns, conditioned on the completed target and the full scene, the query function
\begin{equation}
    F_\theta(P'_k,P;\mathbf t)=(q,\mathbf R,w),
\end{equation}
where the query center $\mathbf t\in\mathbb R^3$ is mapped to grasp quality $q\in[0,1]$, rotation $\mathbf R\in\SO(3)$, and gripper width $w\geq0$. Jointly using $P'_k$ and $P$ lets the model reason about target geometry and surrounding occluders. At inference, we evaluate candidate centers near the target surface and execute the highest-quality retained grasp.

\subsection{Model Architecture}

\Cref{fig:pipeline} (A) shows our TARGO-Net pipeline.
Given a noisy RGB-D image, we first segment the target object from the scene.
Next, we feed the partial target point cloud into our target completion network, outputting a completed target point cloud.
The sparse feature encoder separately extracts target and scene features, which the Target-Scene Transformer (TST) fuses through bidirectional cross-attention. An implicit decoder then predicts grasp quality, orientation, and width, and the sampling module selects the 6-DoF grasp candidate with the highest predicted grasp quality.

\subsubsection{Instance Segmentation}

Given an RGB-D frame from the fixed RealSense D415 camera, we provide SAM~\cite{sam} with a visual prompt identifying the target and obtain its pixel mask. Back-projecting the depth values yields the partial target point cloud $P_k$ and the full scene point cloud $P$. We pass $P_k$ to the target completion network.

\subsubsection{Target Completion Network}
\label{sec:shape_completion}

Following AdaPoinTr~\cite{yu2023adapointr}, our completion network converts $P_k$ into point proxies that encode local regions (\Cref{fig:pipeline}(B)). A geometry-aware Transformer predicts point proxies for the unobserved regions, and a decoder reconstructs the completed target $P'_k$. No external completion dataset is required: for each partial observation in TARGO-Synthetic training, the simulator meshes and object poses provide the corresponding full-surface target point cloud.

We evaluate zero-shot generalization on the Fruits, Sports, and Toys categories of YCB~\cite{calli2015ycb}, all excluded from training. Across occlusion levels, TARGO-Net outperforms GIGA and VGN in both completion accuracy and downstream grasp success; detailed results are reported in~\Cref{sec:sc_generalization}.

\subsubsection{Grasp Detection Module}

TARGO-Net has a transformer-based architecture for robust 6-DoF grasp pose detection under heavy occlusion. The grasp detector jointly reasons over the completed target and its surrounding scene (\Cref{fig:pipeline}(C)). It comprises a sparse feature encoder, Target-Scene Transformer (TST), implicit function decoder, and grasp sampling module.

\paragraph{Sparse Feature Encoder}

Let the \emph{Complete Target} (CT) be $P'_k$ and the \emph{Raw Scene} (RS) be $P$. We combine CT with RS to form the \emph{Complete Target Scene} (CTS). A shared pointwise 1D CNN independently maps the points in CT and CTS to 32-dimensional descriptors, preserving point-level features before voxelization. We quantize each cloud on a $40\times40\times40$ grid and retain its non-empty voxels as a sparse tensor. A shared sparse 3D convolutional network, implemented with the Minkowski Engine~\cite{minkowski_engine}, then encodes the two tensors separately and places both streams in a common feature space. This reduces memory footprint and accelerates downstream processing while preserving the local geometry.

\paragraph{Target-Scene Transformer (TST)}
The TST contains $L=2$ stacked fusion layers~\cite{transformer,xia2023casspr}. Let $\mathbf T^{(0)}$ and $\mathbf S^{(0)}$ be the sparse features of CT and CTS, respectively. Each layer $\ell=1,\ldots,L$ first applies multi-head self-attention (SA) to the two streams independently:
\begin{equation}
  \bar{\mathbf T}^{(\ell)}
    = \mathrm{SA}_{t}^{(\ell)}\!\left(\mathbf T^{(\ell-1)}\right),
  \qquad
  \bar{\mathbf S}^{(\ell)}
    = \mathrm{SA}_{s}^{(\ell)}\!\left(\mathbf S^{(\ell-1)}\right).
\end{equation}
Bidirectional multi-head cross-attention (CA) then exchanges information between them:
\begin{equation}
  \begin{aligned}
    \hat{\mathbf T}^{(\ell)}
      &= \mathrm{CA}_{t}^{(\ell)}\!\left(
          \bar{\mathbf T}^{(\ell)},
          \bar{\mathbf S}^{(\ell)},
          \bar{\mathbf S}^{(\ell)}\right),\\
    \hat{\mathbf S}^{(\ell)}
      &= \mathrm{CA}_{s}^{(\ell)}\!\left(
          \bar{\mathbf S}^{(\ell)},
          \bar{\mathbf T}^{(\ell)},
          \bar{\mathbf T}^{(\ell)}\right),
  \end{aligned}
\end{equation}
where the three arguments to $\mathrm{CA}$ are the query, key, and value streams. The target queries the scene for occluder context, while the scene queries the target to emphasize target-relevant regions. Each attention head computes
\begin{equation}
  \mathrm{Attn}(\mathbf Q,\mathbf K,\mathbf V)
  = \mathrm{softmax}\!\left(\frac{\mathbf Q\mathbf K^\top}{\sqrt d}\right)\mathbf V,
\end{equation}
where $d$ is its feature dimension; learned position embeddings are added to the queries and keys. Residual and normalization operations follow each attention stage, and position-wise feed-forward blocks produce $\mathbf T^{(\ell)}$ and $\mathbf S^{(\ell)}$ for the next layer. Stacking $L$ fusion layers progressively refines these cues: the scene stream learns to highlight feasible target regions, while the target stream absorbs occluder geometry, enabling occlusion-aware reasoning about grasp constraints. The final streams are passed to the implicit decoder. Our ablation over $L\in\{1,2,3\}$ shows that $L=2$ yields the highest GSR, offering the best balance between representational power and trainability (\Cref{sec:number_ca_layers}).

\paragraph{Implicit Function Decoder}

After the TST, we convert the target and scene streams into separate dense $40\times40\times40$ grids with 32 channels. A shared 3D CNN processes each grid, after which we concatenate the two volumes along the channel dimension~\cite{convonet, giga}.

To reduce memory and computation, we project the concatenated volume onto three orthogonal planes, $xy$, $yz$, and $xz$. For each plane, average pooling aggregates the voxels mapping to each of its $40\times40$ cells, and a lightweight 2D U-Net integrates local detail with global context to produce a refined feature map with $C=64$ channels.

For a query center $\mathbf p=(x,y,z)$, we bilinearly sample the three plane features and concatenate them:
\begin{equation}
  \boldsymbol\phi(\mathbf p) = \bigl[
    \phi_{xy}(x,y),\;\phi_{yz}(y,z),\;\phi_{xz}(x,z)
  \bigr]\in\mathbb R^{3C}.
\end{equation}

A lightweight MLP decodes $\bigl(\mathbf p,\boldsymbol\phi(\mathbf p)\bigr)$ into grasp quality $q\in[0,1]$, a unit quaternion $\mathbf r\in\mathbb S^3\subset\mathbb R^4$ representing $\mathbf R$, and gripper width $w\in[0,0.08]$:
\begin{equation}
  f_\theta\colon (\mathbf p,\boldsymbol\phi)\;\mapsto\;
    \bigl(q,\;\mathbf r,\;w\bigr).
\end{equation}
This implicit mapping is fully differentiable, preserves fine-grained geometry, and can be queried efficiently at arbitrary centers.

\subsubsection{Grasp Sampling Module}

Following VGN~\cite{vgn}, we discretize the cubic workspace of side length $0.3\,\mathrm m$ into a $40 \times 40 \times40$ grid, treat every voxel center as a potential grasp center, and query the decoder for its quality, rotation, and width. We smooth the predicted quality grid with a 3D Gaussian kernel to suppress noise and emphasize the near-surface graspable region, discard candidates below a fixed quality threshold, and apply non-maximum suppression to remove nearby lower-quality grasps. The retained candidate with the highest quality is selected for execution.

\subsection{Loss Functions}

Let $g=(q,\mathbf r,w)$ and $\hat g=(\hat q,\hat{\mathbf r},\hat w)$ denote the ground-truth and predicted parameter tuples; hats denote predicted quantities, and $\mathbf r,\hat{\mathbf r}\in\mathbb S^3$ are the corresponding quaternion representations of the rotation. Following VGN~\cite{vgn}, we train on TARGO-Synthetic with
\begin{equation}
  \mathcal{L}(\hat{g}, g)
  = \mathcal{L}_q(\hat{q}, q)
  \;+\;
  q \,\bigl(\mathcal{L}_r(\hat{\mathbf{r}}, \mathbf{r})
  + \mathcal{L}_w(\hat{w}, w)\bigr),
\end{equation}
where $\mathcal L_q$ is binary cross-entropy and $\mathcal L_w$ is mean-squared error. We measure rotation error by the quaternion distance $\mathcal{L}_{\mathrm{quat}}(\hat{\mathbf{r}}, \mathbf{r})=1-|\hat{\mathbf r}\cdot\mathbf r|$~\cite{kuffner2004distance}. A parallel-jaw grasp is unchanged by a $180^\circ$ rotation about the gripper's wrist axis, so both $\mathbf r$ and its $\pi$-rotated equivalent $\mathbf r_\pi$ are valid. We therefore define
\begin{equation}
  \mathcal{L}_r(\hat{\mathbf{r}}, \mathbf{r})
  = \min\!\bigl(
    \mathcal{L}_{\mathrm{quat}}(\hat{\mathbf{r}}, \mathbf{r}),\;
    \mathcal{L}_{\mathrm{quat}}(\hat{\mathbf{r}}, \mathbf{r}_\pi)
  \bigr).
\end{equation}
The factor $q$ applies the rotation and width losses only to positive grasp labels ($q=1$).

%% file: sections/5_exp.tex
\section{Experiments}
\label{sec:exp}

In the following, we present the details of training and inference, evaluation on three test sets, and ablation studies regarding our dataset TARGO and model TARGO-Net.

\subsection{Training Setup}
\label{sec:training_setup}

We train TARGO-Net end-to-end for 15 epochs using the Adam optimizer~\cite{kingma2014adam} with an initial learning rate of $2\times10^{-4}$ and a batch size of 32. On an NVIDIA A40 GPU, this offline training takes approximately five days.

\subsection{Inference Efficiency}

\begin{table}[hbtp]
    \centering
    \caption{Model size and runtime comparison. Inference time covers model inference, while latency also includes grasp post-processing.}
    \label{tab:num_paras}
    \begin{tabular}{lccc}
        \toprule
        Model & Params (M) & Inference (ms) & Latency (ms) \\
        \midrule
        ICGNet~\cite{zurbrugg2024icgnet}         &  3.2 & 137 & 138 \\
        VN-EdgeGraspNet~\cite{huang2023edge}     &  \textbf{1.7} & 264 & 306 \\
        \textbf{TARGO-Net (Ours)}                & 33.34 & \textbf{63.4} & \textbf{127} \\
        \bottomrule
    \end{tabular}
\end{table}

TARGO-Net comprises \SI{33.34}{M} parameters (\SI{0.89}{M} for grasp detection and \SI{32.45}{M} for target completion), compared to \SI{3.2}{M} in ICGNet~\cite{zurbrugg2024icgnet}.
On an NVIDIA Quadro P5000 GPU, TARGO-Net has an inference time of \SI{63.4}{ms} (\SI{14.6}{ms} for target completion and \SI{48.8}{ms} for grasp detection\footnote{The TST accounts for \SI{5.6}{ms} of the grasp-detection time.}), less than half of ICGNet's \SI{137}{ms}. Its end-to-end latency, including grasp post-processing, is \SI{127}{ms}, corresponding to approximately $7.9$ predictions per second and remaining below ICGNet's \SI{138}{ms}.

\subsection{Scalability Analysis of TARGO-Net}

To evaluate how training dataset size affects model performance and training time, we trained TARGO-Net on smaller datasets comprising $\frac{1}{10}$ and $\frac{1}{5}$ of the full TARGO-Synthetic dataset. Training on the $\frac{1}{10}$ subset took approximately 12 hours, while the $\frac{1}{5}$ subset required around 24 hours on an NVIDIA A40 GPU.

\Cref{tab:scalability_analysis} reports grasp success rate (GSR) on the TARGO-Synthetic test set under different occlusion levels. For every dataset size, the drop from the low- to the high-occlusion range is at most $4\%$. Averaged over the three ranges, training on the $\frac{1}{5}$ and $\frac{1}{10}$ subsets reduces GSR by $5.58\%$ and $10.47\%$, respectively, relative to training on the full dataset. More training data therefore raises the overall GSR level, but the degradation across occlusion ranges is similar for all three dataset sizes.

\begin{table}[htbp]
    \centering
    \caption{Scalability of TARGO-Net: GSR (\%) across occlusion bins. \textbf{Decline} is the absolute drop (\%) from the low to high visual occlusion ranges.}
    \label{tab:scalability_analysis}
    \begin{tabular}{lccccc}
        \toprule
        \multirow{2}{*}{\textbf{Method}}
            & \multicolumn{3}{c}{\textbf{Visual Occlusion Level ($\OL$)}}
            & \multirow{2}{*}{\textbf{Decline}}
            & \multirow{2}{*}{\textbf{Average}}\\
        \cmidrule(lr){2-4}
            & $[0,0.3)$ & $[0.3,0.6)$ & $[0.6,0.9)$ & & \\
        \midrule
        TARGO-Net (full)              & \textbf{83.93} & \textbf{81.04} & \textbf{80.07} & 3.86 & \textbf{81.68}\\
        \addlinespace[2pt]
        TARGO-Net ($\tfrac15$ data)   & 77.98 & 75.63 & 74.68 & 3.30 & 76.10\\
        \addlinespace[2pt]
        TARGO-Net ($\tfrac1{10}$ data)& 73.72 & 70.19 & 69.73 & 3.99 & 71.21\\
        \bottomrule
    \end{tabular}
\end{table}

\subsection{Evaluation Results}

\subsubsection{Evaluation on TARGO-Synthetic Test Set}
\label{sec:target_driven_res}
\begin{figure*}[hbtp]
  \centering
  \begin{minipage}[b]{0.51\linewidth}
    \centering
    \includegraphics[width=\linewidth]{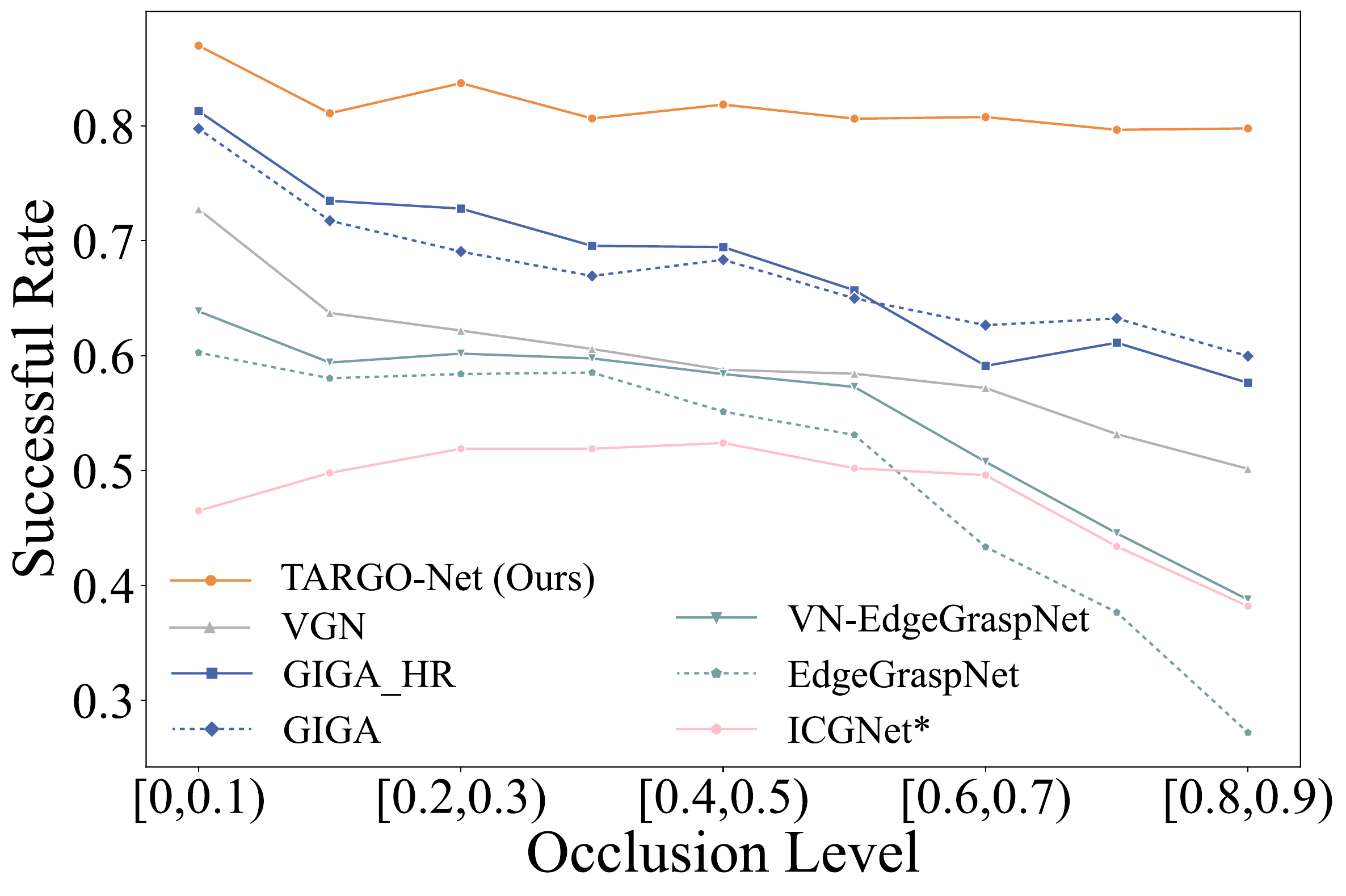}
    \captionof{figure}{GSR of TARGO-Net and the baselines across visual occlusion levels on the TARGO-Synthetic test set.}
    \label{fig:abl_baseline_comparison}
  \end{minipage}
  \hfill
  \begin{minipage}[b]{0.46\linewidth}
    \centering
    \includegraphics[width=\linewidth]{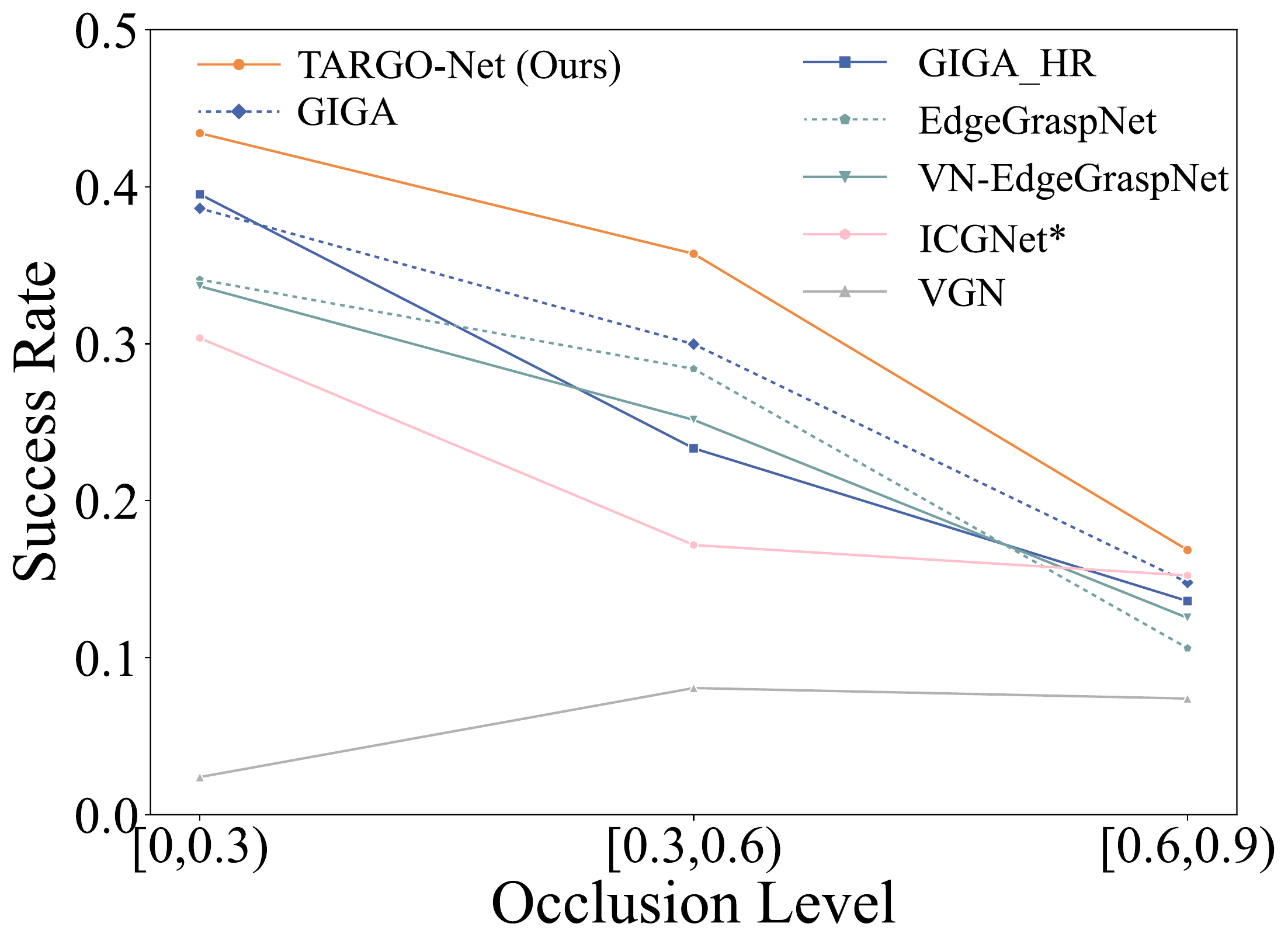}
    \captionof{figure}{GSR of TARGO-Net and the baselines across visual occlusion ranges on TARGO-Real.}
    \label{fig:comparisons_rw}
  \end{minipage}
\end{figure*}

In~\Cref{fig:abl_baseline_comparison}, we compare TARGO-Net with VGN~\cite{vgn}, GIGA and GIGA-HR~\cite{giga}, EdgeGraspNet and VN-EdgeGraspNet~\cite{huang2023edge}, and ICGNet~\cite{zurbrugg2024icgnet} on the TARGO-Synthetic test set.\footnote{For a fair comparison, we exclude target-driven methods that remove occluders before grasping or plan the next best view~\cite{murali2020, liu2022ge, yang2020, breyer2022nbvp, zhang2023nbvp}.}
Because these baselines were originally designed for clutter removal, we adapt them to target-driven grasping. For VGN, GIGA, and GIGA-HR, we restrict grasp queries to the target region after predicting affordances for the entire scene. For EdgeGraspNet and VN-EdgeGraspNet, approach points are sampled only from the target. However, their grasps are inferred from local patches without global scene awareness and may therefore collide with occluders. ICGNet is adapted by selecting its highest-scoring grasp on the target object.

All baselines achieve a lower grasp success rate (GSR) in the $[0.8,0.9)$ occlusion bin than in the $[0,0.1)$ bin. ICGNet has the lowest average GSR across the evaluated bins and is up to $47\%$ below TARGO-Net under severe occlusion. EdgeGraspNet exhibits the largest endpoint decline, approximately $30\%$, while VGN, GIGA, and GIGA-HR decline by roughly $20$--$25\%$. These results reflect the difficulty of predicting collision-free target grasps when global scene context is unavailable or the target geometry is reconstructed inaccurately.

In contrast, TARGO-Net has the smallest endpoint decline, approximately $7\%$, and achieves the highest GSR throughout the tested occlusion range. Its grasp detector jointly processes the scene and completed target point clouds, enabling it to exploit both occluder context and target geometry when predicting collision-free grasps. Overall, the limited robustness of existing methods under severe visual occlusion highlights the importance of explicitly accounting for occlusion in future robotic grasping research.

\subsubsection{Evaluation on Real-world Experiments}
\label{sec:real_world_exp}

We compare TARGO-Net with GIGA in real-world target-driven grasping using a Franka Emika Panda arm. For each method, we conduct at least 10 rounds in packed scenes within each of three visual occlusion ranges: \textbf{Low} $[0,0.3)$, \textbf{Medium} $[0.3,0.6)$, and \textbf{High} $[0.6,0.9)$. In each round, at least five objects are randomly selected from a pool of 18 test objects and placed on the table, and one is randomly designated as the target. For each trial, the model receives a back-projected partial point cloud from a side-view depth image, and the robot executes the highest-scoring predicted grasp. A trial is successful if the robot lifts the target and places it in a nearby container without colliding with other objects.

Before executing a grasp, the robot moves to a pre-grasp pose offset by \SI{5}{cm} from the target along the approach direction, reducing the risk of premature contact. MoveIt~\cite{moveit} uses the observed obstacle point cloud and the robot's current joint state to plan collision-aware motions for grasping and placement.

As reported in~\Cref{tab:real_world_exp}, TARGO-Net achieves observed GSRs of $80.0\%$, $75.0\%$, and $66.7\%$ in the low-, medium-, and high-occlusion ranges, respectively, compared with $70.0\%$, $40.0\%$, and $30.0\%$ for GIGA.\footnote{Videos of our target-driven grasping experiments are available at \url{https://targo-benchmark.github.io}.} TARGO-Net leads in every range, with the margin widening from $10.0\%$ to $36.7\%$ as occlusion increases---largely due to TST and single-scene grasp augmentation.

TARGO-Net nevertheless degrades more steeply in the real world than in simulation: $13.3\%$ from the low to the high range, compared with approximately $7\%$ over the wider bin range evaluated in simulation (\Cref{fig:abl_baseline_comparison}). This may reflect domain gaps such as sensor noise and occlusion artifacts, which pose additional challenges for target completion.

\begin{figure}[!htbp]
  \centering
    \includegraphics[width=0.8\linewidth]{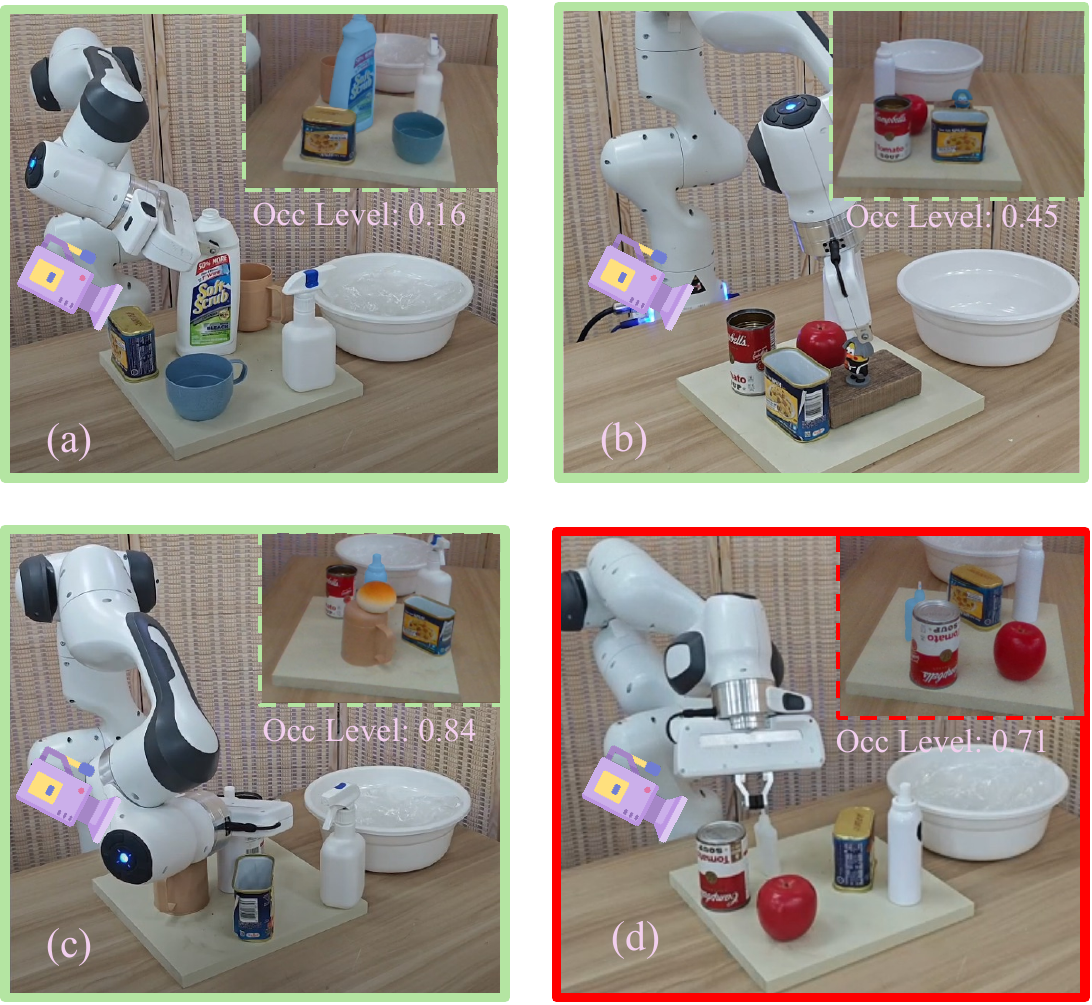}
    \caption{Real-world grasping performance under increasing visual occlusion. Scenes (a)--(c) show successful grasps in the low, medium, and high occlusion ranges, respectively; Scene (d) shows a failed grasp on the glue bottle under high occlusion. The upper-right inset in each scene shows the RGB-D camera view.}
    \label{fig:demos_rw}
\end{figure}

\Cref{fig:demos_rw} shows successful grasps at low ($0.162$), medium ($0.452$), and high ($0.845$) occlusion in Scenes (a)--(c), respectively, and a failure under high occlusion in Scene (d). Scene (b) uses a figurine from the unseen Toys category, providing a qualitative example of transfer to an unseen category. The names, masses, and dimensions of all 18 real-world test objects are provided in~\Cref{sec:details_objects_real}.

\subsubsection{Evaluation on TARGO-Real Dataset}
\label{sec:targo_real}

We evaluate TARGO-Net and six baselines on TARGO-Real: GIGA and GIGA-HR~\cite{giga}, VGN~\cite{vgn}, EdgeGraspNet and VN-EdgeGraspNet~\cite{huang2023edge}, and ICGNet~\cite{zurbrugg2024icgnet}.
All methods are trained on the TARGO-Synthetic training set. As shown in~\Cref{fig:comparisons_rw}, every method has a GSR below $50\%$, but TARGO-Net has the highest GSR in each occlusion range. Averaged across the occlusion ranges, TARGO-Net outperforms GIGA, GIGA-HR, EdgeGraspNet, VN-EdgeGraspNet, VGN, and ICGNet by $4.21\%$, $6.51\%$, $7.63\%$, $8.22\%$, $26.04\%$, and $12.04\%$, respectively.

The lower absolute GSRs may reflect differences in object geometry, scene arrangement, and depth sensing between TARGO-Real and the synthetic training data. Further, the evaluation workspace is also larger ($\SI{0.7}{m}\times\SI{0.7}{m}\times\SI{0.7}{m}$ rather than $\SI{0.3}{m}\times\SI{0.3}{m}\times\SI{0.3}{m}$).

\subsection{Ablation Studies}
\label{sec:ablation_study}

\begin{figure*}[hbtp]
  \centering
  \begin{minipage}[b]{0.48\linewidth}
    \centering
    \includegraphics[width=\linewidth]{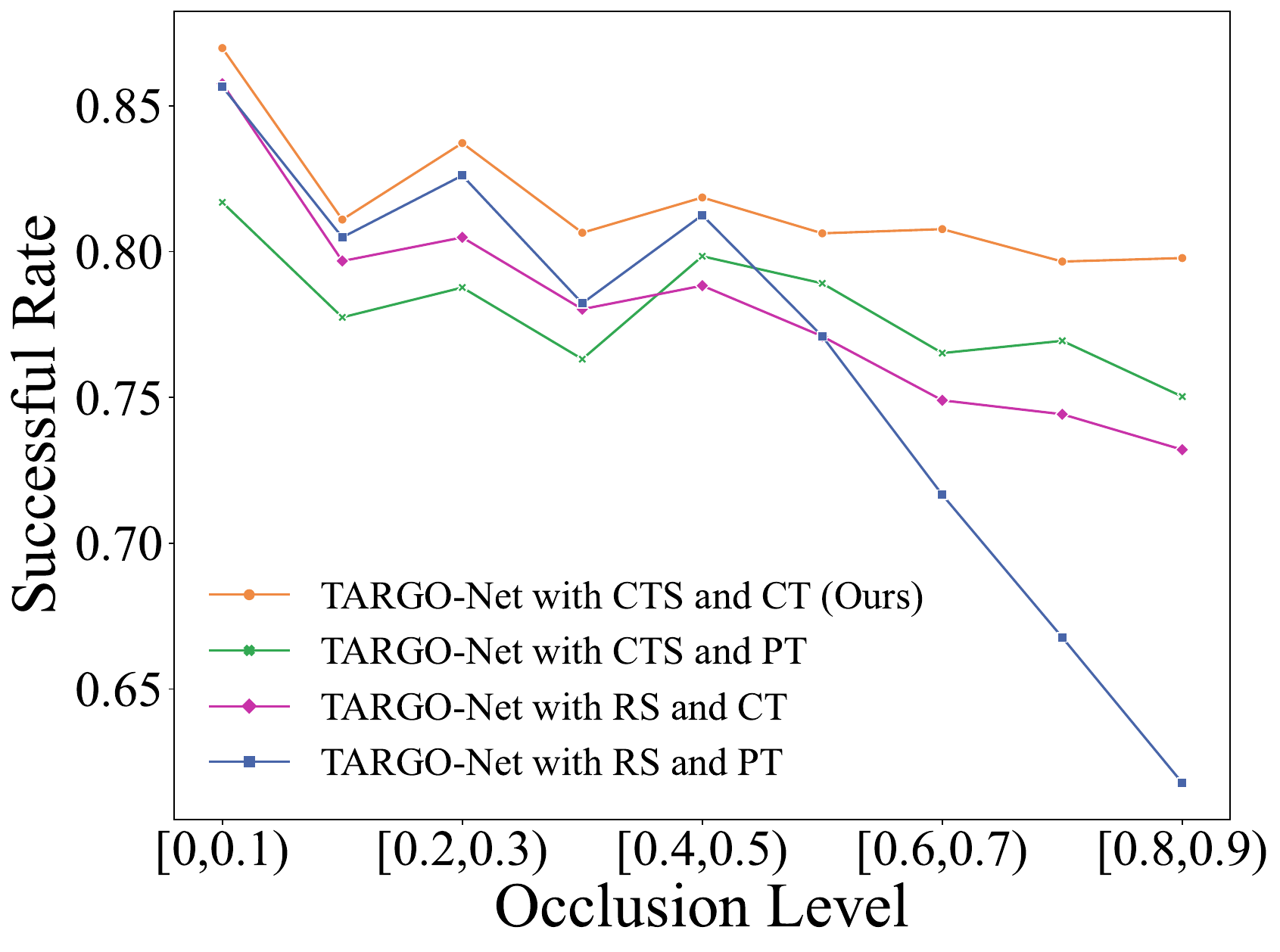}
    \captionof{figure}{GSR of TARGO-Net variants with different scene and target input completeness on the TARGO-Synthetic test set.}
    \label{fig:abl_target}
  \end{minipage}
  \hfill
  \begin{minipage}[b]{0.48\linewidth}
    \centering
    \includegraphics[width=\linewidth]{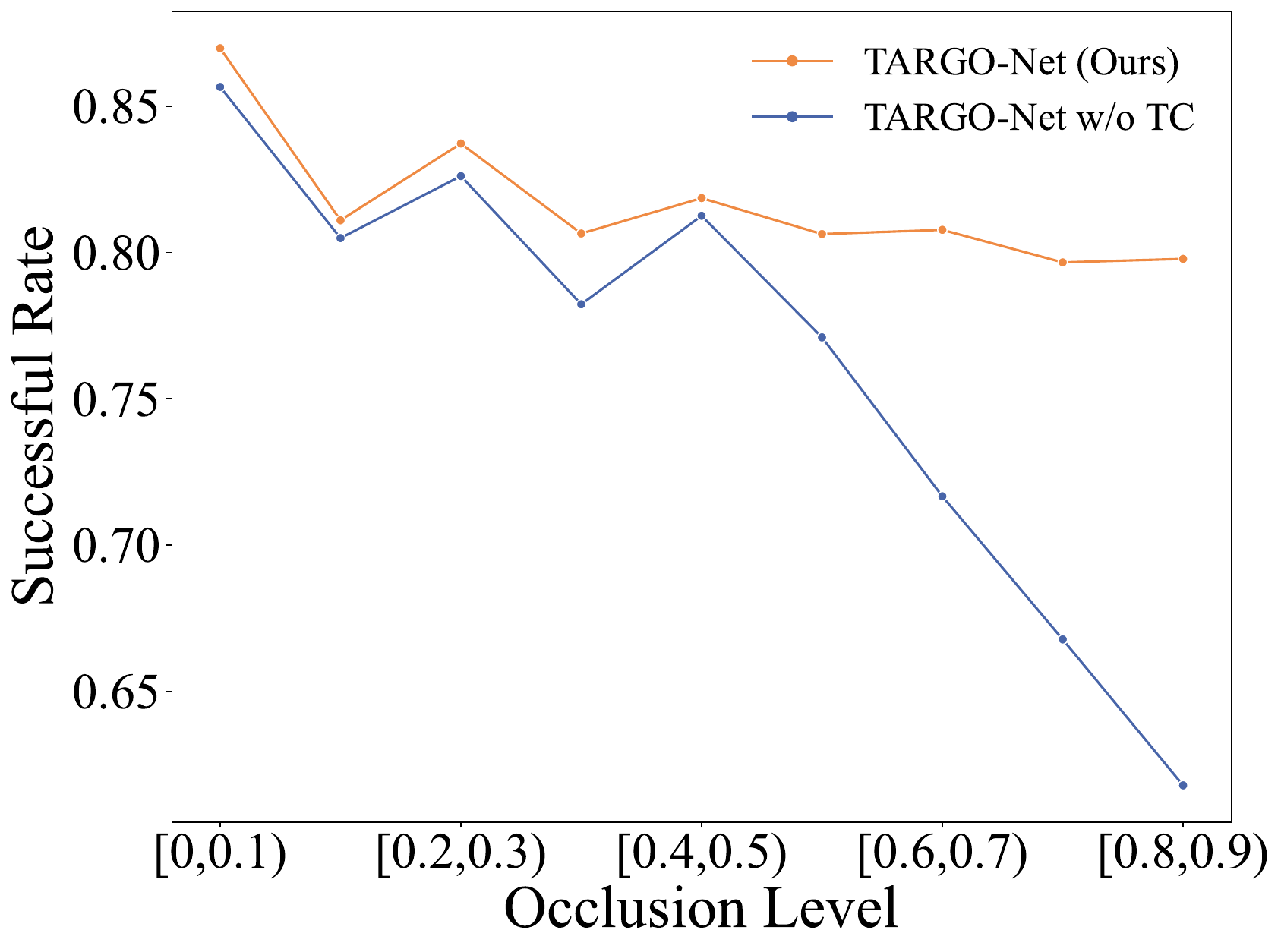}
    \captionof{figure}{GSR with and without target completion on the TARGO-Synthetic test set.}
    \label{fig:abl_sc}
  \end{minipage}
\end{figure*}

\subsubsection{Single-Scene Grasp Augmentation}
\label{sec:single_scene_analysis}
\begin{figure*}[htbp]
  \centering
  \begin{minipage}[b]{0.48\linewidth}
    \centering
    \includegraphics[width=\linewidth]{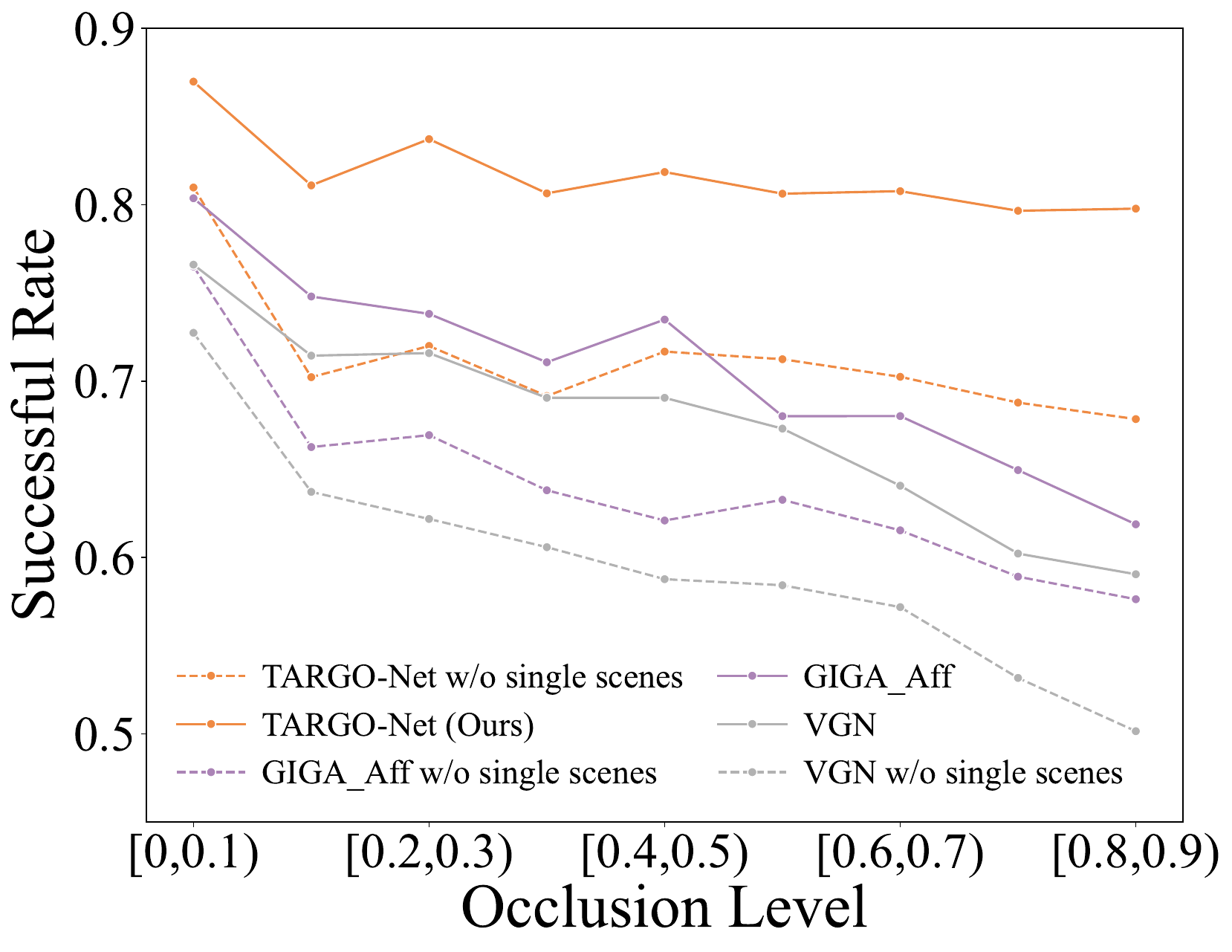}
    \captionof{figure}{GSR with and without single-scene grasp augmentation on the TARGO-Synthetic test set.}
    \label{fig:abl_data_aug}
  \end{minipage}
  \hfill
  \begin{minipage}[b]{0.48\linewidth}
    \centering
    \includegraphics[width=\linewidth]{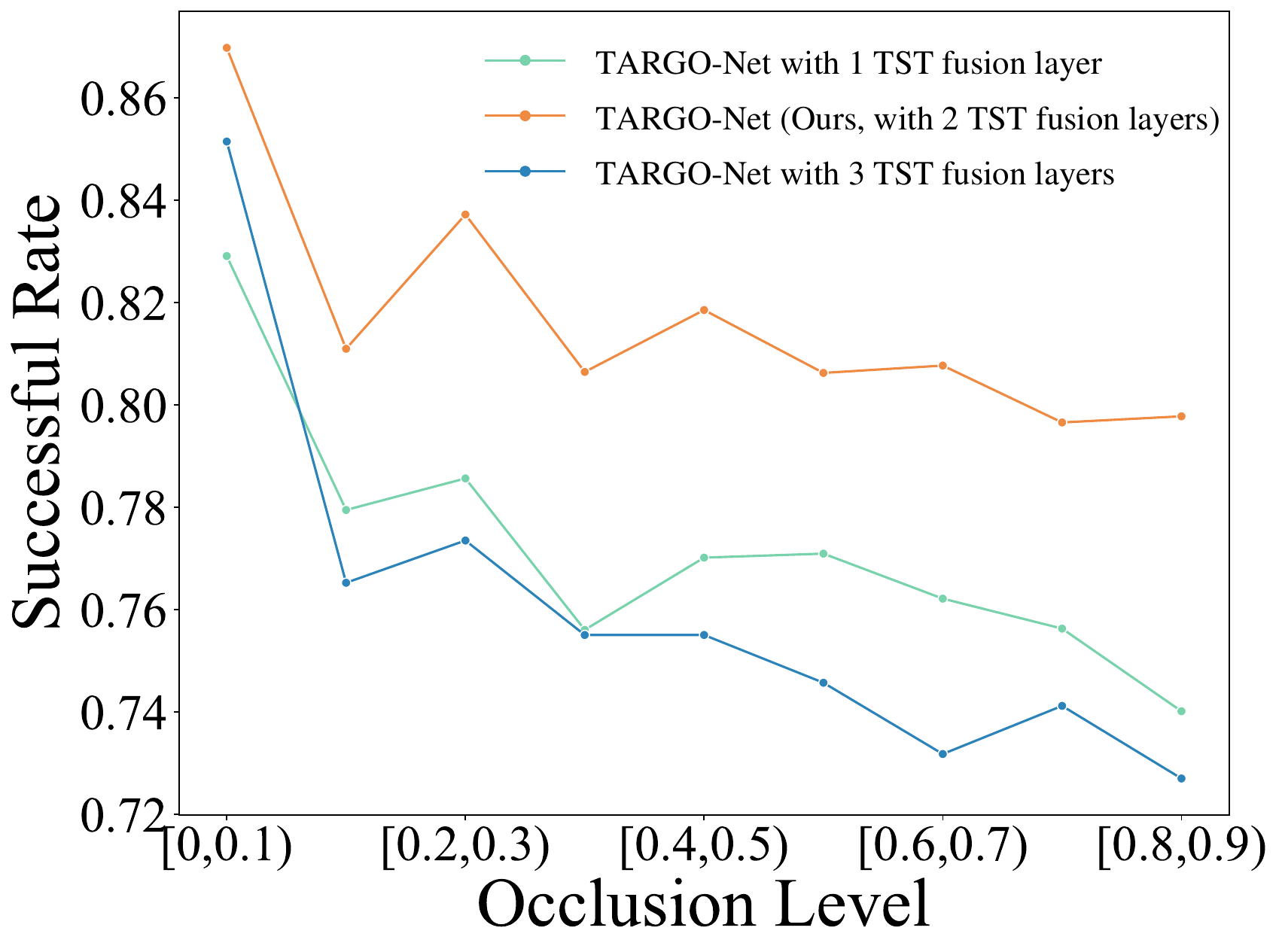}
    \captionof{figure}{GSR for different numbers $L$ of TST fusion layers. The figure legend abbreviates cross-attention as CA.}
    \label{fig:abl_transformer_layer}
  \end{minipage}
\end{figure*}

We compare TARGO-Net, GIGA-Aff, and VGN trained with and without single-scene grasp augmentation to isolate its impact (see~\Cref{sec:existence_occluders} for motivation). GIGA-Aff is the affordance-prediction branch of GIGA trained without its scene-completion branch. As shown in~\Cref{fig:abl_data_aug}, single-scene grasp augmentation increases average GSR by $10.2\%$, $5.6\%$, and $7.9\%$ for TARGO-Net, GIGA-Aff, and VGN, respectively.

\begin{table*}[htbp]
    \centering
    \begin{minipage}[t]{0.53\textwidth}
        \centering
        \caption{Target completion accuracy measured by CD-$\ell_1$ ($\times 1000$, lower is better) and volumetric IoU (\%, higher is better).}
        \label{tab:target_completion}
        \vspace{0.2cm}
        \setlength{\tabcolsep}{4pt}
        \resizebox{\textwidth}{!}{
        \begin{tabular}{lcc}
            \toprule
            Method & CD-$\ell_1$ \(\downarrow\) & IoU (\%) \(\uparrow\) \\
            \midrule
            GIGA~\cite{giga}          & 8.722 & 78.77 \\
            GIGA-Geo~\cite{giga}      & 5.588 & 85.67 \\
            \textbf{TARGO-Net (ours)} & \textbf{2.443} & \textbf{93.99} \\
            \bottomrule
        \end{tabular}}
    \end{minipage}%
    \hfill
    \begin{minipage}[t]{0.45\textwidth}
        \centering
        \caption{GSR (\%) of GIGA and TARGO-Net over three visual occlusion ranges in the real-world experiments.}
        \label{tab:real_world_exp}
        \setlength{\tabcolsep}{4pt}
        \resizebox{\textwidth}{!}{
        \begin{tabular}{lccc}
            \toprule
            Method & \makecell{Low \\ $[0,0.3)$}
                   & \makecell{Medium \\ $[0.3,0.6)$}
                   & \makecell{High \\ $[0.6,0.9)$} \\
            \midrule
            GIGA~\cite{giga}          & 70.0 & 40.0 & 30.0 \\
            \textbf{TARGO-Net (ours)} & 80.0 & 75.0 & 66.7 \\
            \bottomrule
        \end{tabular}}
    \end{minipage}
\end{table*}

\begin{figure}[htbp]
    \centering
    \includegraphics[width=\linewidth]{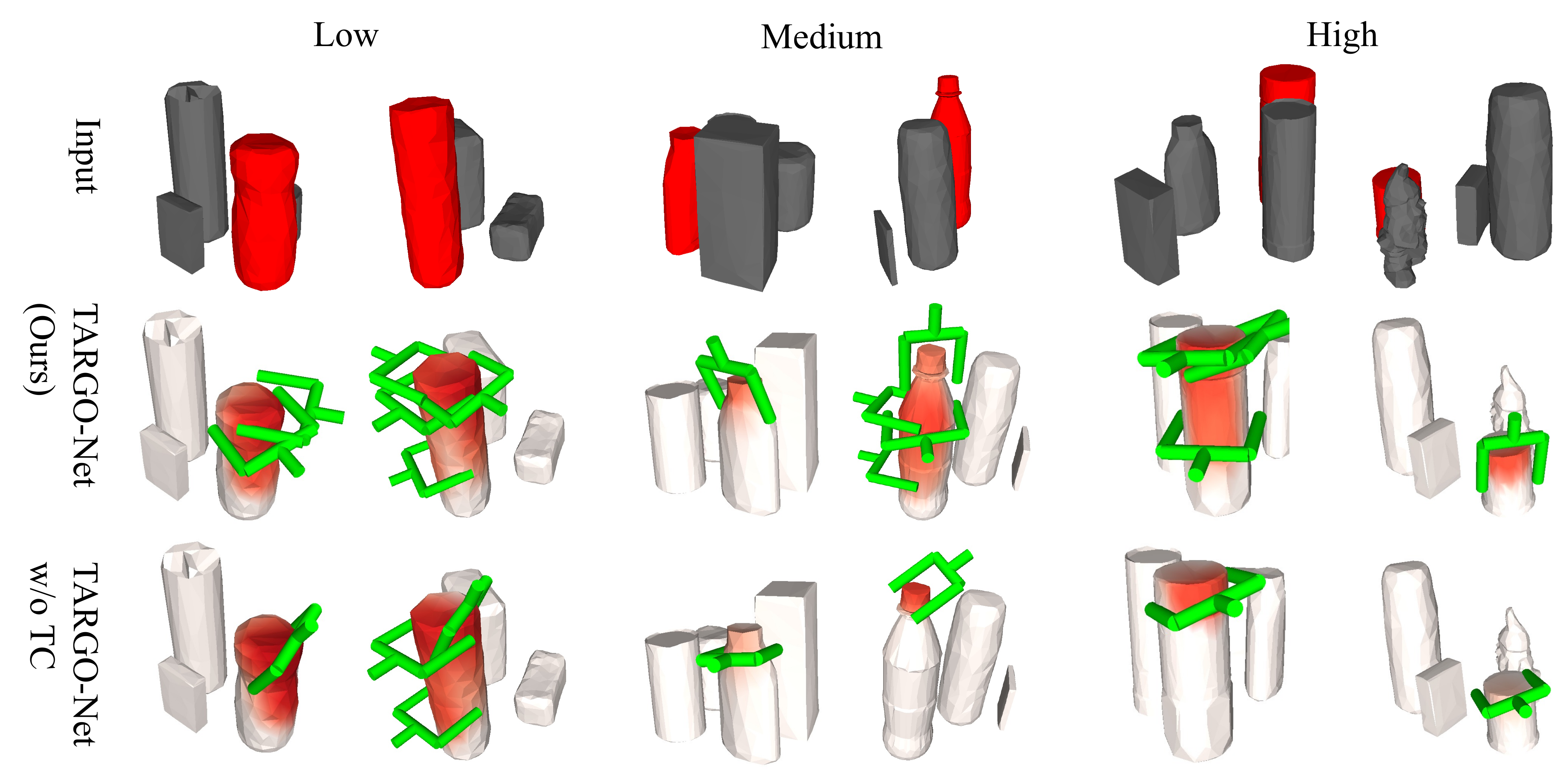}
    \caption{Effects of target completion on the TARGO-Synthetic test set. \emph{TARGO-Net w/o TC} removes the target completion module in TARGO-Net. The red target in the input view indicates the target, while for model predictions in the second and third rows, the red region represents the grasp affordance predictions.}
    \label{fig:sc_vs_no_sc}
\end{figure}

\subsubsection{Impact of Input Completeness}
We analyze how scene and target input completeness affects four TARGO-Net variants, all trained on the TARGO-Synthetic training set. The target completion module maps the \emph{Partial Target (PT)} to the \emph{Complete Target (CT)}; placing CT back into the \emph{Raw Scene (RS)} forms the \emph{Complete Target Scene (CTS)}. The four input pairs are \emph{CTS and CT} (ours), \emph{CTS and PT}, \emph{RS and CT}, and \emph{RS and PT}.

As shown in~\Cref{fig:abl_target}, \emph{TARGO-Net with CTS and CT} has the highest GSR across all occlusion levels. From the first to the last bin, \emph{RS and PT} declines by approximately $24\%$, compared with about $7\%$ for \emph{CTS and PT} and $13\%$ for \emph{RS and CT}. With both inputs complete, TARGO-Net retains approximately $80\%$ GSR in the highest-occlusion bin.

\subsubsection{Impact of Target Completion}
The preceding results show that input completeness is associated with robustness under visual occlusion. To isolate target completion, we compare the full \emph{TARGO-Net} with \emph{TARGO-Net w/o TC}, which is equivalent to the \emph{RS and PT} variant (\Cref{fig:abl_sc}).
At low visual occlusion ($0$--$30\%$), the two variants perform similarly. Beyond $40\%$ occlusion, however, \emph{TARGO-Net w/o TC} is lower by as much as $18\%$, while the full model degrades more gradually.

Besides quantitative analysis, we qualitatively compare the grasp and affordance predictions between our \emph{TARGO-Net} and \emph{TARGO-Net w/o TC}, which removes the target completion module. We categorize the comparisons into three occlusion ranges, each containing two examples: Low $[0,0.3)$, Medium $[0.3,0.6)$, and High $[0.6,0.9)$.
As illustrated in~\Cref{fig:sc_vs_no_sc}, the affordance predictions of \emph{TARGO-Net w/o TC} are concentrated on geometry visible in the input. With target completion, \emph{TARGO-Net} also predicts affordance in reconstructed regions and identifies candidate grasps on the far side of the target in the medium- and high-occlusion examples. These examples are consistent with the quantitative benefit of target completion under heavier occlusion.

\subsubsection{Target Completion vs. Scene Completion}

GIGA (with scene completion and affordance prediction) and GIGA-Geo (trained for scene completion without the affordance branch) reconstruct the full \textbf{scene} using Convolutional Occupancy Networks~\cite{convonet}, whereas TARGO-Net completes the partial \textbf{target} with a geometry-aware Transformer. For GIGA and GIGA-Geo, we evaluate the target component extracted from the reconstructed scene. \Cref{tab:target_completion} shows that TARGO-Net has lower CD-$\ell_1$ and higher volumetric IoU than both scene-completion baselines. This improvement coincides with higher downstream grasp performance.

\begin{figure}[htbp]
    \centering
    \includegraphics[width=0.8 \linewidth]{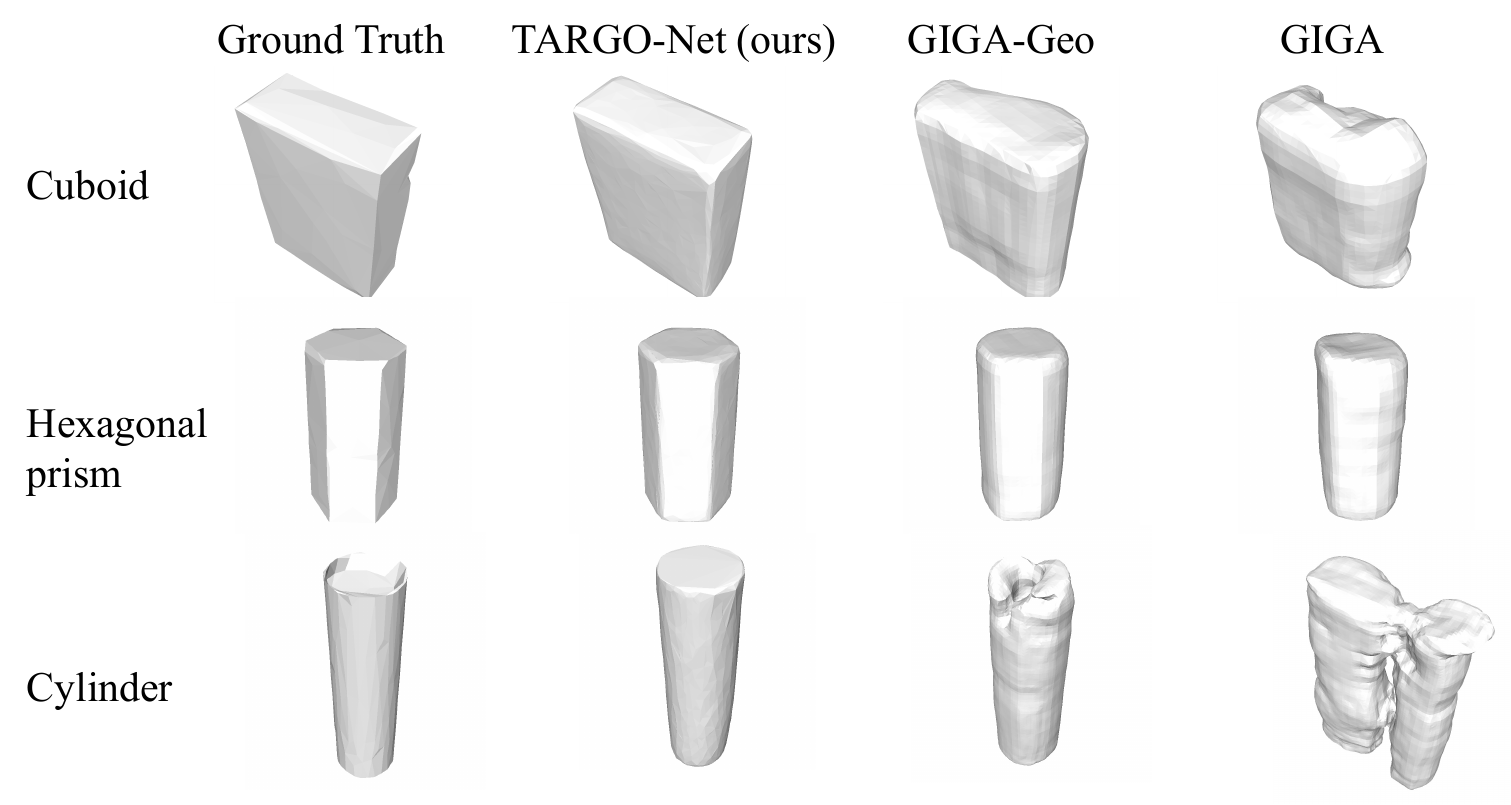}
    \caption{Qualitative target completion comparisons between TARGO-Net, GIGA-Geo, and GIGA.}
    \label{fig:sc_comp}
\end{figure}

In~\Cref{fig:sc_comp}, we \emph{qualitatively} compare the target completion performance of TARGO-Net with GIGA-Geo and GIGA.
TARGO-Net shows better completion performance than GIGA-Geo and GIGA, which is consistent with the quantitative results in~\Cref{tab:target_completion}.
For the completion of cuboid and hexagonal prism targets, TARGO-Net achieves near-perfect results, whereas GIGA-Geo and GIGA hardly preserve the shape.
In the case of cylinder reconstruction, TARGO-Net loses some details on the top but to a much lesser extent than GIGA-Geo.
GIGA, however, predicts a target shape connected with other occluders since it takes all scene voxels as input and occasionally fails to separate occluders near the target.
These findings suggest that the target completion approach used by our TARGO-Net is more effective than scene completion approaches for the target-driven grasping task.

\subsubsection{Effects of Segmentation Accuracy on Grasping}
We compare ground-truth (GT) target masks with masks predicted by SAM to assess sensitivity to segmentation errors. As shown in~\Cref{tab:segmentation}, reducing the reported segmentation accuracy from $100\%$ to $95.7\%$ coincides with a GSR decrease of $4.4\%$. This two-condition comparison quantifies the observed sensitivity but does not establish a general correlation.

\begin{table}[t]
    \centering
    \caption{Target segmentation accuracy and downstream GSR for ground-truth masks and SAM predictions.}
    \label{tab:segmentation}
    \begin{tabular}{lcc}
        \toprule
        \textbf{Target Mask} & \textbf{Segmentation Accuracy (\%)} & \textbf{GSR (\%)} \\
        \midrule
        Ground truth & 100.0 & 81.7 \\
        SAM & 95.7 & 77.3 \\
        \bottomrule
    \end{tabular}
\end{table}

\subsubsection{Effects of Gaussian Noise on Grasping}
\begin{figure}[t]
    \centering
    \includegraphics[width=0.8\linewidth]{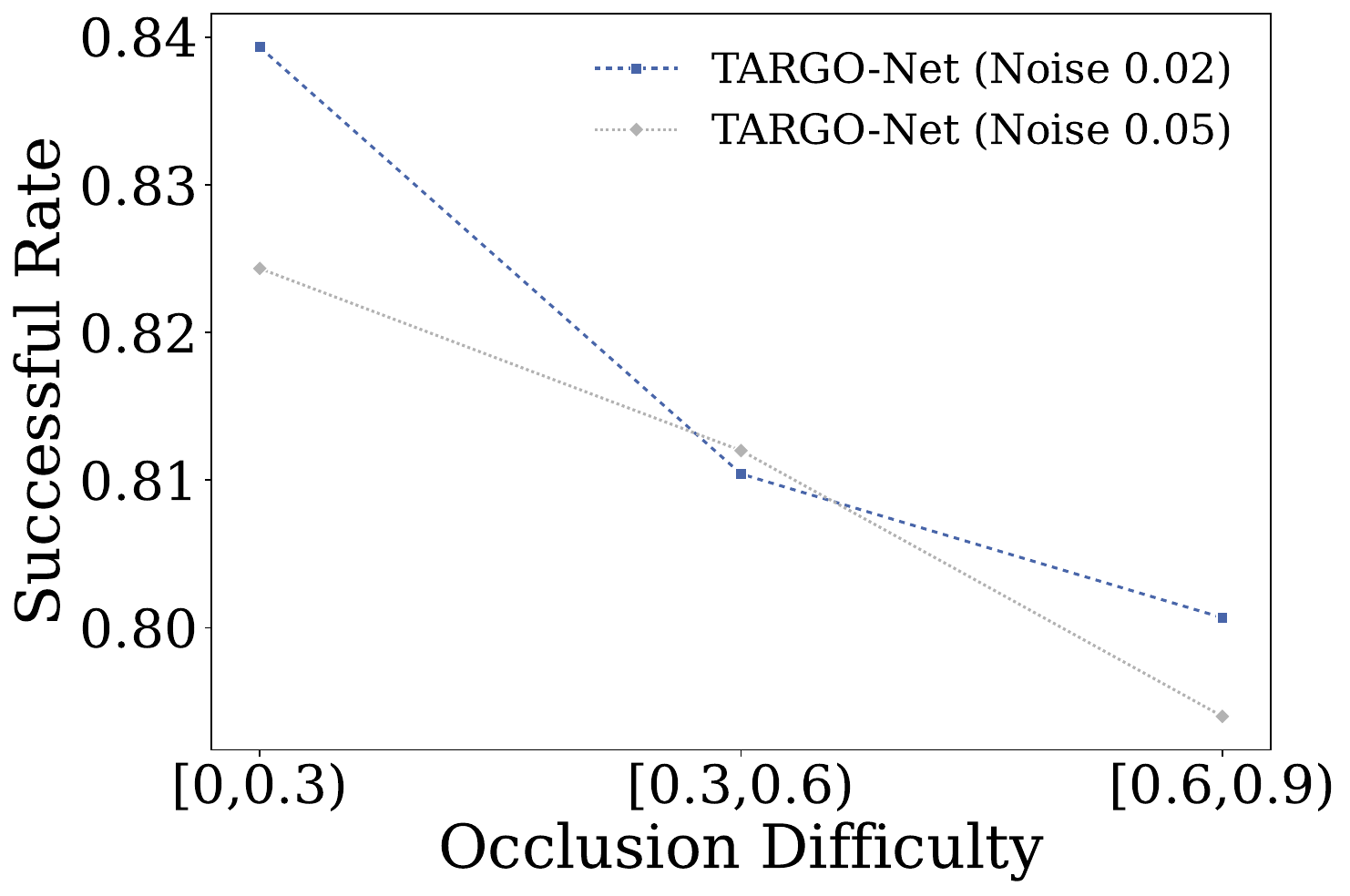}
    \caption{GSR of TARGO-Net with Gaussian depth-noise standard deviations $\sigma=\SI{0.02}{m}$ and $\sigma=\SI{0.05}{m}$.}
    \label{fig:noise_effect}
\end{figure}

We evaluate TARGO-Net after adding zero-mean Gaussian noise directly to metric depth images, using standard deviations of $\sigma=\SI{0.02}{m}$ and $\sigma=\SI{0.05}{m}$. As shown in~\Cref{fig:noise_effect}, the two settings differ by at most approximately $1.5\%$ across the three occlusion ranges, indicating limited sensitivity within the tested noise levels.

\subsubsection{Effects of the Number of TST Fusion Layers}
\label{sec:number_ca_layers}
We train variants with $L\in\{1,2,3\}$ TST fusion layers and evaluate them on the TARGO-Synthetic test set. As shown in~\Cref{fig:abl_transformer_layer}, $L=2$ yields the highest GSR in every evaluated occlusion bin. The figure legend refers to each TST fusion layer as a cross-attention (CA) layer.

\subsection{Generalization of Target Completion to Unseen Categories}
\label{sec:sc_generalization}

\begin{figure}[htbp]
    \centering
    \includegraphics[width=\linewidth]{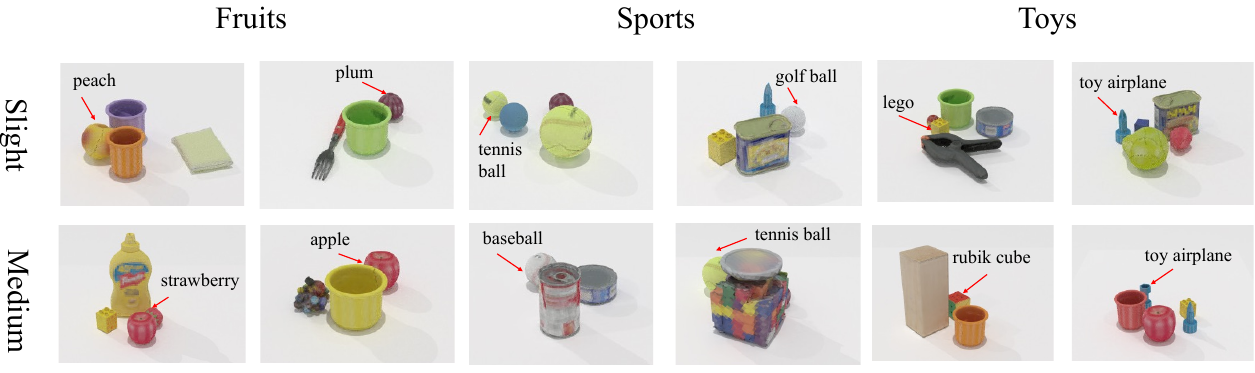}
    \caption{Object categories used to evaluate category-level generalization.}
    \label{fig:category_overview}
    \vspace{-0.3cm}
\end{figure}

We test TARGO-Net's zero-shot generalization on YCB objects from three categories excluded from training: \emph{fruits}, \emph{sports equipment}, and \emph{toys} (\Cref{fig:category_overview}). In~\Cref{tab:category_generalization}, we compare TARGO-Net with GIGA and VGN under low and medium visual occlusion.

Higher completion accuracy coincides with higher GSR in these aggregate results. For \emph{fruits} and \emph{sports equipment}, TARGO-Net achieves over $70\%$ GSR under low occlusion ($72.87\%$ and $70.83\%$) and over $64\%$ under medium occlusion ($67.22\%$ and $64.17\%$), exceeding GIGA by $20$--$35\%$. TARGO-Net also has higher IoU (e.g., $84.98\%$ vs. $61.26\%$) and lower CD-$\ell_1$ (e.g., $6.3$ vs. $19.0$).

\emph{Toys} remain the most challenging category, yet TARGO-Net still achieves 49.12\% (low) and 46.03\% (medium) success, compared to GIGA's 40.74\% and 22.92\%. Overall, TARGO-Net consistently outperforms GIGA and VGN across all categories and occlusion levels, highlighting the benefit of accurate target completion for generalizable grasping under visual occlusion.

\begin{table}[htbp]
\centering
\caption{Zero-shot GSR and target completion accuracy across unseen object categories. Each method row reports GSR; the following parenthetical row reports (IoU \%, CD-$\ell_1\times1000$). VGN does not predict completed geometry.}
\label{tab:category_generalization}
\begin{tabular}{llccc}
\toprule
\textbf{Occlusion} & \textbf{Method} & \textbf{Fruits} & \textbf{Sports} & \textbf{Toys} \\
\midrule
\multirow{5}{*}{Low} & \textbf{TARGO-Net} & \textbf{72.87\%} & \textbf{70.83\%} & \textbf{49.12\%} \\
& & (84.98, 6.3) & (86.11, 6.5) & (52.66, 13.1) \\
& \textbf{GIGA}~\cite{giga} & 46.96\% & 45.31\% & 40.74\% \\
& & (61.26, 19.0) & (60.55, 21.4) & (44.92, 29.8) \\
& \textbf{VGN}~\cite{vgn} & 36.84\% & 41.67\% & 28.24\% \\
\midrule
\multirow{5}{*}{Medium} & \textbf{TARGO-Net} & \textbf{67.22\%} & \textbf{64.17\%} & \textbf{46.03\%} \\
& & (80.77, 9.9) & (81.00, 11.2) & (47.25, 14.2) \\
& \textbf{GIGA}~\cite{giga} & 32.22\% & 33.33\% & 22.92\% \\
& & (51.83, 27.8) & (52.43, 23.4) & (36.39, 31.9) \\
& \textbf{VGN}~\cite{vgn} & 22.22\% & 22.50\% & 17.01\% \\
\bottomrule
\end{tabular}
\end{table}

%% file: sections/6_disc.tex
\section{Discussion}
\label{sec:discussion}

\subsection{Additional Factors Influencing Target-driven Grasping}

Beyond the \emph{visual occlusion level} ($\OL$) studied throughout this paper, we investigate three further factors that may affect the grasp success rate (GSR) of TARGO-Net: \emph{target completion} (whether the completed target shape is provided to the grasp detector), \emph{minimum dimension} (the smallest extent of the box bounding the target), and \emph{number of occluders} (the count of non-target objects in the cluttered scene). In the following, we quantify how much each of the four factors contributes to GSR, both overall and at different factor values, using a SHAP summary plot and line plots.

\subsubsection{Explanation Method}
To attribute the GSR of TARGO-Net to these factors, we use Kernel SHAP~\cite{shap}, which produces SHAP (SHapley Additive exPlanation) values as a unified measure of factor importance. Each SHAP value quantifies how much one factor raises or lowers the predicted GSR of a specific scene relative to the mean GSR over all scenes.

Concretely, a scene is represented by four features $\mathbf{x}=[x_1,x_2,x_3,x_4]^\top$:
\begin{enumerate*}[label=\arabic*)]
\item $x_1$: target completion indicator ($1$ if the completed target is used, $0$ otherwise), \item $x_2$: visual occlusion level, \item $x_3$: target minimum dimension, and \item $x_4$: number of occluders.
\end{enumerate*}

Treating the predicted GSR as a function $f$ of these features, Kernel SHAP fits an additive explanation that satisfies
\begin{equation}
\label{eq:additive_fa}
f(\mathbf{x}) = \phi_0 + \sum_{i=1}^{4}\phi_i(\mathbf{x}),
\end{equation}
where $\phi_0=0.733$ is the mean predicted GSR over all scenes and $\phi_i(\mathbf{x})\in\mathbb{R}$ is the contribution of feature $i$ for that scene. The contributions are estimated by weighted linear regression over feature coalitions in the vicinity of $\mathbf{x}$.

The mean SHAP values of target completion, visual occlusion, minimum dimension, and number of occluders are $0.0791$, $-0.107$, $-0.0911$, and $-0.0397$, respectively. Target completion is the only factor whose average contribution is positive, whereas visual occlusion has the largest negative average contribution and the number of occluders the smallest. Because positive and negative contributions of the same factor can cancel across scenes, these signed means describe the average direction of each effect rather than a strict importance ranking; the per-scene distributions below give a more complete picture.

\subsubsection{Summary Plot}

\begin{figure}[b]
    \centering
    \includegraphics[width=0.7\linewidth]{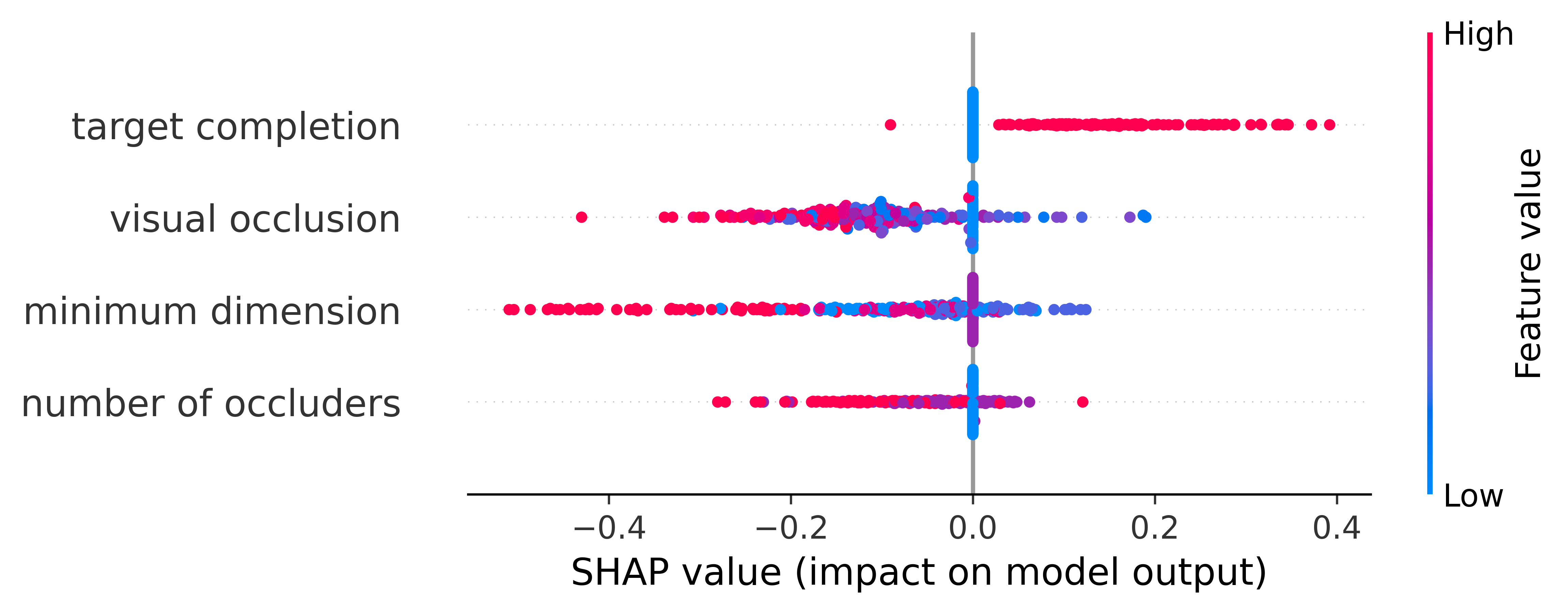}
    \caption{SHAP summary plot. Each point represents a scene. The horizontal position is the factor's SHAP value for that scene: points left of the gray vertical line lower the predicted GSR relative to the mean prediction, while points to the right raise it. Color denotes the raw factor value (blue = low, red = high).}
    \label{fig:shap_summary}
\end{figure}

The summary plot in~\Cref{fig:shap_summary} shows the SHAP value of each factor for every scene, revealing both the magnitude and the direction of its effect.

Target completion ($x_1$): When completion is present (feature value $1$, red points), the SHAP values are almost all positive, indicating a clear gain in GSR. When it is absent (feature value $0$, blue points), the values cluster near zero. This agrees with the target completion ablation in~\Cref{fig:abl_sc}.

Visual occlusion ($x_2$): Highly occluded scenes (red points) lie far to the left and consistently lower the predicted GSR, whereas lightly occluded scenes (blue points) lie near zero or slightly to the right. Visual occlusion is thus the strongest negative factor.

Minimum dimension ($x_3$): Targets whose minimum dimension falls in the $(0.0277,0.0393]$ m bin (about $3$ cm) have the highest mean SHAP value and are the easiest to grasp. Very large targets produce more negative contributions than very small ones.

Number of occluders ($x_4$): The mean SHAP value decreases as the number of occluders grows from three to five, reflecting the increased difficulty of reaching the target, although the effect is milder than that of visual occlusion.

\subsubsection{Line Plots}
\begin{figure*}[hbtp]
  \centering
  \begin{subfigure}[b]{0.495\linewidth}
    \centering
    \includegraphics[width=\linewidth]{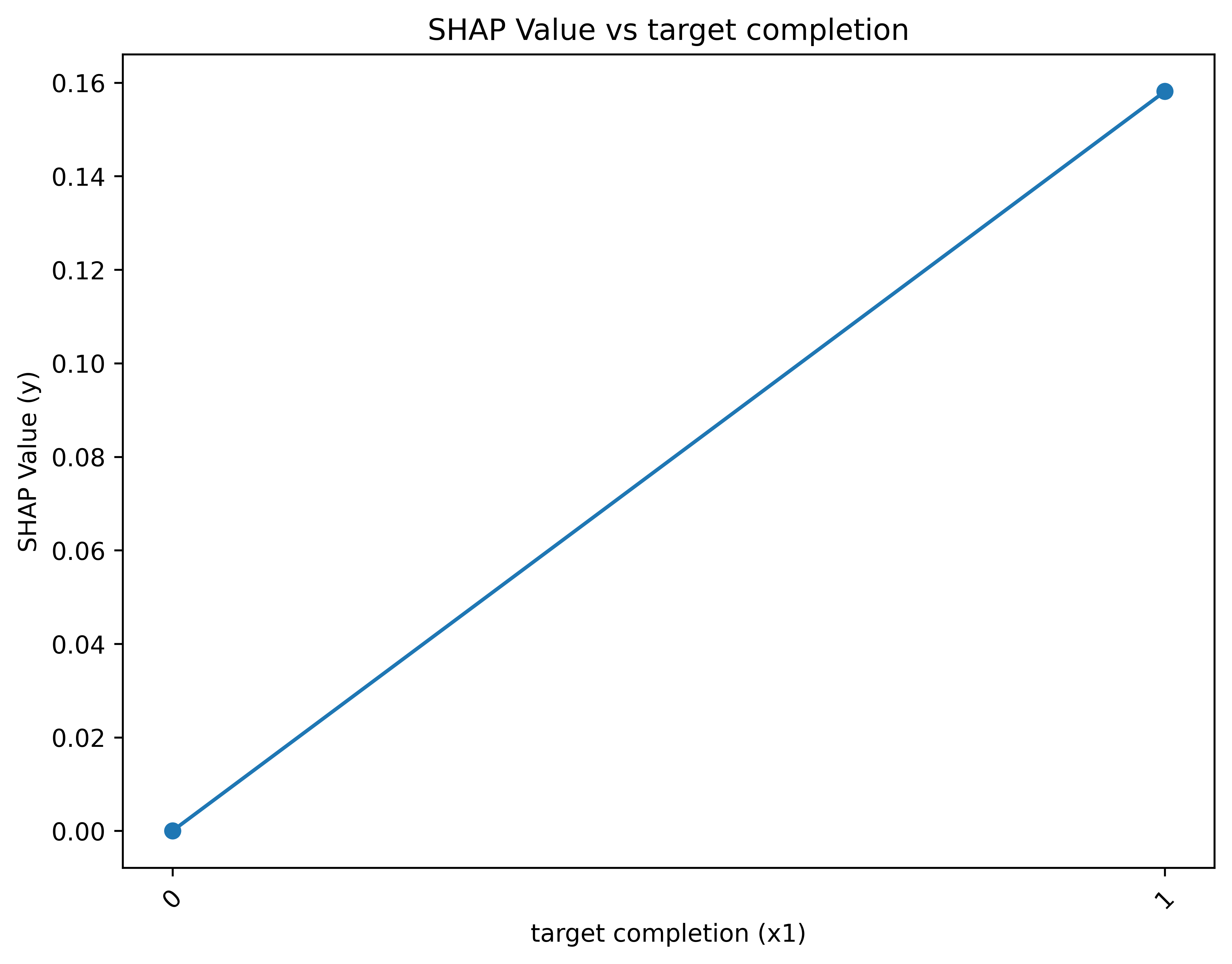}
    \caption{$x_1$: target completion.}
    \label{fig:shap_line_target_completion}
  \end{subfigure}
  \hfill
  \begin{subfigure}[b]{0.465\linewidth}
    \centering
    \includegraphics[width=\linewidth]{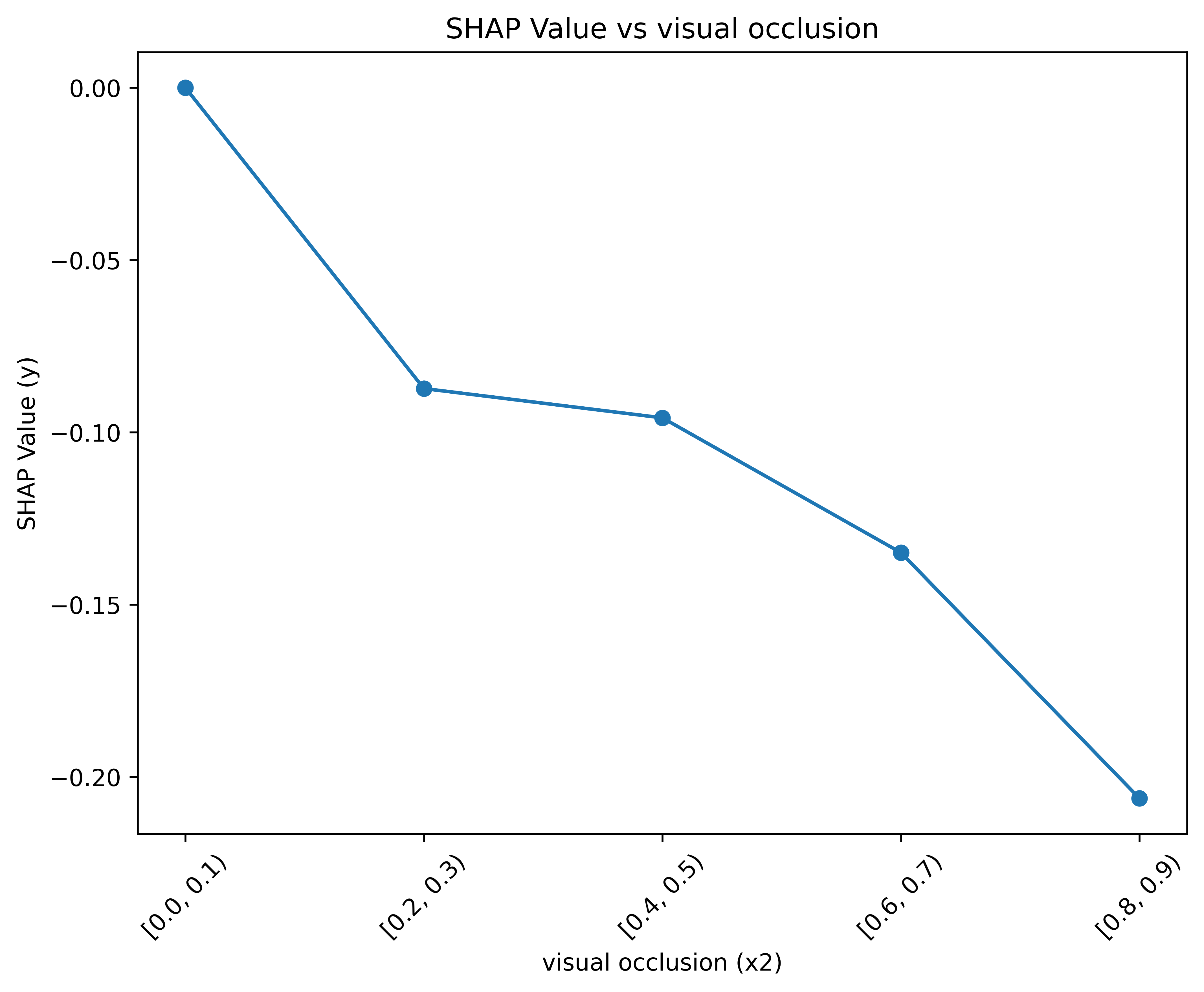}
    \caption{$x_2$: visual occlusion level.}
    \label{fig:shap_line_vis_occ}
  \end{subfigure}

  \begin{subfigure}[b]{0.455\linewidth}
    \centering
    \includegraphics[width=\linewidth]{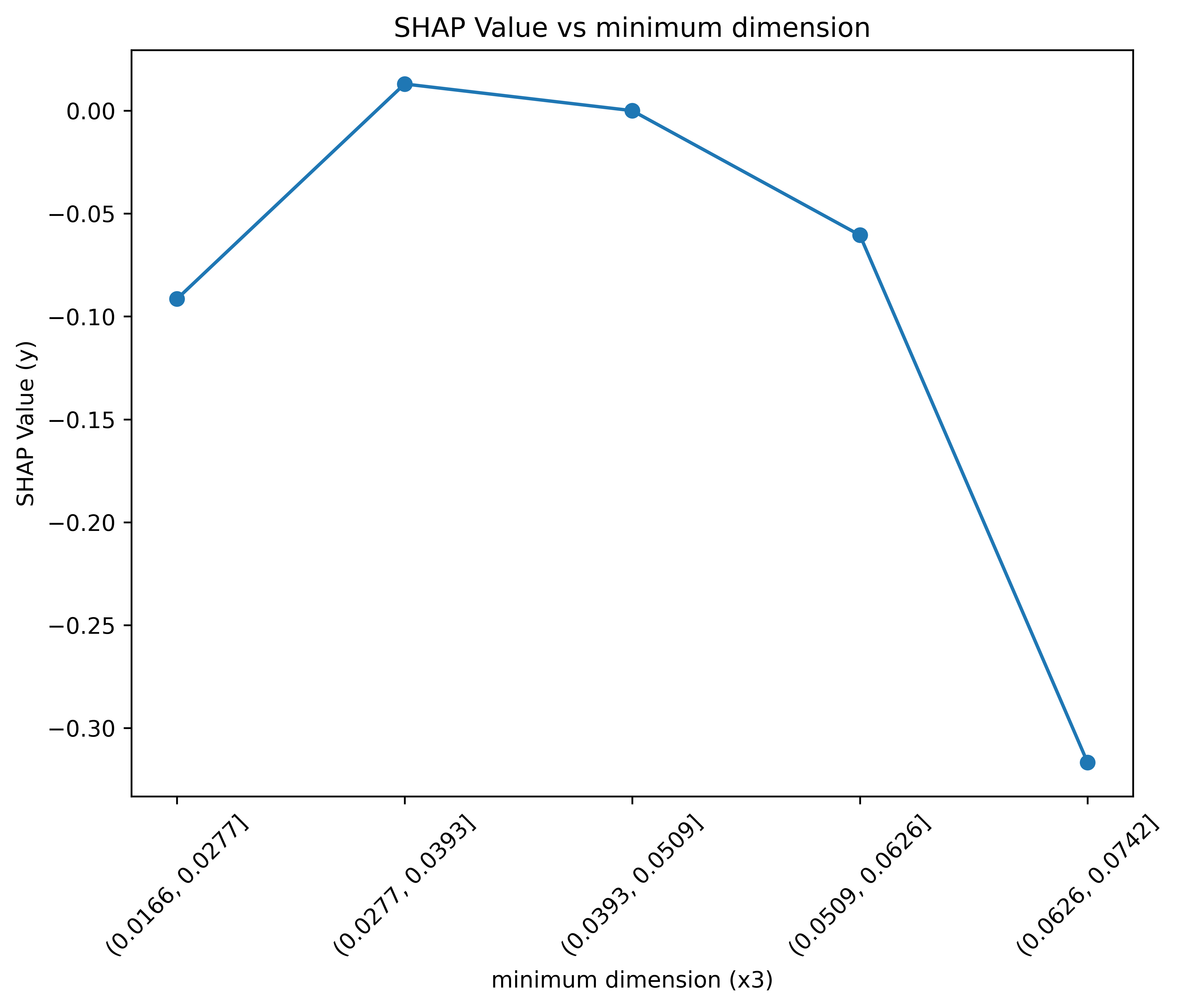}
    \caption{$x_3$: minimum dimension.}
    \label{fig:shap_line_min_dim}
  \end{subfigure}
  \hfill
  \begin{subfigure}[b]{0.505\linewidth}
    \centering
    \includegraphics[width=\linewidth]{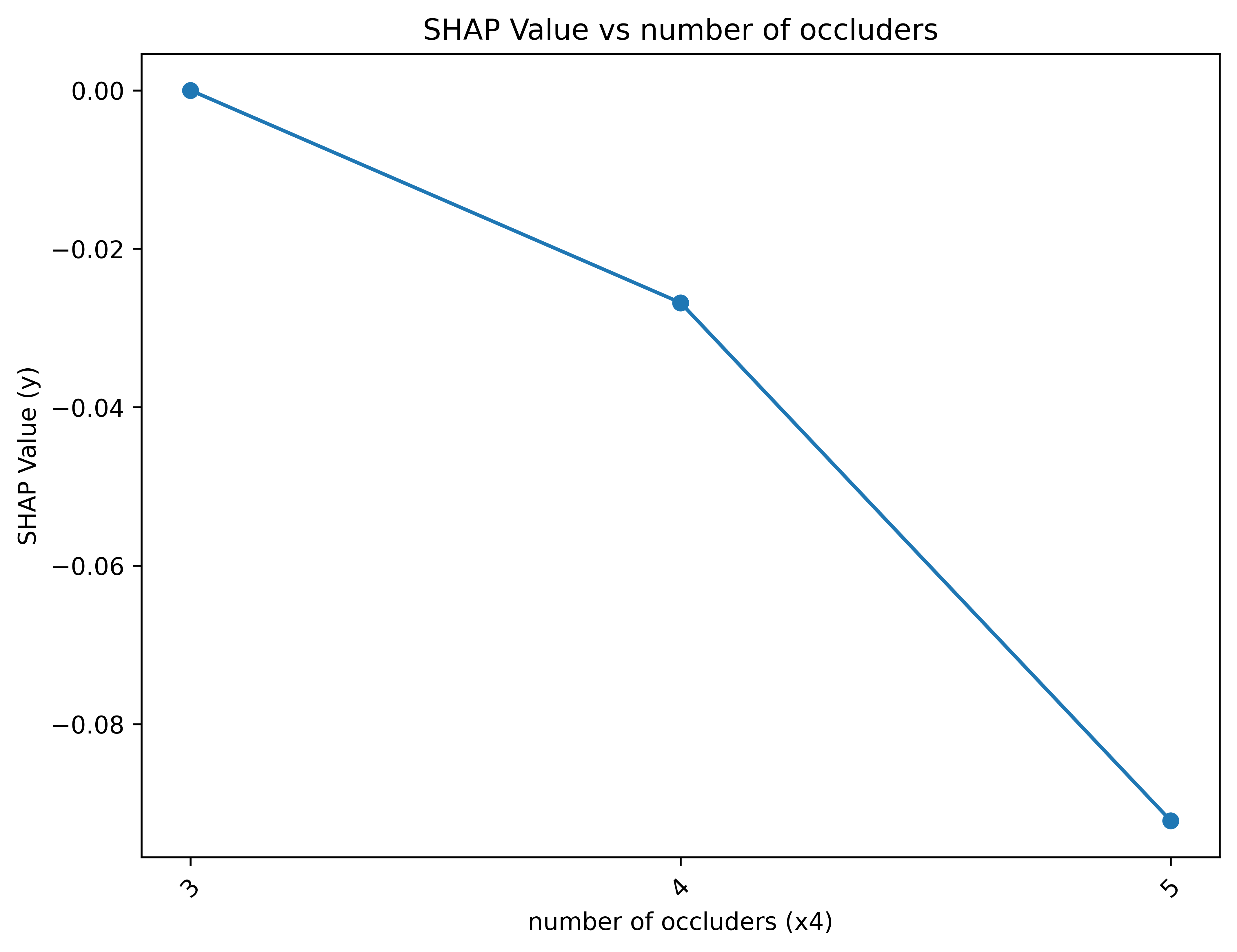}
    \caption{$x_4$: number of occluders.}
    \label{fig:shap_line_numocc}
  \end{subfigure}

  \caption{Mean SHAP values as a function of the four factors: \textit{target completion}, \textit{visual occlusion level}, \textit{target minimum dimension}, and \textit{number of occluders}.}
  \label{fig:shap_line}
\end{figure*}

The line plots in~\Cref{fig:shap_line} show the mean SHAP value at each factor value or bin. Presence of target completion yields the highest mean contribution. The mean contribution drops steadily as visual occlusion or the number of occluders increases. For minimum dimension, both the smallest and the largest bins lower GSR, with the largest bin showing the strongest decrease.

\subsection{Limitations of this Work}

\subsubsection{Sim2Real Gap}
Although our pipeline transfers to real-world experiments (\Cref{sec:real_world_exp}), its performance declines on TARGO-Real (\Cref{sec:targo_real}). We attribute this gap mainly to two factors: \emph{dataset size} and \emph{sensor noise patterns}. Both the target completion and grasp detection networks are trained on TARGO-Synthetic, which contains only $114$ training objects and thus limits the shape diversity seen during training. Moreover, we only add Gaussian noise to the synthetic depth maps, whereas real depth sensors also produce other artifacts, such as gamma noise and missing depth values. Future datasets with broader coverage of tabletop objects, calibrated sensor-noise models or domain adaptation, and a more robust target completion network could narrow the sim-to-real gap.

Moreover, the contact and friction models of the PyBullet physics simulator~\cite{pybullet} do not reproduce every physical interaction. VGN~\cite{vgn}, for example, reports failures caused by a lack of friction when the gripper fingers are not deep enough. Because real-world robotic data is expensive to collect, current grasp learning methods, including ours, are trained mainly on synthetic data. Building realistic and efficient simulators, complemented by real-world robotic data, is therefore key to the generalization of grasping models from simulation to the real world.

\subsection{Future Directions}

\subsubsection{Simple Metric}

Our definition of \textit{visual occlusion level} ($\OL$) is intentionally simple and does not capture which parts of an object are occluded. A more nuanced \textit{actionable visual occlusion level} could instead weight task-relevant regions, such as the handle of a mug in functional grasping. HANDAL~\cite{guo2023handal}, for instance, introduces the notion of \textit{actionable affordance} and provides affordance segmentations for real-world manipulable objects. Our benchmark, however, addresses general rigid-object grasping, where any surface patch is considered graspable and no functional grasp intent is imposed, so a part-agnostic $\OL$ is appropriate for this setting.

The occlusion score of~\cite{occ_def} for 3D object recognition assumes access to the full object mesh. By contrast, our $\OL$ requires no mesh or ground-truth 3D geometry, only the paired single scene as a reference, and measures the fraction of target pixels hidden in the cluttered view rather than the fraction of the 3D surface that is invisible. It can therefore be computed from depth images alone, without object models.

Despite its simplicity, our $\OL$ metric serves the purpose of effectively evaluating model performance under visual occlusion: in~\Cref{fig:abl_baseline_comparison}, all baseline models exhibit a steep decline in success rate as $\OL$ increases. We hope our benchmark will spur the development of more sophisticated occlusion metrics and methods within the robotics community.

\subsubsection{Full Distribution of Grasp Parameters}
During training, each grasp center is paired with a single ground-truth grasp pose. This supervision is sufficient for target-driven grasping, but it does not represent the potentially multimodal distribution of feasible orientations and widths at a grasp center. Generative modeling is a natural way to capture this full distribution. GenPose~\cite{genpose}, for example, uses score-based diffusion to generate multiple object-pose hypotheses; we plan to extend our work with analogous techniques to generate and score multiple grasp hypotheses per target.

%% file: sections/7_concl.tex
\section{Conclusion}

To the best of our knowledge, this is the first study to systematically evaluate target-driven 6-DoF grasping under controlled, target-specific visual occlusion levels.
In this work, we introduced TARGO, a benchmark dataset that combines large-scale simulated data derived from VGN~\cite{vgn} (TARGO-Synthetic) with real RGB-D observations selected from DexYCB~\cite{dexycb} (TARGO-Real) to evaluate model robustness against visual occlusion. We further developed TARGO-Net, which predicts robust grasp poses at all levels of visual occlusion and outperforms the evaluated baselines on both TARGO-Synthetic and TARGO-Real. Real-world robot experiments confirm that its advantage grows as occlusion increases.
Specifically, our study yields several key findings:
\begin{enumerate*}[label=\arabic*)]
    \item Single-scene grasp augmentation improves GSR for all evaluated models and helps most under heavy occlusion.
    \item Attending to both occluders and the target, as our Target-Scene Transformer (TST) does through bidirectional cross-attention, captures their mutual geometry, which is critical for accurate target-driven grasping.
    \item In addition to \emph{visual occlusion level}, \emph{target completion}, \emph{number of occluders}, and \emph{target minimum dimension} are also determining factors for grasp success, as shown by our ablations and SHAP attribution analysis.
\end{enumerate*}

%% file: sections/appendix.tex
\newpage
\appendix
\section{Dataset and Experimental Details}
\subsection{Objects Used in Real-World Experiments and TARGO-Real}
\label{sec:details_objects_real}
\Cref{tab:obj_details_rw,tab:obj_details_targo_real} list the names, masses, and measured dimensions of the $18$ objects used in the real-world experiments (\Cref{sec:real_world_exp}) and the $21$ YCB objects in TARGO-Real (\Cref{sec:targo_real}). Three values denote the extents along three axes, two values denote the width (or diameter) and height of a roughly cylindrical object, and a single value denotes the diameter of a roughly spherical object.

\subsection{Objects Used in the Category-Generalization Experiment}
\label{sec:details_objects_categories}
\Cref{tab:obj_details_categories} lists the $20$ YCB objects used in the category-generalization experiment (\Cref{sec:sc_generalization}), which belong to three categories held out from training: fruits, sports equipment, and toys.

\setlength{\tabcolsep}{4.5pt}

\begin{longtable}{|c|>{\centering\arraybackslash}m{2.5cm}|c|c|c|}
    \caption{Details of objects used in real-world experiments.} \label{tab:obj_details_rw} \\
    \hline
    \textbf{ID} & \textbf{Picture} & \textbf{Object Name} & \textbf{Mass (g)} & \textbf{Dimensions (mm)} \\
    \hline
    \endfirsthead
    \multicolumn{5}{c}{\tablename~\thetable\ (continued)} \\
    \hline
    \textbf{ID} & \textbf{Picture} & \textbf{Object Name} & \textbf{Mass (g)} & \textbf{Dimensions (mm)} \\
    \hline
    \endhead
    \hline
    \multicolumn{5}{r}{Continued on next page} \\
    \endfoot
    \hline
    \endlastfoot
    1 & \includegraphics[width=2cm]{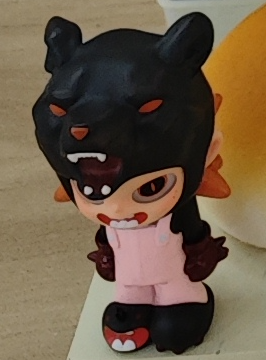} & Figurine & 110 & $38 \times 46 \times 85$ \\
    \hline
    2 & \includegraphics[width=2cm]{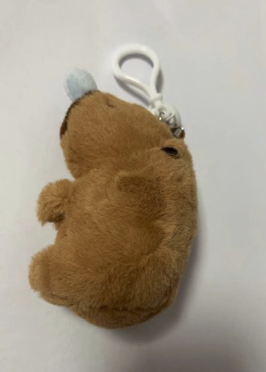} & Guinea pig & 72 & $62 \times 70 \times 115$ \\
    \hline
    3 & \includegraphics[width=2cm]{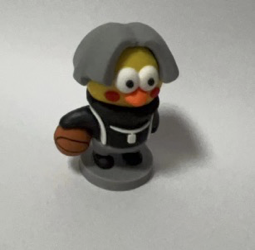} & Kunkun & 80 & $31 \times 43 \times 60$ \\
    \hline
    4 & \includegraphics[width=2cm]{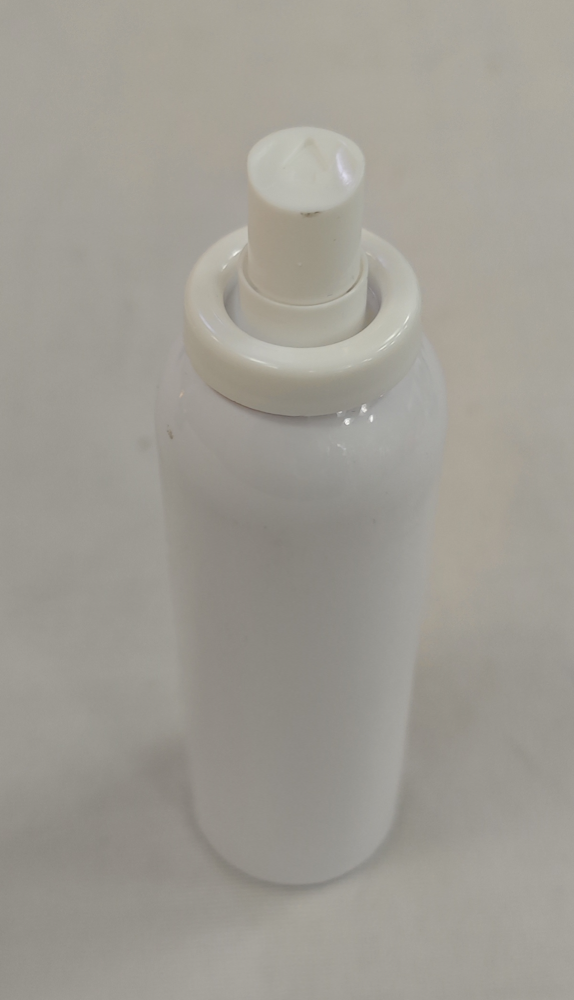} & Sprayer & 39.1 & $44.7 \times 165$ \\
    \hline
    5 & \includegraphics[width=2cm]{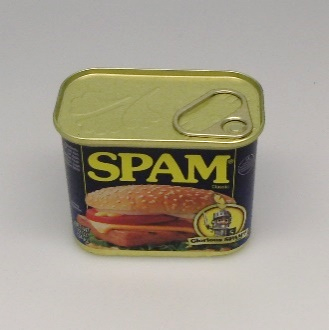} & Spam & 370 & $50 \times 97 \times 82$ \\
    \hline
    6 & \includegraphics[width=2cm]{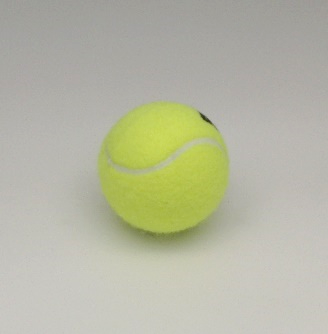} & Tennis ball & 58 & 64.7 \\
    \hline
    7 & \includegraphics[width=2cm]{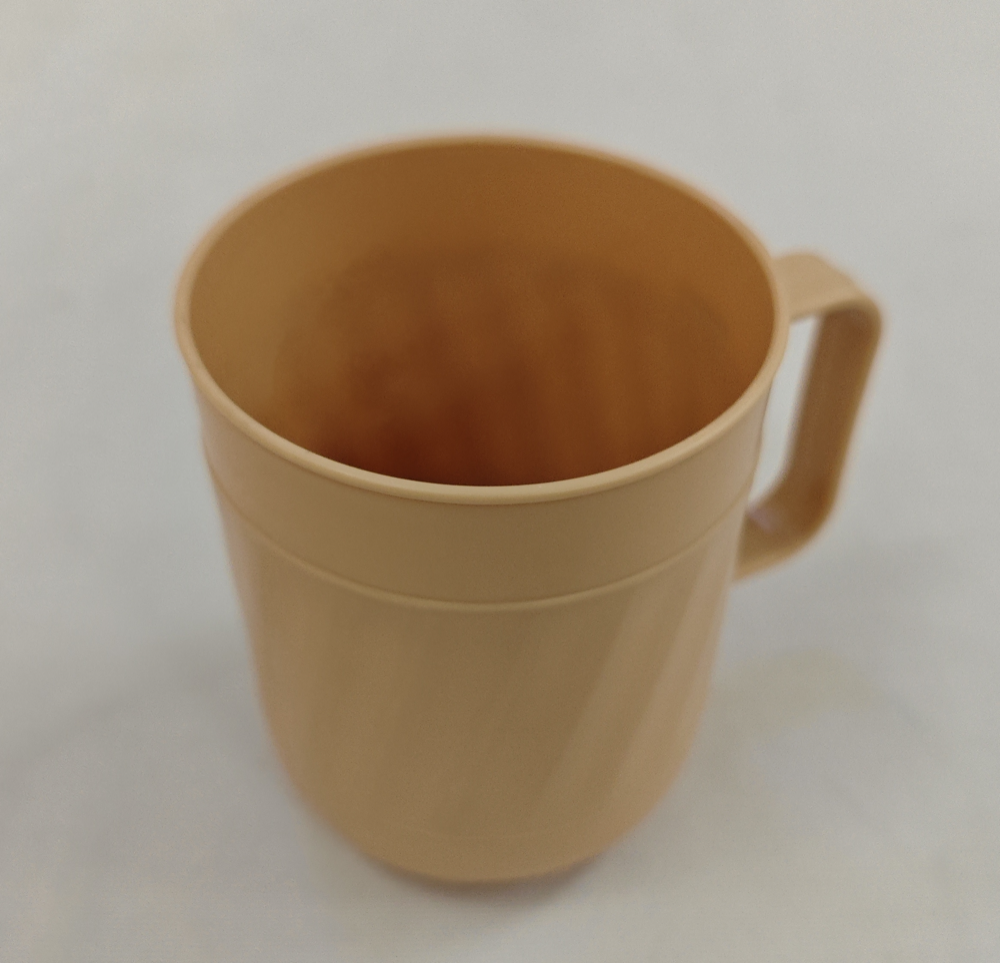} & Brown cup & 43.2 & $79.8 \times 105$ \\
    \hline
    8 & \includegraphics[width=2cm]{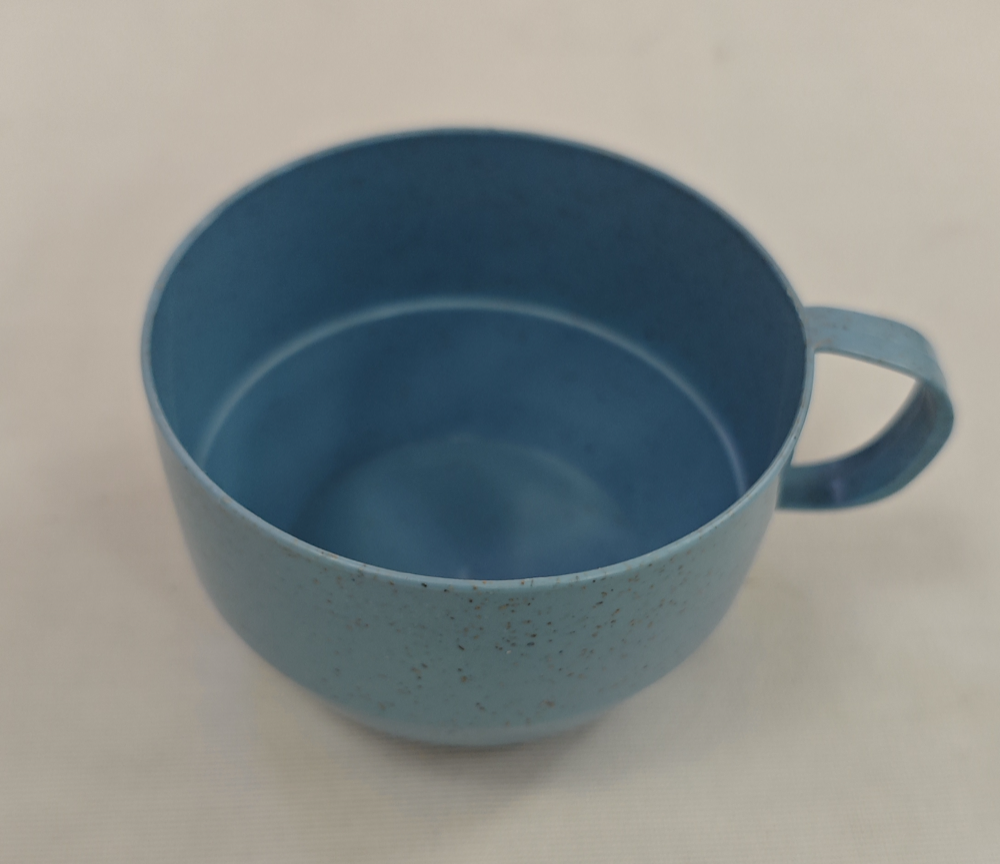} & Blue cup & 26.3 & $94 \times 61.5$ \\
    \hline
    9 & \includegraphics[width=2cm]{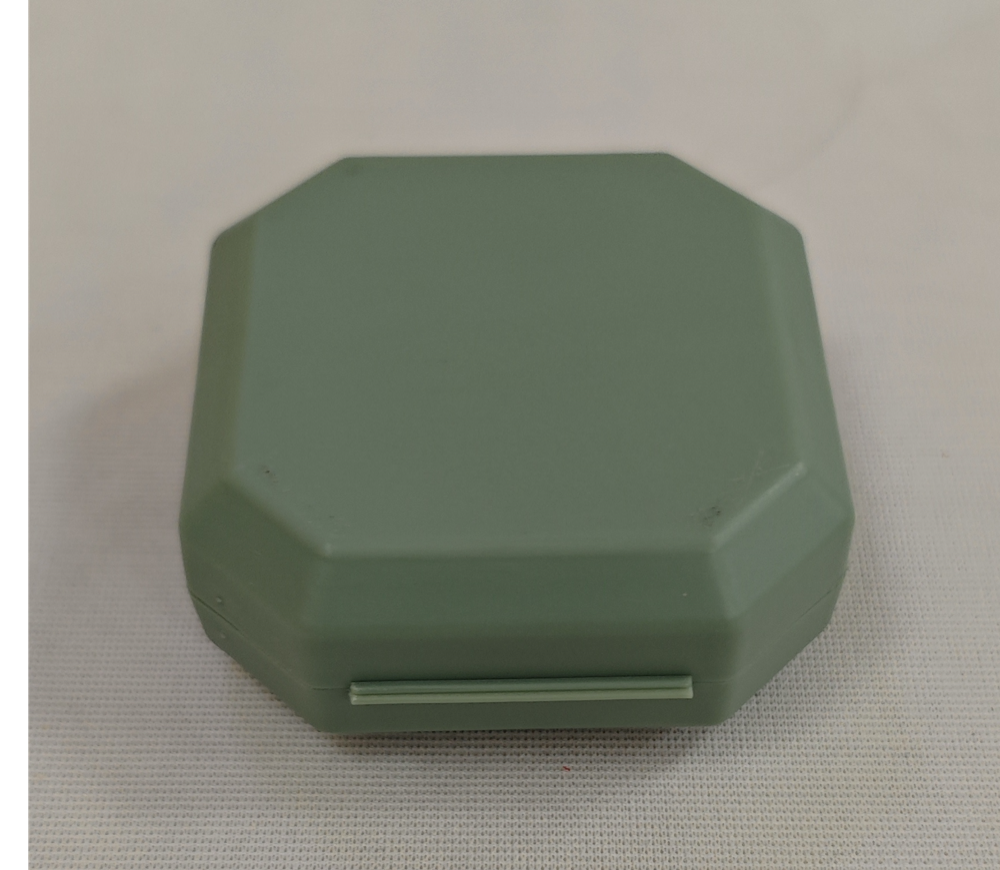} & Green box & 38.2 & $29.5 \times 70 \times 70$ \\
    \hline
    10 & \includegraphics[width=2cm]{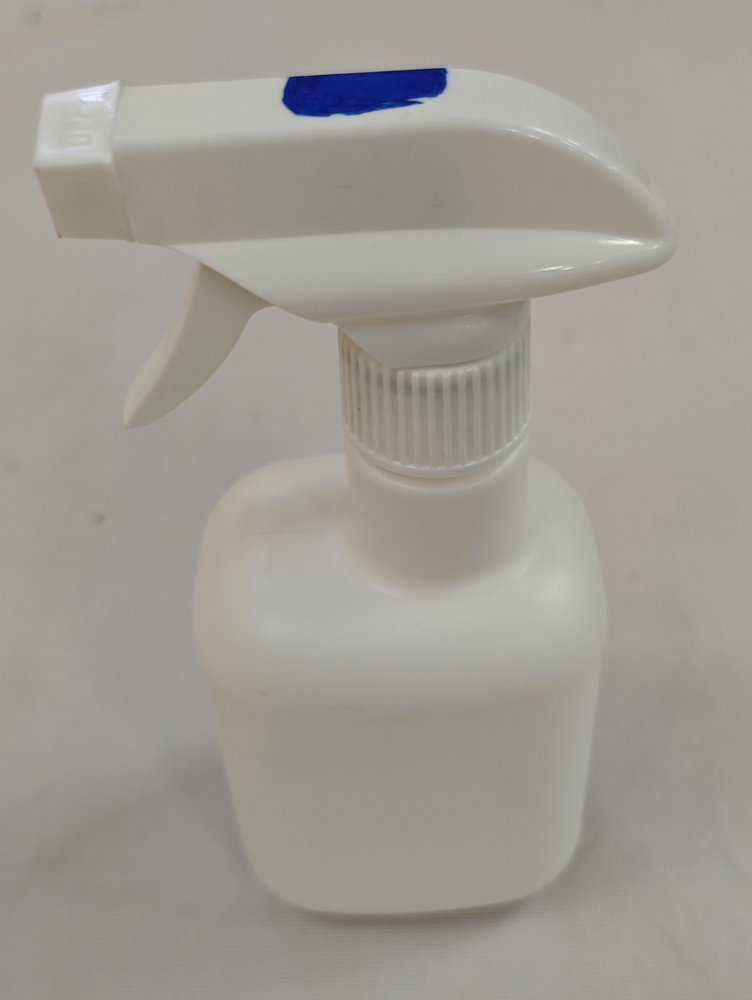} & Watering bottle & 47 & $58.1 \times 76.1 \times 175$ \\
    \hline
    11 & \includegraphics[width=2cm]{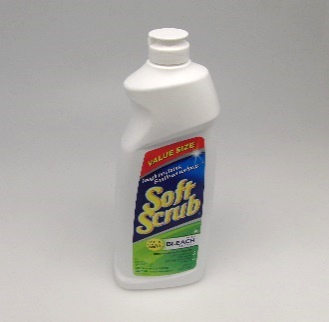} & Bleach & 1131 & $250 \times 98 \times 65$ \\
    \hline
    12 & \includegraphics[width=2cm]{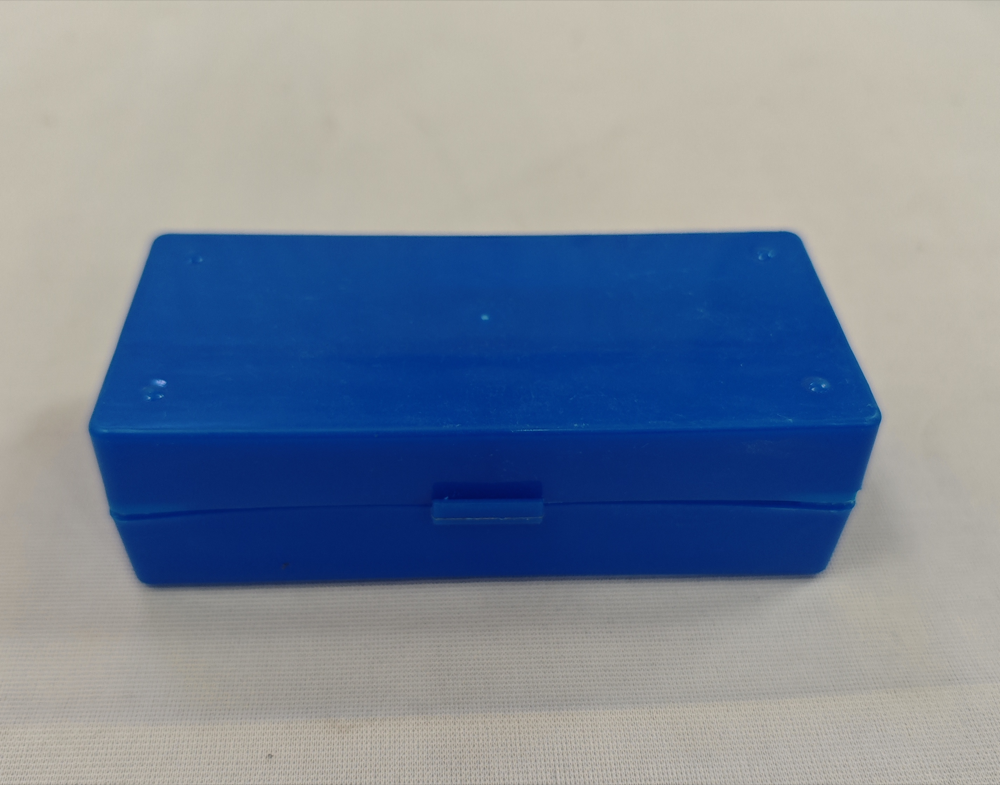} & Blue box & 56.9 & $41 \times 59.1 \times 132.7$ \\
    \hline
    13 & \includegraphics[width=2cm]{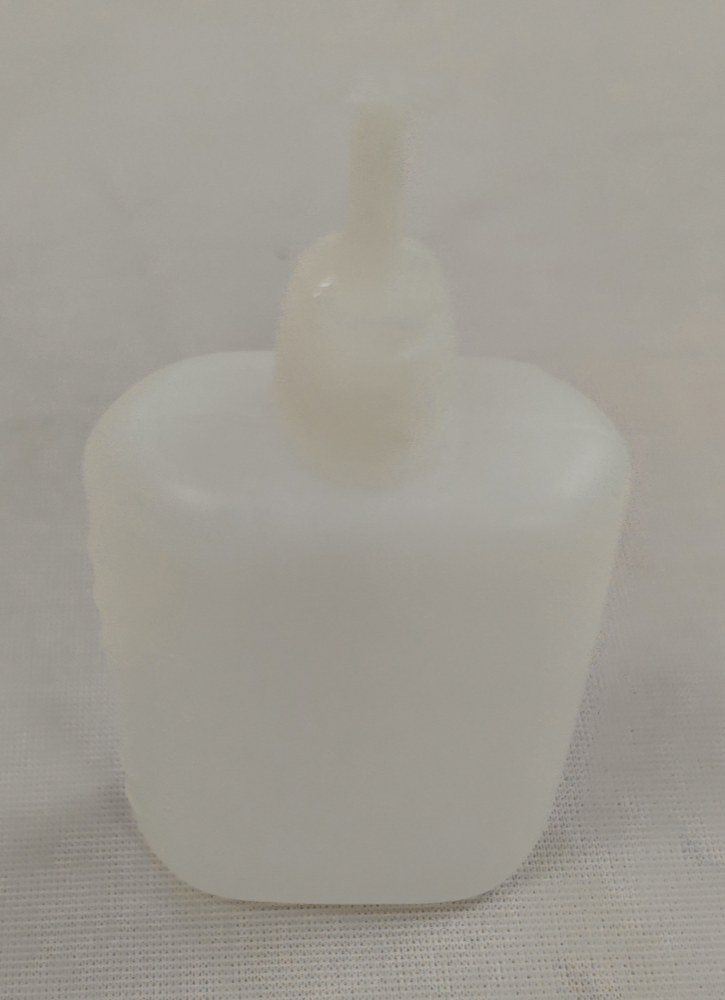} & Glue & 48 & $19.7 \times 50.5 \times 108$ \\
    \hline
    14 & \includegraphics[width=2cm]{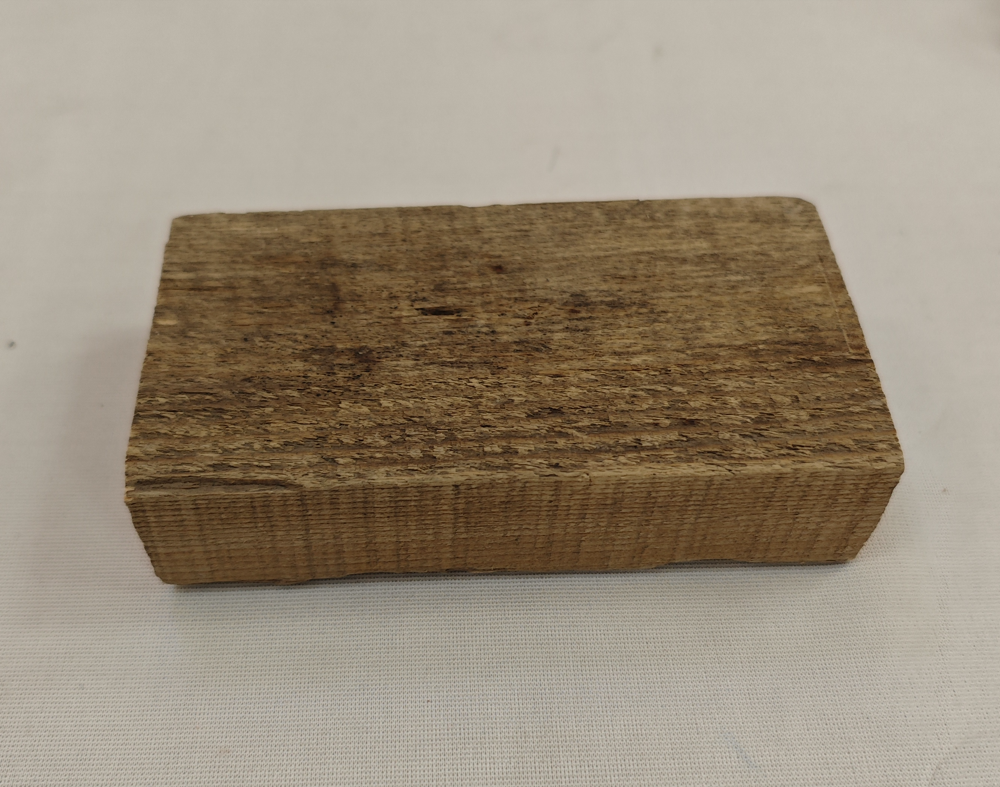} & Wooden block & 191 & $34.3 \times 83.7 \times 147.8$ \\
    \hline
    15 & \includegraphics[width=2cm]{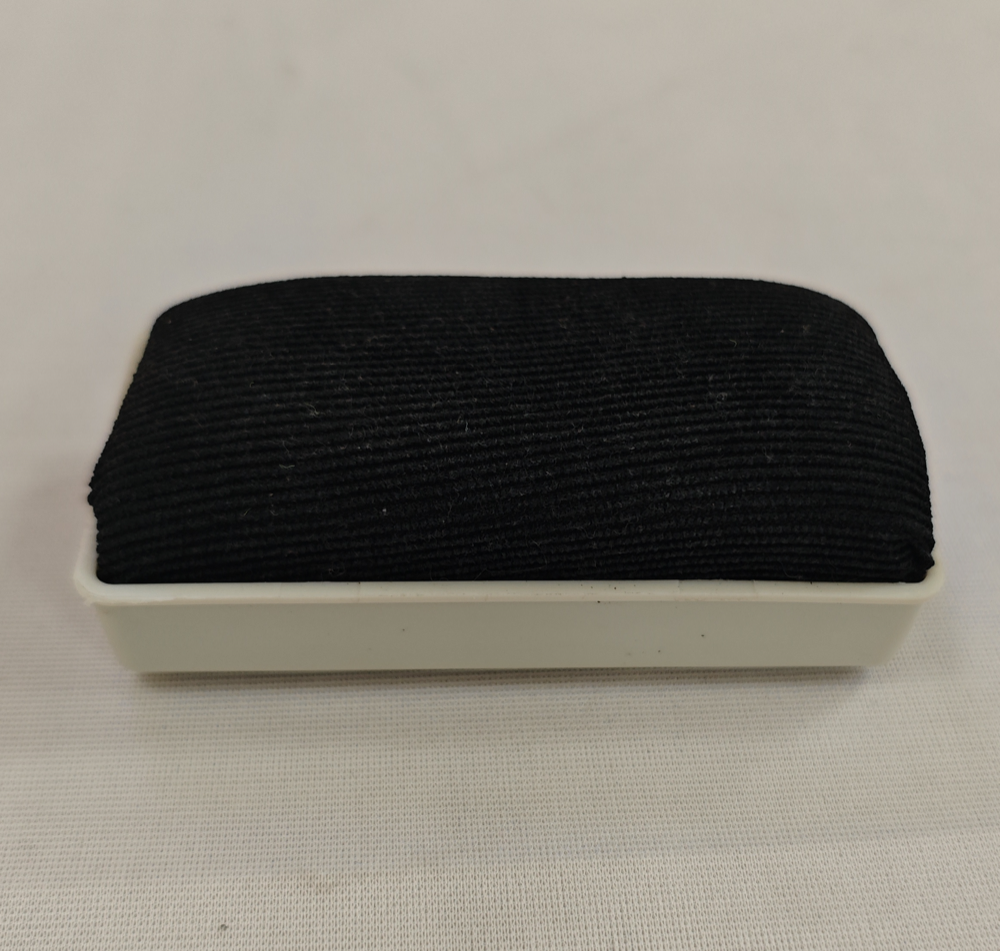} & Eraser & 28.7 & $39.7 \times 53 \times 110.1$ \\
    \hline
    16 & \includegraphics[width=2cm]{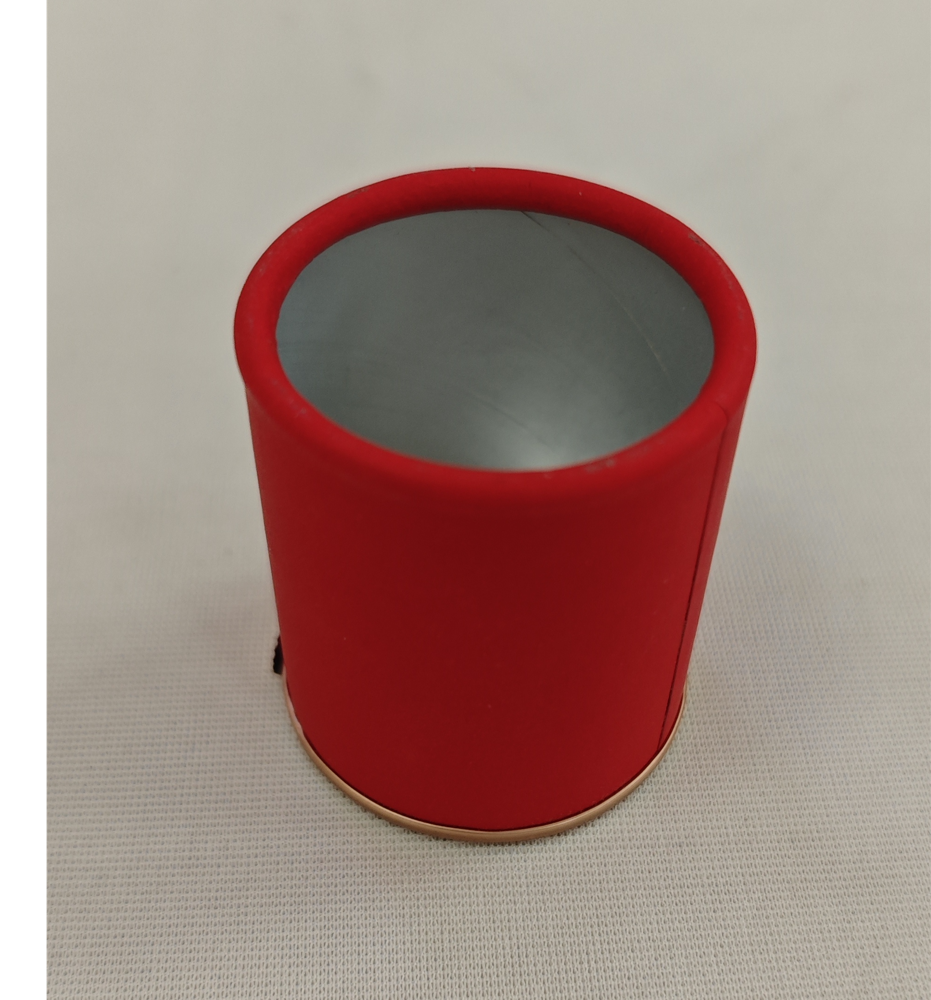} & Red can & 86 & $46.9 \times 53.1$ \\
    \hline
    17 & \includegraphics[width=2cm]{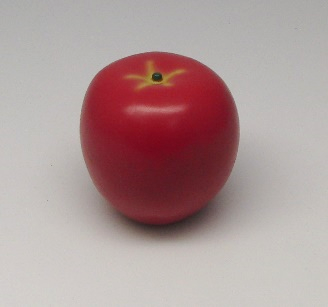} & Apple & 68 & 75 \\
    \hline
    18 & \includegraphics[width=2cm]{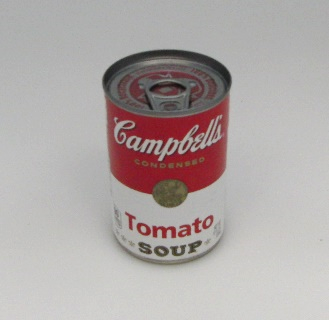} & Tomato can & 349 & $66 \times 101$ \\
    \hline
\end{longtable}

\begin{longtable}{|c|>{\centering\arraybackslash}m{2.5cm}|c|c|c|}
\caption{Details of objects used in TARGO-Real.} \label{tab:obj_details_targo_real} \\
    \hline
    \textbf{ID} & \textbf{Picture} & \textbf{Object Name} & \textbf{Mass (g)} & \textbf{Dimensions (mm)} \\
    \hline
    \endfirsthead
    \multicolumn{5}{c}{\tablename~\thetable\ (continued)} \\
    \hline
    \textbf{ID} & \textbf{Picture} & \textbf{Object Name} & \textbf{Mass (g)} & \textbf{Dimensions (mm)} \\
    \hline
    \endhead
    \hline
    \multicolumn{5}{r}{Continued on next page} \\
    \endfoot
    \hline
    \endlastfoot
    1 & \includegraphics[width=2.5cm]{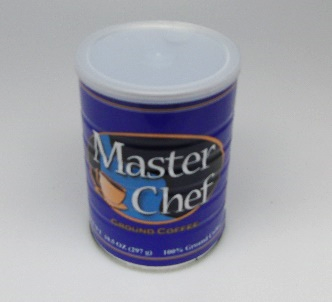} & Coffee can & 414 & $102 \times 139$ \\
    \hline
    2 & \includegraphics[width=2.5cm]{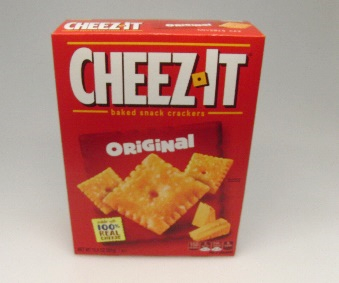} & Cracker box & 411 & $60 \times 158 \times 210$ \\
    \hline
    3 & \includegraphics[width=2.5cm]{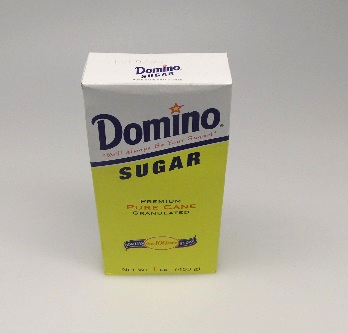} & Sugar box & 514 & $38 \times 89 \times 175$ \\
    \hline
    4 & \includegraphics[width=2.5cm]{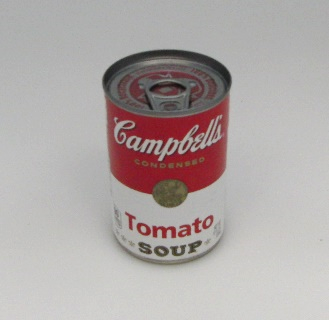} & Tomato soup can & 349 & $66 \times 101$ \\
    \hline
    5 & \includegraphics[width=2.5cm]{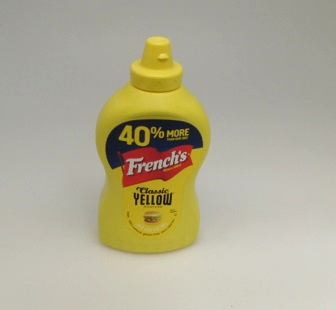} & Mustard bottle & 603 & $50 \times 85 \times 175$ \\
    \hline
    6 & \includegraphics[width=2.5cm]{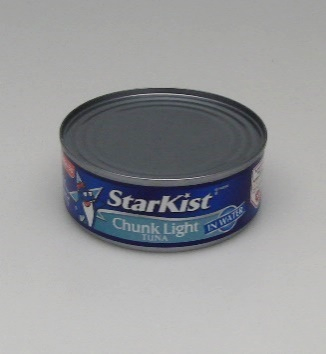} & Tuna fish can & 171 & $85 \times 33$ \\
    \hline
    7 & \includegraphics[width=2.5cm]{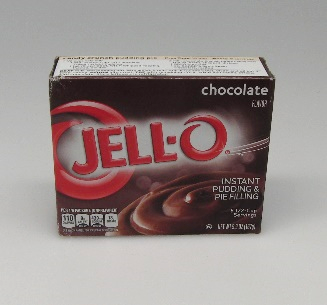} & Chocolate pudding box & 187 & $35 \times 110 \times 89$ \\
    \hline
    8 & \includegraphics[width=2.5cm]{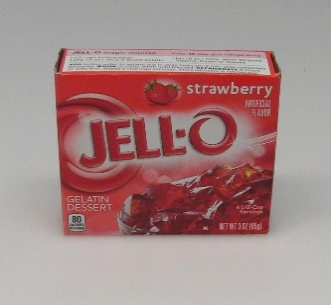} & Strawberry gelatin box & 97 & $28 \times 85 \times 73$ \\
    \hline
    9 & \includegraphics[width=2.5cm]{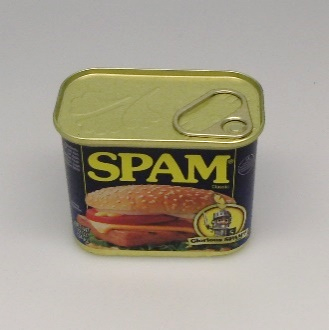} & Spam can & 370 & $50 \times 97 \times 82$ \\
    \hline
    10 & \includegraphics[width=2.5cm]{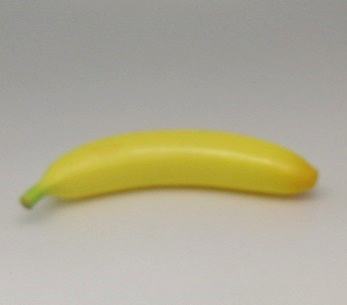} & Plastic banana & 66 & $36 \times 190$ \\
    \hline
    11 & \includegraphics[width=2.5cm]{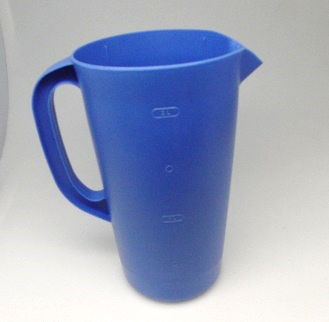} & Pitcher base & 178 & $108 \times 235$ \\
    \hline
    12 & \includegraphics[width=2.5cm]{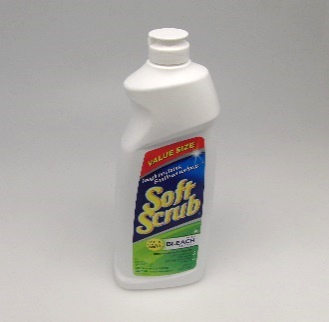} & Cleanser bottle & 1131 & $250 \times 98 \times 65$ \\
    \hline
    13 & \includegraphics[width=2.5cm]{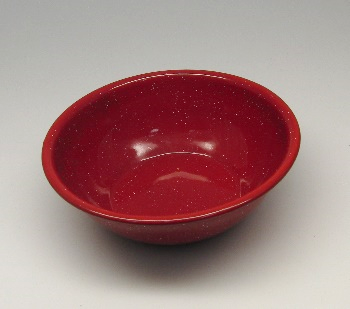} & Bowl & 147 & $159 \times 53$ \\
    \hline
    14 & \includegraphics[width=2.5cm]{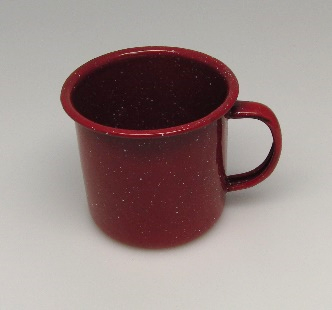} & Mug & 118 & $80 \times 82$ \\
    \hline
    15 & \includegraphics[width=2.5cm]{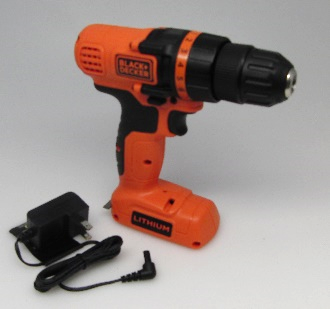} & Power drill & 895 & $35 \times 46 \times 184$ \\
    \hline
    16 & \includegraphics[width=2.5cm]{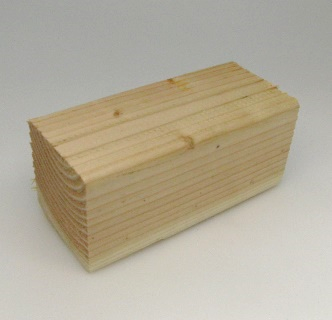} & Wood block & 729 & $85 \times 85 \times 200$ \\
    \hline
    17 & \includegraphics[width=2.5cm]{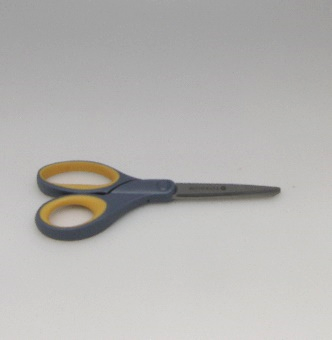} & Scissors & 82 & $87 \times 200 \times 14$ \\
    \hline
    18 & \includegraphics[width=2.5cm]{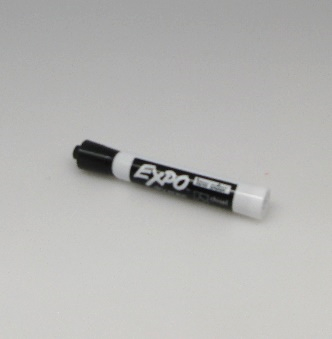} & Large marker & 15.8 & $18 \times 121$ \\
    \hline
    19 & \includegraphics[width=2.5cm]{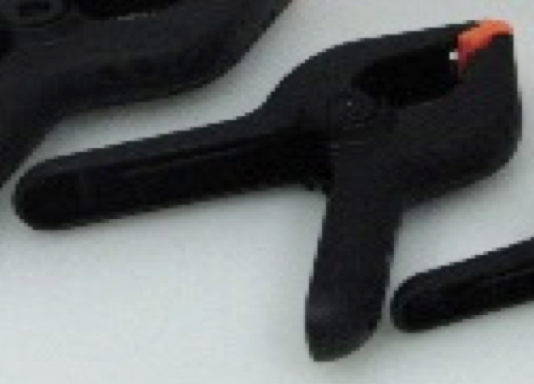} & Large clamp & 125 & $125 \times 165 \times 32$ \\
    \hline
    20 & \includegraphics[width=2.5cm]{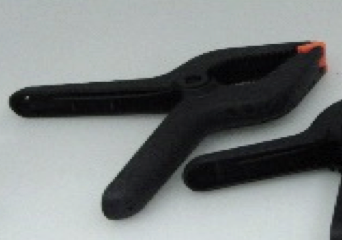} & Extra large clamp & 202 & $165 \times 213 \times 37$ \\
    \hline
    21 & \includegraphics[width=2.5cm]{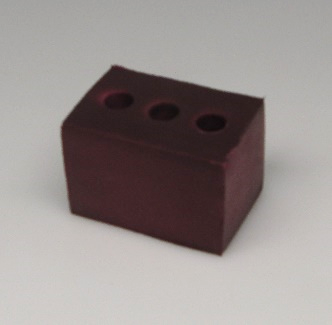} & Foam brick & 28 & $50 \times 75 \times 50$ \\
    \hline
\end{longtable}

\begin{longtable}{|c|>{\centering\arraybackslash}m{2.5cm}|c|c|c|}
    \caption{Details of objects used in the category-generalization experiment.} \label{tab:obj_details_categories} \\
    \hline
    \textbf{ID} & \textbf{Picture} & \textbf{Object Name} & \textbf{YCB ID} & \textbf{Category} \\
    \hline
    \endfirsthead
    \multicolumn{5}{c}{\tablename~\thetable\ (continued)} \\
    \hline
    \textbf{ID} & \textbf{Picture} & \textbf{Object Name} & \textbf{YCB ID} & \textbf{Category} \\
    \hline
    \endhead
    \hline
    \multicolumn{5}{r}{Continued on next page} \\
    \endfoot
    \hline
    \endlastfoot
    1 & \includegraphics[width=2.5cm]{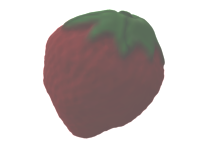} & Strawberry & 012 & Fruits \\
    \hline
    2 & \includegraphics[width=2.5cm]{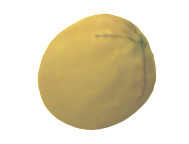} & Lemon & 014 & Fruits \\
    \hline
    3 & \includegraphics[width=2.5cm]{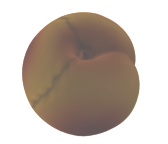} & Peach & 015 & Fruits \\
    \hline
    4 & \includegraphics[width=2.5cm]{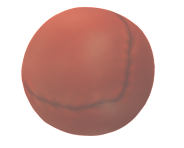} & Orange & 017 & Fruits \\
    \hline
    5 & \includegraphics[width=2.5cm]{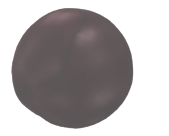} & Plum & 018 & Fruits \\
    \hline
    6 & \includegraphics[width=2.5cm]{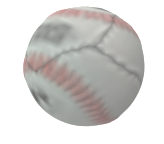} & Baseball & 055 & Sports \\
    \hline
    7 & \includegraphics[width=2.5cm]{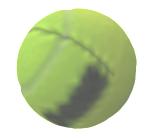} & Tennis ball & 056 & Sports \\
    \hline
    8 & \includegraphics[width=2.5cm]{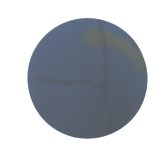} & Racquetball & 057 & Sports \\
    \hline
    9 & \includegraphics[width=2.5cm]{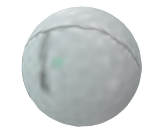} & Golf ball & 058 & Sports \\
    \hline
    10 & \includegraphics[width=2.5cm]{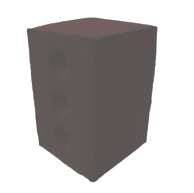} & Foam brick & 061 & Toys \\
    \hline
    11 & \includegraphics[width=2.5cm]{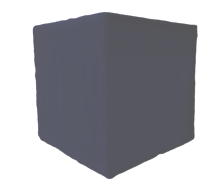} & Colored wood blocks & 070-b & Toys \\
    \hline
    12 & \includegraphics[width=2.5cm]{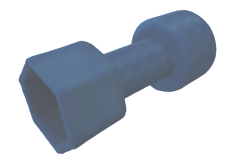} & Toy airplane & 072-c & Toys \\
    \hline
    13 & \includegraphics[width=2.5cm]{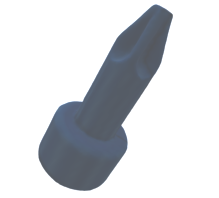} & Toy airplane & 072-d & Toys \\
    \hline
    14 & \includegraphics[width=2.5cm]{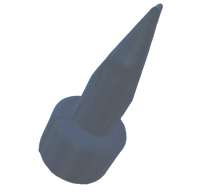} & Toy airplane & 072-e & Toys \\
    \hline
    15 & \includegraphics[width=2.5cm]{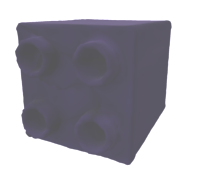} & Lego Duplo & 073-b & Toys \\
    \hline
    16 & \includegraphics[width=2.5cm]{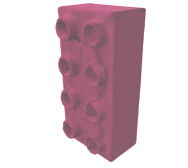} & Lego Duplo & 073-c & Toys \\
    \hline
    17 & \includegraphics[width=2.5cm]{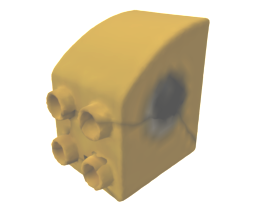} & Lego Duplo & 073-d & Toys \\
    \hline
    18 & \includegraphics[width=2.5cm]{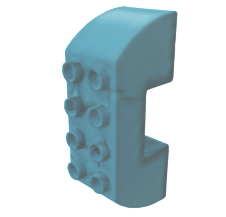} & Lego Duplo & 073-e & Toys \\
    \hline
    19 & \includegraphics[width=2.5cm]{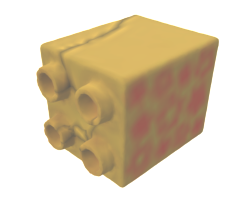} & Lego Duplo & 073-f & Toys \\
    \hline
    20 & \includegraphics[width=2.5cm]{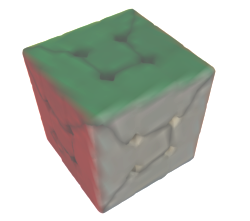} & Rubik's cube & 077 & Toys \\
    \hline
\end{longtable}

\subsection{Why TARGO Uses Packed Scenes}
\label{sec:reason_packed}

\begin{figure}[htbp]
    \centering
    \begin{subfigure}[b]{0.3\textwidth}
        \centering
        \includegraphics[width=\textwidth]{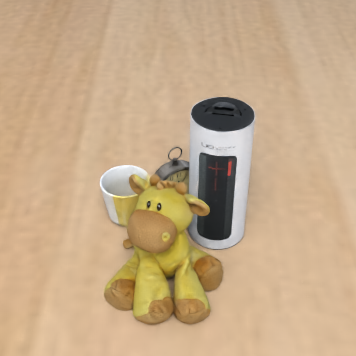}
        \caption{Packed setup}
        \label{fig:packed_setup}
    \end{subfigure}
    \hspace{0.05\textwidth}
    \begin{subfigure}[b]{0.3\textwidth}
        \centering
        \includegraphics[width=\textwidth]{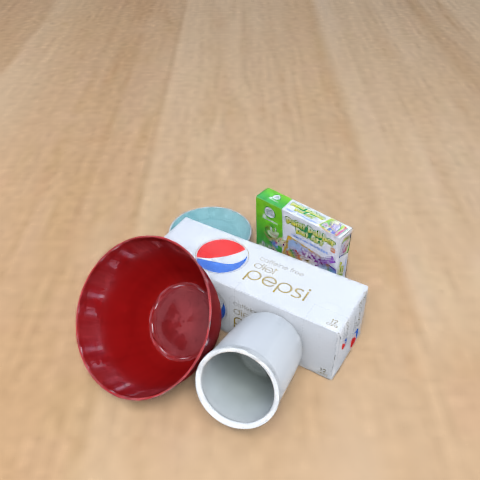}
        \caption{Pile setup}
        \label{fig:pile_setup}
    \end{subfigure}
    \caption{Packed and pile scenarios.}
    \label{fig:packed_pile}
\end{figure}

VGN~\cite{vgn} defines two scene types, shown in~\Cref{fig:packed_pile}. The \textit{pile} setup drops objects randomly into a box the size of the workspace and then removes the box, leaving a cluttered pile. The \textit{packed} setup places objects in their canonical poses at random tabletop positions. TARGO uses only packed scenes because constructing a paired single scene requires the target pose to stay unchanged after the occluders are removed. In a pile, an occluder may support the target, so removing it can shift the target and invalidate the pose-matched pair.

%% file: sections/da_statement.tex
\section*{Data and Code Availability Statement}

The TARGO dataset introduced in this paper (comprising TARGO-Synthetic training set, TARGO-Synthetic test set, and TARGO-Real set), including all object meshes, synthetic and real-world RGB-D images from cluttered and single scenes, and 6-DoF ground-truth grasp poses, is publicly available at \url{https://targo-benchmark.github.io}. The TARGO-Synthetic dataset is provided under the BSD 3-Clause License, and the TARGO-Real dataset is available under the Creative Commons Attribution-NonCommercial 4.0 (CC BY-NC 4.0) license.

All source code for dataset generation and loading, training, inference, and visualization to reproduce the experimental results presented in this paper is openly accessible at \url{https://github.com/targo-benchmark/targo} under the MIT License. Description of TARGO data schema and instructions for downloading pre-trained TARGO-Net model weights are included in the repository's README. The code repository is released under the MIT license.

Additional materials or clarifications can be obtained from the corresponding author upon reasonable request.